\documentclass{article}

\PassOptionsToPackage{numbers, compress}{natbib}

\usepackage[preprint]{neurips_2024}
\usepackage[T5]{fontenc}
\usepackage{amsmath} 
\usepackage{graphicx}
\usepackage{multirow}
\usepackage{array} 
\usepackage{xr-hyper}

\usepackage{hyperref} 
\usepackage{times}
\usepackage{latexsym}
\usepackage{booktabs}
\usepackage{array}
\usepackage{amsmath}
\usepackage{mathabx}
\usepackage{multirow}
\usepackage{multicol}
\usepackage[numbers]{natbib}
\usepackage{tabularx}  
\usepackage{booktabs}  
\usepackage{makecell}  

\usepackage[T5]{fontenc}

\usepackage[utf8]{inputenc}
\usepackage{enumitem} 

\usepackage{microtype}

\usepackage{inconsolata}

\usepackage{graphicx}

\usepackage{mathrsfs}
\usepackage{lipsum} 
\usepackage[normalem]{ulem}
\usepackage{wrapfig}
\usepackage{float}
\usepackage[T5]{fontenc}

\usepackage[utf8]{inputenc}
\usepackage{enumitem} 

\usepackage{microtype}

\usepackage{inconsolata}

\usepackage{graphicx}

\usepackage{mathrsfs}
\usepackage{lipsum} 
\usepackage[normalem]{ulem}
\usepackage{tcolorbox}

\usepackage[switch]{lineno}

\usepackage{booktabs} 
\usepackage{array}
\usepackage{booktabs}
\usepackage{multirow}
\usepackage{siunitx}

\usepackage{color,soul}
\usepackage{caption}
\usepackage{subcaption}
\usepackage{array}

\usepackage[utf8]{inputenc}
\usepackage{mathtools, nccmath}

\usepackage{adjustbox}

\usepackage{pythonhighlight}
\usepackage{fontawesome}
\usepackage{xcolor}
\hypersetup{
    colorlinks=true,
    linkcolor=orange,
    citecolor=green,
    urlcolor=cyan
}

\title{A Wearable Device Dataset for Mental Health Assessment Using Laser Doppler Flowmetry and Fluorescence Spectroscopy Sensors}

\author{Minh Ngoc Nguyen$^{*,1,7}$, Khai Le-Duc$^{*,2,3}$, Tan-Hanh Pham$^{*,4}$, Trong Nhan Nguyen$^{9}$,  \\ {\bf Bailey Trang$^{6}$, Ba Kien Tran$^{5}$, Viktor Dremin$^1$, Sergei Sokolovsky$^{1}$,} \\ {\bf Edik Rafailov$^{\dagger, 1}$,  Truong-Son Hy$^{\dagger, @, 8}$}\\
$^1$Aston University, Birmingham, UK
$^2$University of Toronto, Toronto, Canada \\
$^3$University Health Network, Toronto, Canada
$^4$Florida Institute of Technology, Melbourne, USA \\
$^5$Hai Duong Central College of Pharmacy, Hai Duong, Vietnam
$^6$Stanford University, Stanford, USA \\
$^7$Industrial University of Ho Chi Minh City, Ho Chi Minh City, Vietnam \\
$^8$The University of Alabama at Birmingham, Birmingham, USA\\
$^9$Pukyong National University, Busan, Republic of Korea \\
$*$ These authors contributed equally. \\
$\dagger$ These authors jointly supervised this work. \\
$@$ Correspondence to \textbf{thy@uab.edu}
\\[0.5em]
\Large {\faGithubSquare}   \href{https://github.com/leduckhai/Wearable_LDF-FS}{Wearable\_LDF-FS}\\
}

\begin{document}
\maketitle
\begin{abstract}

\textbf{Background:}
Mental health conditions such as depression, anxiety, and stress are commonly assessed using self-reported questionnaires and limited wearable physiological measures. However, reliance on subjective reporting, restricted sensor modalities such as heart rate variability and electrodermal activity, and small or homogeneous datasets may limit generalizability. We aimed to evaluate whether wearable optical sensing of microcirculation and tissue metabolism enables objective assessment of stress-related mental health states.

\textbf{Methods:}
We conducted a prospective observational study including 132 adults aged 18 to 94 years (58\% female) from 19 countries. Participants underwent repeated fingertip measurements using a non-invasive wearable device combining laser Doppler flowmetry and fluorescence spectroscopy to capture microvascular perfusion and metabolic signals. Frequency-domain features were extracted using wavelet analysis. Depression, anxiety, and stress levels were assessed using a standardized 21-item questionnaire. Multiple machine learning models were evaluated under subject-wise validation, and model interpretability was assessed using Shapley-based feature attribution.

\textbf{Results:}
Here we show that ensemble-based models distinguish individuals with stress-related symptoms from those without with a receiver operating characteristic area under the curve of 0.72 and a precision–recall area under the curve of 0.89 under subject-wise validation. Microcirculatory variability and metabolic fluorescence features contribute substantially to prediction performance. Demographic variables, including sex, age, body mass index, and heart rate, are associated with increased stress-related risk.

\textbf{Conclusions:}
Wearable optical sensing combined with interpretable machine learning provides physiological signatures associated with stress-related mental health conditions. This framework supports development of scalable and data-driven tools for objective mental health monitoring. 

\end{abstract}

\section*{Plain Language Summary}

Mental health problems such as stress, anxiety, and depression affect millions of people worldwide. These conditions are usually assessed using questionnaires, which rely on how people describe their own feelings. In this study, we explore whether a wearable device can help measure mental health using physical signals from the body. The device records small changes in blood flow and tissue activity from the fingertip. We collected data from 132 adults across 19 countries and compared these signals with mental health questionnaire results. We found that patterns in blood flow and tissue activity are linked to stress-related symptoms. This approach may help develop new tools for simple, non-invasive mental health monitoring in everyday life.

\section{Introduction}
\label{sec.intro}


During the past two decades, the global incidence of common mental disorders (CMDs), particularly anxiety and depression, has fluctuated significantly and increased substantially due to improved awareness and diagnosis in healthcare settings \cite{wu2023changing}. However, the increase in CMDs is not uniform across age groups, with higher rates among younger individuals due to changing social pressures and lifestyle factors\cite{krokstad2022divergent}. Economic conditions and public health crises also influence mental health trends, highlighting the need for adaptable and accessible mental health services in the healthcare system. \cite{dykxhoorn2024temporal}. Mental health has gained significant attention, particularly after the COVID-19 pandemic, which exacerbated mental health problems \cite{lange2021coronavirus, kola2021covid}.

In the United Kingdom, more than 25\% of people experience a mental health disorder annually, with 1 in 6 adults facing anxiety or depression weekly; 
stress leading to overeating (46\%), increased alcohol consumption (29\%) and elevated smoking rates (16\%). CMDs harm various body systems, including elevated blood pressure and increased heart risks in the cardiovascular system, altering learning and mood in the nervous system, causing tension and fatigue in muscles, resulting in shallow breathing, and leading to weight changes and risk of diabetes in metabolism. Ultimately, stress has an extensive affect on mental and physical well-being \cite{Mentalhealthstatistics2024}.

Stress can have a detrimental impact on various body systems \cite{chrousos2009stress}. Prolonged stress can elevate blood pressure and heart rate, increasing the risk of cardiovascular diseases \cite{steptoe2012stress}. It also affects the nervous system, leading to cognitive decline, mood disorders, and an increased risk of mental disorders \cite{calabrese2009neuronal}. Muscular tension, soreness, and fatigue can result from stress, affecting daily activities \cite{umer2022quantifying}. Changes in breathing patterns due to stress can lead to respiratory problems \cite{pedersen2010influence}. Stress can also alter metabolism, potentially leading to weight changes and a higher risk of diabetes \cite{harris2017stress}. In conclusion, stress negatively affects both mental and physical health, impacting systems such as cardiovascular, nervous, muscular, respiratory, and metabolic.

Mental health assessment encompasses various methods to ensure a comprehensive and accurate understanding. Standardized tests such as DASS (Depression Anxiety Stress Scales) \cite{lovibond1995depression},  the Beck Depression Inventory (BDI) \cite{beck1987beck}, and the Beck Anxiety Inventory (BAI) \cite{beck1993beck} Clinical interviews, structured, semi-structured, or unstructured, measure levels of depression, anxiety, and stress, helping psychologists gather detailed information through specific questions and conversations. Biological assessments, including tests for neurotransmitter levels such as serotonin and dopamine, and electroencephalograms (EEGs) to monitor brain activity, also play a crucial role \cite{miranda2019overview}, and functional magnetic resonance imaging (fMRI) to observe brain activity during psychological tasks \cite{whitten2012functional}. Biosensors for psychiatric biomarkers (e.g. cortisol, dopamine, serotonin) can diagnose and manage disorders via samples of blood, saliva, urine, and sweat. They offer high sensitivity, selectivity, and real-time monitoring, but face challenges such as environmental accuracy, high costs, and data integration. Therefore, further development is needed for better effectiveness \cite{wang2024biosensors}.

Among wearable-based approaches, physiological signals such as heart rate variability (HRV) and electrodermal activity (EDA) are widely adopted for mental health assessment due to their ability to reflect autonomic nervous system activity. However, these modalities primarily provide indirect markers of psychological states and are highly sensitive to confounding factors, including motion artifacts, ambient temperature, skin hydration, and device-specific calibration \cite{elgendi2026wearable, pei2026quantitative}. Such limitations may reduce robustness in real-world monitoring scenarios.

In contrast, laser Doppler flowmetry (LDF) enables direct assessment of microvascular blood perfusion, while fluorescence spectroscopy (FS) provides information related to biochemical and metabolic tissue characteristics. These signals are influenced by stress-induced vascular regulation and metabolic alterations, offering complementary physiological perspectives that are not directly accessible through HRV or EDA alone. Rather than aiming to replace existing wearable modalities, this study explores the feasibility and potential added value of integrating LDF and FS signals for mental health assessment in a non-invasive wearable setting.

The DASS-21 questionnaire, a short version of the 42-item DASS, includes 21 items divided into three subscales: Depression, Anxiety, and Stress. It assesses motivation loss, anxiety symptoms, and irritability, respectively. Validated in clinical and community settings, DASS-21 shows excellent internal consistency with Cronbach's alpha values of 0.94 for depression, 0.87 for anxiety, and 0.91 for stress. The severity levels and cut-off points of DASS-21 classify and quickly support patients \cite{monteiro202312}. Intense emotions such as anxiety or anger can affect the hands by altering blood flow and electrical activity of the muscles, causing muscle tension or relaxation \cite{mcgaugh2013emotions}. Despite many articles on blood circulation in such individuals, none compare blood circulation variability in stressed vs. non-stressed people. This study demonstrated the ability of the wearable device to differentiate cardiovascular parameters between stress and non-stress groups on both the middle fingers.

Wearable devices with LDF and FS channels offer a promising and complementary approach to assess microcirculation and obtain physiological and metabolic information relevant to mental health assessment. Although these studies demonstrate their potential under normal and pathological conditions, more research with larger cohorts is essential for clinical implementation. One of the crucial tasks is to investigate the effects of various treatment protocols and lifestyle changes on microcirculatory and metabolic parameters using these wearable devices. Another important direction is to develop machine learning algorithms for automated data analysis and interpretation, which can significantly enhance the diagnostic capabilities of wearable devices. Our research focuses on building a diverse dataset for mental health detection using a non-invasive wearable device equipped with LDF and FS channels. By exploring subcutaneous blood microcirculation across demographics, our aim is to provide valuable information and pioneer the development of a large dataset for mental health assessment.


Professor E. Rafailov's research group at Aston University has developed LDF/FS wearable devices using VCSELs, showing signal responses comparable to conventional monitors in volunteer assessments \cite{inproceedings19,zharkikh2022,dremin2025}. These devices employ LDF and FS for non-invasive early detection of vascular complications in diabetes and other conditions. LDF assesses tissue perfusion and blood flow regulatory mechanisms \cite{low2020historical,rajan2009}, while FS can detect changes in metabolic activity, accumulation of AGEs, and other pathology-associated fluorophores \cite{dremin2017,dremin2023chapter}.

Introduced more than 30 years ago, LDF uses laser radiation to probe tissue and analyze backscattering from moving red blood cells. The main parameter recorded is the microcirculation or perfusion index, essential for organ nutrition, adaptation, and regulation. The method uses wavelet transform, specifically adaptive wavelet analysis with complex-valued Morlet wavelets, to assess microvessel oscillatory processes over a wide frequency range. This has been the standard for over 20 years, replacing the Fast Fourier Transform (FFT) and Butterworth filters \cite{stefanovska1999,kralj2023wavelet}. The continuous wavelet transform is preferred for non-stationary LDF signals (perfusion) due to its optimal ``time-frequency'' resolution, which effectively tracks frequency and amplitude fluctuations in blood flow signals. Currently, several principal frequency bands are distinguished in microvascular oscillations, reflecting various regulatory mechanisms: endothelial endothelial ($0.0095$--$0.02$\,Hz), neurogenic ($0.02$--$0.06$\,Hz), myogenic ($0.06$--$0.16$\,Hz), respiratory ($0.16$--$0.4$\,Hz), and cardiac ($0.4$--$1.6$\,Hz) \cite{bagno2015wavelet, kvandal2006}. The FS method uses laser probing to record fluorescence spectra of tissue fluorophores, including metabolic coenzymes, such as NADH and FAD. This detects changes in endothelial cell metabolic activity, indicating various physiological and pathological processes and identifying cellular metabolic disorders associated with disease \cite{dunaev2015,zharkikh2020biophotonics}.

Several studies have employed these wearable optical devices to assess blood microcirculation across diverse patient populations, revealing distinct physiological and pathological patterns. In diabetic populations, wearable LDF devices have consistently detected microvascular dysfunction, including reduced perfusion in the lower extremities, impaired endothelial and neurogenic regulatory mechanisms, and decreased complexity of blood flow oscillations in areas prone to ulceration \cite{zharkikh2024,Zherebtsov2023,Xing2024}. The sensitivity of wearable LDF technology was further confirmed by Saha et al., who showed that smokers exhibit lower blood perfusion and reduced amplitudes of endothelial, neurogenic, and myogenic oscillations compared to non-smokers, indicating nicotine-induced vasoconstriction \cite{saha2020wearable}. Also, Fedorovich et al. demonstrated that body position significantly affects microcirculatory parameters, with orthostatic changes decreasing perfusion in the lower extremities while forehead perfusion remains stable, highlighting the importance of standardized protocols \cite{BodyPosition}. Collectively, these studies validate developed wearable devices as a powerful tool for detecting microvascular dysfunction in various conditions. Further description of device parameters and related works are provided in the Supplementary Informations Section.


In this study, we make three key contributions to the field of mental health assessment, placing particular emphasis on our data collection methods and the application of Explainable AI (XAI):

\begin{itemize}
    \item \textbf{Establishment of one of the largest and most comprehensively characterized multimodal LDF \- FS datasets for mental health research}: 
    We present a novel framework for objective mental health assessment through the construction of what is, to our knowledge, one of the largest and most systematically characterized datasets integrating LDF and FS signals in psychiatric research. The dataset comprises physiological measurements from 132 participants, acquired using wearable sensing devices and rigorously linked with validated psychological assessments using the Depression, Anxiety, and Stress Scale (DASS-21). By simultaneously capturing microvascular dynamics and tissue metabolic characteristics, this multimodal dataset enables comprehensive investigation of vascular–metabolic signatures associated with mental health status and provides a robust foundation for physiological biomarker development.

    \item \textbf{Development of a rigorously evaluated machine learning framework for multimodal physiological classification}:
    We benchmarked multiple machine learning architectures under consistent validation schemes to assess their ability to classify mental health conditions using structured physiological features derived from both LDF and FS signals. Ensemble-based models demonstrated robust generalization within this multimodal feature space.
    
    \item \textbf{Unveiling the \textit{``AI black box''} by using XAI}: Recognizing the critical role of interpretability in mental health applications, we employ XAI techniques to investigate the decision-making behind a machine learning model. Using XAI, we aim to illuminate the specific features within the wearable device data that exert the strongest influence on the prediction of health issues of a person. 
\end{itemize}

All related code and data are published online.

\section{Methods}

\subsection{Study Design and Dataset Description}
\label{sec.study_design_and_data}


\begin{figure}[h]
    \centering
    \includegraphics[width=\linewidth]{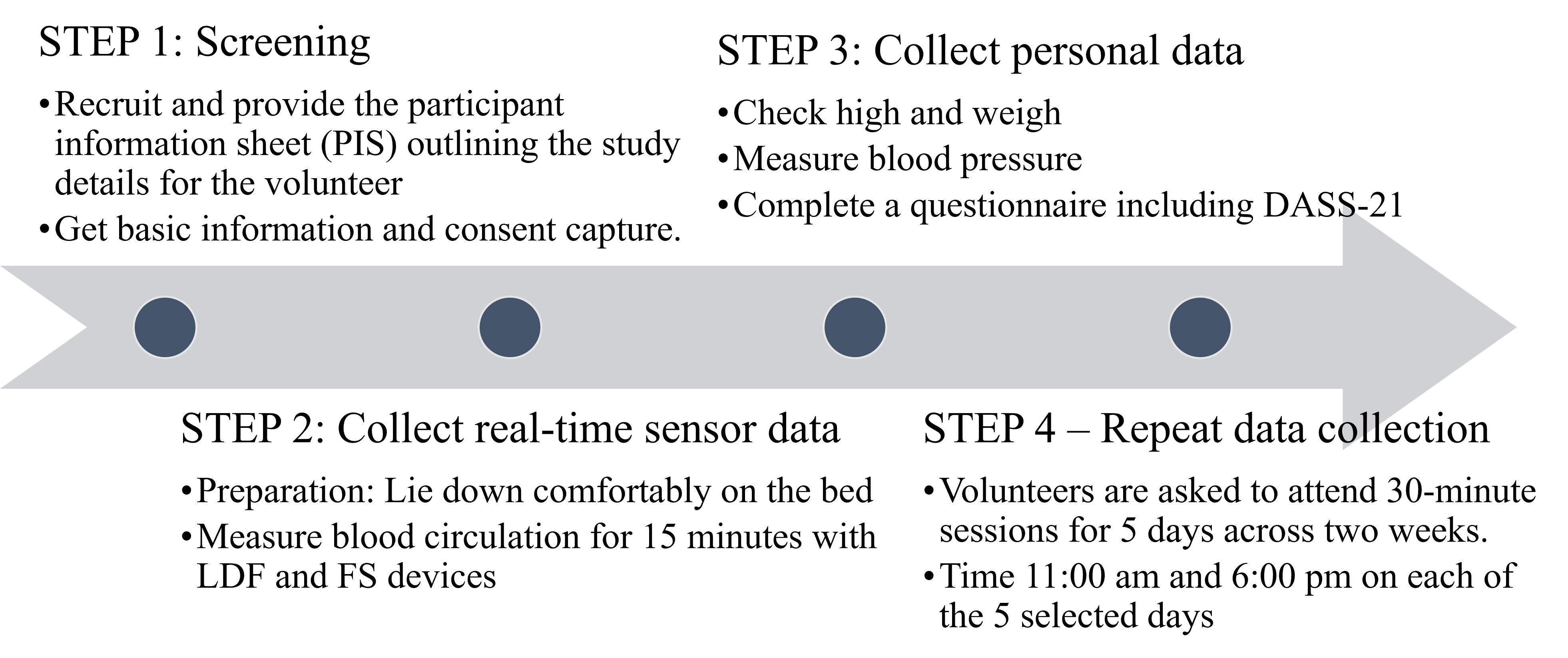}
    \caption{\textbf{Data collection workflow.} The four-step protocol includes participant recruitment, wearable device measurement, anthropometric assessment, and DASS-21 questionnaire completion.}
    \label{fig.datastep}
\end{figure}

{This study included a total of 132 participants recruited from the general population across 19 countries. The cohort comprised individuals aged 18 to 94 years, with a balanced gender distribution. Participants were enrolled to ensure diversity in demographic characteristics, including age, gender, and geographic background. Inclusion criteria required participants to be adults (greater than 18 years) and capable of providing informed consent. Exclusion criteria included the presence of dermatological conditions affecting the hands or fingertips, which could interfere with optical signal acquisition. Additional participant information, including physiological and lifestyle factors (e.g., body mass index, heart rate, smoking status), was collected to support downstream analysis.} There are four steps in data collection, as shown in Figure~\ref{fig.datastep}. Firstly, participants were recruited from the general population and included volunteers aged 18 and older. To ensure accurate blood perfusion measurements, people with dermatological conditions in the hands and middle fingers were excluded from the study. Before starting the study, all participants received a detailed explanation of the study design and its objectives. After giving informed consent, participants completed a questionnaire detailing their current health status, including medication history, alcohol consumption within the last 24 hours, and exercise habits such as cycling, treadmill, or jogging.

\begin{figure}[h]
    \centering
    \includegraphics[width=\linewidth]{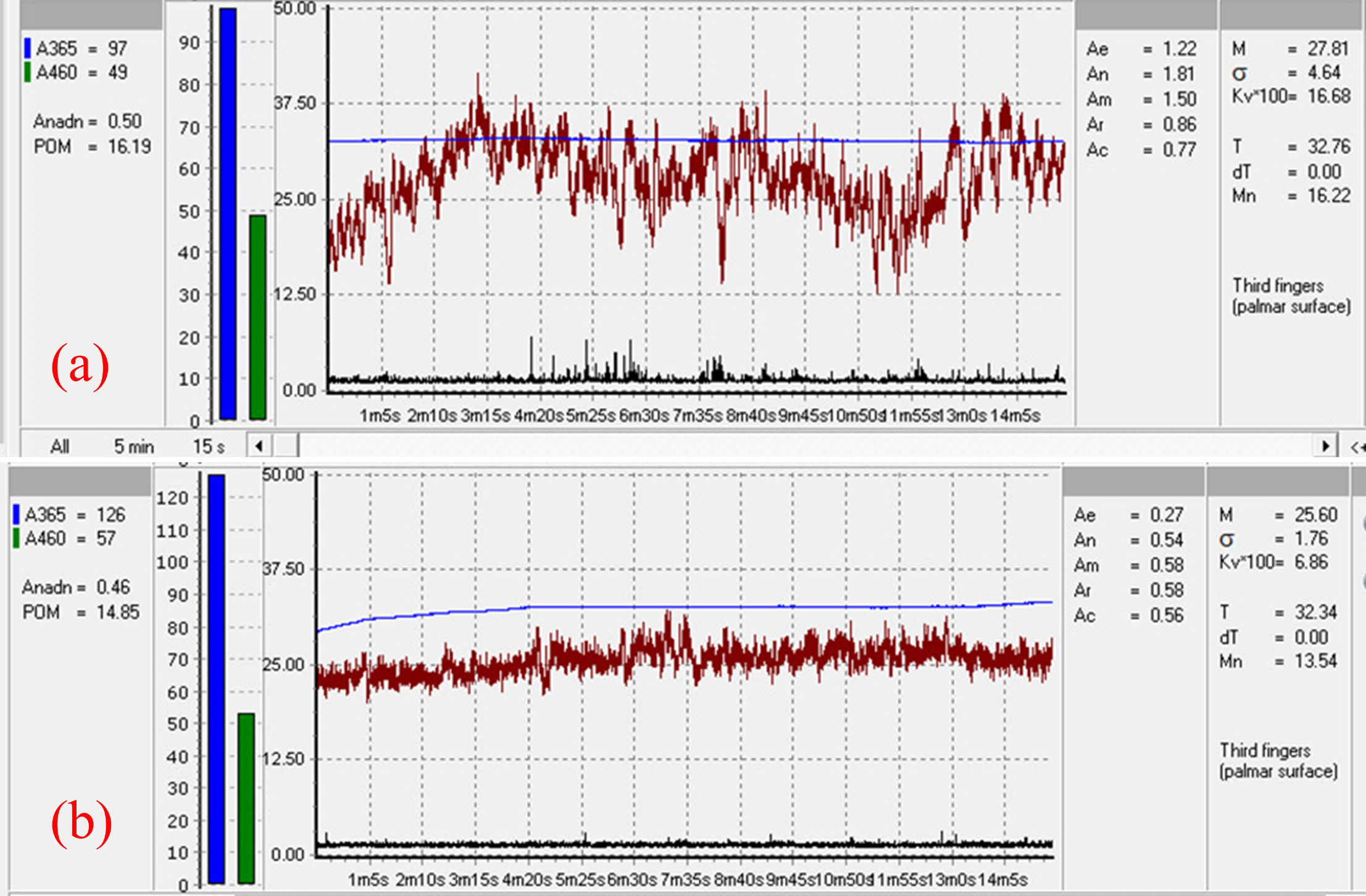}
    \caption{{\textbf{Representative wearable device data samples.} 
(a) Data from a stressed individual (36-year-old female, moderate stress, 
anxiety, and depression, right hand). (b) Data from a well-being individual 
(27-year-old female, right hand). Each panel shows raw signals recorded 
from both left (top) and right (bottom) hands. Notably, the stressed 
individual exhibits greater signal fluctuation compared to the well-being 
individual. M: microcirculation index (perfusion units, PU); T: temperature; 
LDF: Laser Doppler Flowmetry; FS: Fluorescence Spectroscopy.}}
    \label{fig.datacollection}
\end{figure}

Sequentially, blood perfusion parameters were measured non-invasively with participants in a supine position to ensure physical and mental rest. To minimize external stimuli, participants were instructed not to read, write, or talk during the test.  Blood perfusion data was collected from sensors placed on the middle fingertips of the left and right hands for a duration of eight minutes. To control possible confounders, participants were asked not to consume caffeine and alcohol-containing beverages at least twelve hours before the designated measurement time. 

Figure~\ref{fig.datacollection}(a) shows the data measured from a stressed individual, with data from the left hand illustrated on the top and data from the right hand on the bottom. Similarly, Figure~\ref{fig.datacollection}(b) presents an example of well-being data collected using wearable devices. As observed, the data from the stressed individual exhibits significant fluctuations, while the data from the well-being individual are more stable. In addition, the definitions of the parameters of the measurement device are described in Table \ref{table.wearabeldeviceparameters}.

\begin{table}[h]
\centering
\caption{Definitions of the measurement device parameters.}
\renewcommand{\arraystretch}{1.15} 
\begin{tabular}{lm{6.5cm}}
\toprule
\multicolumn{1}{c}{{Parameters}} & \multicolumn{1}{c}{{Definition}}\\ 
\midrule
M & Microcirculation index, indicating the average perfusion of microvessels (in PU). \\ 
$\sigma$ & Mean square deviation of blood flow oscillation amplitude (in PU). \\ 
Kv & Coefficient of blood flow variability. \\
A365 & Backscatter amplitude at the laser source wavelength for NADH excitation. \\ 
A460 & NADH fluorescence amplitude at 460 nm. \\
NADH & Relative amplitude of NADH fluorescence, considering the optical characteristics of the study tissue region. \\ 
POM & Index of oxidative metabolism linked to the nutritional component of blood perfusion and NADH coenzyme fluorescence amplitude. \\ 
Ae & Average maximum amplitude of blood flow within the endothelial oscillation range. \\ 
An & Average maximum amplitude of blood flow within the neurogenic oscillation range. \\ 
Am & Average maximum amplitude of blood flow within the myogenic oscillation range. \\ 
Ar & Average maximum amplitude of blood flow within the respiratory oscillation range. \\ 
Ac & Average maximum amplitude of blood flow within the cardiac oscillation range. \\ 
Fe & Endothelial oscillation frequency (0.0095 - 0.02 Hz). \\ 
Fn & Neurogenic oscillation frequency (0.02 - 0.06 Hz). \\ 
Fm & Myogenic oscillation frequency (0.06 - 0.16 Hz). \\ 
Fr & Respiratory oscillation frequency (0.16 - 0.4 Hz). \\ 
Fc & Cardiac oscillation frequency (0.4 - 1.6 Hz). \\ 
T & Temperature at the measurement site. \\ 
\bottomrule
\end{tabular}
\label{table.wearabeldeviceparameters}
\end{table}

Following the 15-minute blood circulation measurement, we measured height and weight. The participants then completed the DASS-21 questionnaire, which assesses how much each statement applied to them over the past week. After completing the questionnaire, their blood pressure was measured. The measurements were taken twice a day: in the morning (around 11.00, before lunch) and in the afternoon (around 15.00, after lunch) for any five days over two consecutive weeks.

The DASS-21 is used to assess key symptoms of depression, anxiety, and stress, as well as patient reactions to treatment. It has been proven to have adequate psychometric properties and is equivalent to other accurate scales. The 21 elements comprise three self-reported scales, each with seven elements classified on a Likert scale from 0 to 3. The depression, anxiety, and stress scores are measured by summing the scores of the related items. It is important to note that participants classified as 'normal' based on DASS-21 scores represent individuals who did not report significant symptoms of depression, anxiety, or stress during the past week. This classification does not confirm clinical mental wellness or rule out subclinical conditions, transient psychological states, or self-reporting biases such as symptom suppression or social desirability effects. Since DASS-21 is a shorter version of the original 42-item DAS, the score for each subscale must be multiplied by 2 to calculate the final score. The recommended cut-off scores for conventional severity labels (normal, moderate, severe) are calculated using Table \ref{tab:DAS21results}. The scores on the DASS-21 will need to be multiplied by 2 to calculate the final score.

\begin{table}[h]
    \centering
    \caption{Scores on the DASS-21 will need to be multiplied by two to calculate the final score.}
    \begin{tabular}{lccc}
        \toprule
        \multicolumn{1}{c}{{Level}} & \centering {Depression} & {Anxiety} & {stress} \\ \midrule
        Normal & 0-9 & 0-7 & 0-14\\
        Mild & 10-13 & 8-9 & 15-18\\ 
        Moderate & 14-20 & 10-14 & 19-25 \\ 
        Severe & 21-27 & 15-19 & 26-33 \\ 
        Extremely Severe & 28+ & 20+ & 34+ \\ \bottomrule
    \end{tabular}
    \label{tab:DAS21results}
\end{table}

According to the manual, the ratings are classified as: \textit{normal}, \textit{mild}, \textit{moderate}, \textit{severe}, or \textit{extremely severe}; all those who exhibit any signs of stress, anxiety, or depression, we call the stressed group, and the remaining individuals will be classified as the well-being group. This allowed real-time control of the course of the experiment and analysis of the recorded parameters. 

The displayed parameters show the raw data on blood perfusion, temperature, and movement of the fingertip and wrist. After acquiring the data, the oscillation rhythms of each measurement were analyzed using the built-in module ``wavelet analysis''. Wavelet analysis is used to determine the maximum amplitude of blood perfusion in five characteristic oscillations. Five rhythmic oscillations are isolated from LDF recordings with the help of wavelet analysis: endothelial (frequency interval 0.0095 to 0.02 Hz), neurogenic (0.02-0.06 Hz), myogenic (0.06 to 0.16 Hz), respiratory (0.16 to 0.4 Hz) and cardiac or pulse rhythm (0.4 to 1.6 Hz).

\begin{figure}[h]
    \centering
    \includegraphics[width=\linewidth]{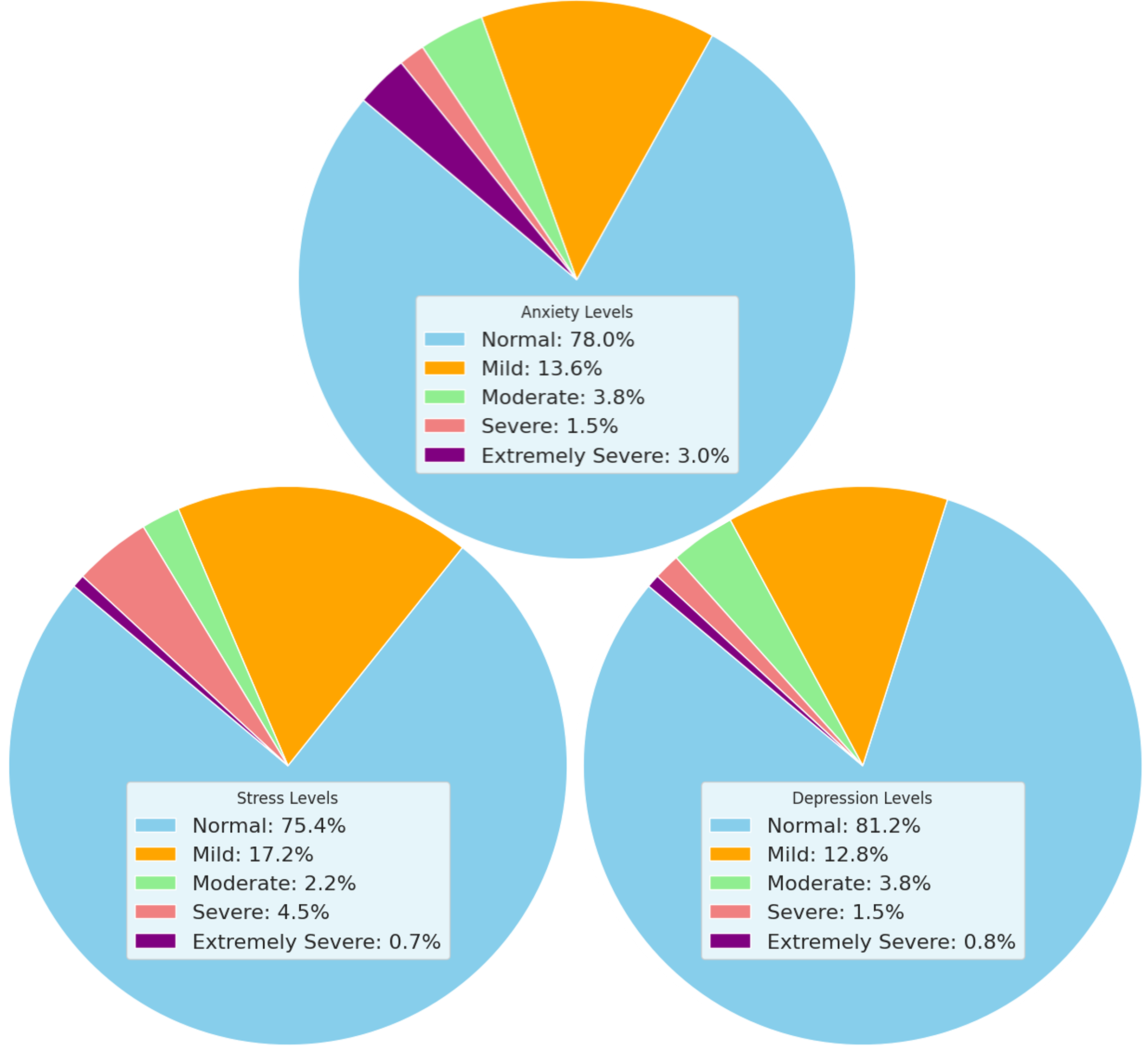}
    \caption{{\textbf{Distribution of mental health severity levels across the study cohort.} Bar charts show the proportion of participants (n = 132) classified at each severity level for stress, anxiety, and depression based on DASS-21 scoring. Severity categories follow established cut-offs: Normal, Mild, Moderate, Severe, and Extremely Severe.}}
    \label{fig.datadistribution0}
\end{figure}

As illustrated in Figure~\ref{fig.datadistribution0}, the total number of people with mental health problems reaches 27.3\% of the population, with more than 50\% of them experiencing combined stress, anxiety and depression. The incidence rates of stress, anxiety, and depression are 24.5\%, 22\% and 18.2\%, respectively, mostly at mild levels, accounting for 17.2\%, 13.6\% and 12.8\% in these groups. The extremely severe level is highest in the anxiety group at 3.0\%, while in the other two groups, it is below 1\%.

Further details of data collection and data analysis are described in Supplementary Information Section \ref{sec.appendix_study_design_and_data}.

\subsection{Machine Learning and Explainable Artificial Intelligence}
\label{sec.machinelearningandXAI}
Further details of the experimental setup are described in Supplementary Information.

\subsubsection{Machine Learning Models for DAS Prediction}
\label{sec.MLmodel}

To identify the most effective approach for predicting depression, anxiety, and stress levels, we explored various machine learning algorithms including Support Vector Machine (SVM), Random Forest Classification, Gradient Boosting Classifier, CatBoost, LightGBM, as well as Multi-layer Perceptron (MLP) \cite{mahesh2020machine}. In addition, we employ two primary approaches to train machine learning models for predicting DAS levels: binary classification and multi-class classification. Both approaches leverage data from the DASS-21 questionnaire along with potentially other features of the collected data set. In addition, we consider three cases to investigate the models' performances: using all features collected, using only features extracted from wearable devices, and using the top-10 important features. 

For binary classification, this approach simplifies the prediction task by transforming the DAS levels into a binary classification problem. We categorize participants into two classes based on their DASS-21 scores:
\begin{itemize}
    \item \textbf{Normal:} This class comprises participants who score within the normal range for depression, anxiety, and stress according to established DASS-21 scoring guidelines.
    \item \textbf{Abnormal:} This class includes participants whose DASS-21 scores indicate possible symptoms of depression, anxiety, or stress.
\end{itemize}

For multi-class classification, this approach aims for a more granular prediction by treating DAS levels as a multi-class classification problem. Instead of dividing mental health states into two categories, we define multiple classes based on established DASS-21 scoring ranges. Normal, stress, stress anxiety, and stress anxiety depression. We employ two standard strategies: One-vs-Rest (OvR) and One-vs-One (OvO). These approaches enable binary classifiers to handle multi-class problems.

\begin{itemize}
    \item \textbf{OvR Strategy:} In the OvR approach, the classifier trains $N$ separate binary models for $N$ classes. For each model, one class is treated as the ``positive'' class while all other classes are combined into a single ``negative'' class.
    
    \item \textbf{OvO Strategy:} In the OvO approach, the classifier trains $N \times (N-1)/2$ binary models, one for each pair of classes. Each model learns to distinguish between only two specific classes, ignoring all other classes during training.
\end{itemize}

In machine learning, dividing the data set into training and testing subsets is crucial to evaluate the performance of the model. In our ablation study, we use three train-evaluate techniques: Split 80:20, 5-fold patient-wise (not sample-wise) and Leave-one-patient-out (LOPO) \cite{hastie2009elements}. In doing this, we ensure that the model is evaluated on its ability to perform on new patients not seen during training.

To assess the performance of machine learning models to predict depression, anxiety, and stress (DAS) levels, we employ two key evaluation metrics: Receiver Operating Characteristic (ROC) AUC (Area Under the Curve) and Precision-Recall (PR) AUC. These metrics provide a comprehensive assessment of the model's discriminative ability and its performance in handling class imbalances.

\subsubsection{Explainable AI}

In healthcare applications, understanding the reasoning behind a model's predictions for DAS levels is crucial to building trust and confidence in its outputs. This empowers healthcare professionals and researchers to make informed decisions based on the predicted levels of DAS and the underlying factors that influence these predictions. In this study, we used SHAP (Shapley Additive Explanations) to achieve interpretability and gain insight into the model decision-making process for the prediction of DAS \cite{lundberg2017unified}. SHAP assigns an attribution value (SHAP value) to each feature for a given DAS prediction. High positive SHAP values indicate that the feature has a strong positive influence on the predicted DAS level (which could indicate a higher probability of depression, anxiety, or stress). In contrast, low negative SHAP values signify a negative influence (indicating a lower likelihood). This interpretability allows us to answer several key questions:

\begin{itemize}
    \item \textbf{Identification of key physiological and psychological indicators:} What are the features of wearable sensor data and questionnaire scores of a patient that have the most significant influence on the model's predictions? 
    \item \textbf{Validation of Model Fairness and Mitigation of Bias:} Are the model's predictions fair across different demographics (age, gender, etc.)? Examining SHAP values across these groups helps ensure that the model is not unfairly biased toward certain populations.
    \item \textbf{Enhanced model transparency:} How does the model arrive at its predictions? By explaining the rationale behind the predictions of the model through SHAP values, we can foster trust and confidence in its use among healthcare professionals and researchers. 
\end{itemize}

\subsection{Statistics and Reproducibility}

All statistical analyses were conducted using Python (version 3.9) and relevant machine learning libraries. Model performance was evaluated using receiver operating characteristic area under the curve (ROC-AUC) and precision recall area under the curve (PR-AUC). To ensure robustness and generalizability, we employed subject-wise validation strategies, including 5-fold cross-validation and leave-one-patient-out (LOPO) validation, ensuring that data from the same participant did not appear in both training and testing sets.

The study included 132 participants, and repeated measurements were collected across multiple sessions. Each measurement session was treated as an independent sample, while subject-wise validation ensured independence at the participant level.

No statistical methods were used to predetermine sample size; however, the sample size is comparable to previous studies in wearable-based mental health assessment. All experiments were conducted with consistent preprocessing and model configurations to ensure reproducibility.

To ensure reproducibility, all source code and dataset are publicly available at: \url{https://github.com/leduckhai/Wearable_LDF-FS}.

\subsection{Ethics Approval}

All procedures involving human participants were conducted in accordance with the Declaration of Helsinki. Ethical approval for this study was obtained from the Human Ethics Committee of Aston University, Birmingham, United Kingdom, and from the Ethics Committee of Hai Duong Central College of Pharmacy, Vietnam.  Ethics approval number: EPS21056.

All participants were provided with detailed information about the study objectives and procedures prior to participation.Written informed consent was obtained from all participants before data collection. All data were anonymized prior to analysis to ensure participant confidentiality and privacy.

\section{Results}
\label{sec.result}

\subsection{All Features with 80:20 Split}
\label{sec.result.8020}
In this section, we present the results of our investigation on the use of machine learning models to predict stress levels based on data from the DASS-21 questionnaire and potentially other features within our dataset. We employed both binary and multi-class classification approaches, evaluating the models on a random 80/20 train-test split to ensure generalizability.

\subsubsection{Binary Classification}
Our initial focus was on binary classification, with the objective of identifying people with potential mental health concerns based on their DASS-21 scores. For this task, the performance of each model is summarized in Table \ref{tab:ml_metrics_All_features_binary_classification}. 

From the table, LightGBM emerged as the model that performed the best, achieving the highest ROC AUC of 0.9941 and PR AUC of 0.9982. Gradient Boosting and MLP also demonstrated strong performance, with ROC AUC values of 0.9751 and 0.9322, respectively. In contrast, Catboost and Random Forest showed relatively lower performance, indicating that they might not be as effective for this particular binary classification task.
\begin{table}[ht]
    \centering
    \caption{Performance for different models: All features with 80:20 split, binary classification}
    \begin{adjustbox}{max width=1\textwidth}
    \setlength{\tabcolsep}{2pt} 
    \begin{tabular}{lccccccc}
        \toprule
        Model & Gradient Boosting & Catboost & LightGBM & SVM & Random Forest & MLP & EEGNet\\
        \midrule
        {ROC AUC} & 0.9751 & 0.7320 & 0.9941 & 0.9199 & 0.8145 & 0.9322 & 0.4034\\
        {PR AUC} & 0.9911 & 0.9104 & 0.9982 & 0.9720 & 0.9330 & 0.9767 & 0.8128\\
        \bottomrule
    \end{tabular}
    \end{adjustbox}
    \label{tab:ml_metrics_All_features_binary_classification}
\end{table}

\subsubsection{Multi-class Classification}
In addition to predicting whether a person has a mental issue or not, we also explored a multi-class classification task, aiming to predict not only the presence of stress but also its severity level. In particular, {Supplementary Table \ref{tab:ml_metrics_All_features_multi_classification}} details the performance metrics of the models on multi-class classification tasks, with the notable absence of MLP results.

LightGBM again stands out, achieving near-perfect Macro ROC AUC scores and high precision, recall, and F1 scores. Gradient Boosting and SVM also performed well, with Gradient Boosting showing balanced performance across all metrics. Catboost and Random Forest had lower scores, suggesting limitations in handling the complexities of multi-class classification in this context. The disparity in F1-scores, particularly between LightGBM (0.9391) and the other ensemble methods (ranging from 0.1783 to 0.5152), underscores the difficulty of the multi-class task given the complex nature of LDF and FS features. The superior F1-score of LightGBM confirms its unique capability to effectively categorize stress severity, whereas other models struggle with the overlapping patterns of different stress levels.

\subsubsection{Feature Importance}
To understand the factors that influence the predictions of the models, we analyzed the importance of various characteristics. Feature importance was assessed using Gradient Boosting, Catboost, and LightGBM models, as summarized in Table {Supplementary Table \ref{tab:merged_featuresbinary_prediction}} and {Supplementary Table \ref{tab:multiclass_features}}. The tables highlight the top 10 important features identified by each model. In both tables, features such as heart rate, BMI, weight, T (temperature), and type of skin consistently rank high in the top ten importance for most models. This suggests that physiological factors significantly influence the models' stress predictions. Other features, such as age, POM, A365, and Anadn, also appear to be relevant to some extent, depending on the model.

\subsection{All Features with Cross-Validation}
\label{sec.result.allfeatures_crossval}
In the field of health issue analysis, ensuring the robustness and reliability of predictive models is paramount. To achieve this, we employ cross-validation techniques such as k-fold cross-validation and LOPO cross-validation. 

\subsubsection{Binary Classification with LOPO}

LOPO cross-validation is particularly relevant in medical studies, where patient-specific variations can significantly impact the model's predictions. Table \ref{tab:BC_LOPO} presents the performance metrics for various machine learning models when evaluated using the LOPO cross-validation method for binary classification. LOPO is a stringent evaluation method in which the model is trained in all patients except one, who is then used as the test set. This process is repeated for each patient, ensuring that the model's performance is tested on unseen data in each iteration.

\begin{table}[ht]
\centering
\caption{Performance for different models: All features with LOPO, binary classification.}
    \begin{adjustbox}{max width=1\textwidth}
    \setlength{\tabcolsep}{2pt} 
    \begin{tabular}{lccccccc}
    \toprule
    Model & Gradient Boosting & Catboost & LightGBM & SVM & Random Forest & MLP & EEGNet\\
    \midrule
    {ROC AUC} & 0.6556 & 0.6001 & 0.6773 & 0.5316 & 0.6209 & 0.5313  & 0.4403\\
    {PR AUC} & 0.8806 & 0.8287 & 0.8998 & 0.8214 & 0.8630 & 0.8425 & 0.8303 \\
    \bottomrule
    \end{tabular}
    \end{adjustbox}
    \label{tab:BC_LOPO}
\end{table}

From the results, LightGBM shows the highest ROC AUC (0.6773) and PR AUC (0.8998), indicating better performance in distinguishing between the two classes compared to other models. Gradient Boosting and Random Forest also perform reasonably well, with ROC AUC values of 0.6556 and 0.6209, respectively. SVM and MLP perform the worst in terms of ROC AUC, indicating they might struggle more with the variability in patient data.

\subsubsection{Binary Classification with 5-folds}

As mentioned above, we also use 5-fold cross-validation to investigate the performance of the models. In 5-fold cross-validation, the dataset is divided into 5 subsets, and the model is trained and tested k times, each time using a different subset as the validation set and the remaining subsets for training, providing a thorough assessment of the model's performance. This method helps to mitigate overfitting and ensures that the model is not overly dependent on any particular subset of the data. 

Table \ref{tab:BC_K_fold} provides the performance metrics for the same machine learning models but evaluated using 5-fold cross-validation. In this method, the dataset is split into five equal parts, and the model is trained on four parts and tested on the remaining one. This process is repeated five times, and each part is used exactly once as a test set.

\begin{table}[ht]
\centering
\caption{Performance for different models: All features with 5-fold, binary classification.}
    \begin{adjustbox}{max width=1\textwidth}
    \setlength{\tabcolsep}{2pt} 
    \begin{tabular}{lccccccc}
    \toprule
    Model & Gradient Boosting & Catboost & LightGBM & SVM & Random Forest & MLP & EEGNet \\
    \midrule
    {ROC AUC} & 0.6292 & 0.5462 & 0.6892 & 0.5571 & 0.6257 & 0.5182 & 0.3995\\
    {PR AUC} & 0.8529 & 0.8255 & 0.8833 & 0.8184 & 0.8597 & 0.8318  & 0.7839\\
    \bottomrule
    \end{tabular}
    \end{adjustbox}
    \label{tab:BC_K_fold}
\end{table}

In this evaluation, LightGBM again outperforms other models with a ROC AUC of 0.6892 and a PR AUC of 0.8833. Gradient Boosting and Random Forest show comparable ROC AUC values of 0.6292 and 0.6257, respectively. Catboost and SVM exhibit lower performance, while MLP remains the lowest-performing model based on ROC AUC.

\subsubsection{Multi-class Classification with LOPO}
In addition to the binary classification, we also investigate the performance of models' prediction using multi-level severity following DAS21. {Supplementary Table \ref{tab:Multi_LOPO}} details the performance of the models on multi-class classification tasks using the LOPO cross-validation. The approach is even more challenging in a multi-class setting as the model must correctly classify multiple classes for each patient left out during testing.

LightGBM exhibits the best performance for multi-class classification with LOPO, achieving a Macro ROC AUC of 0.5678 in the One-vs-Rest approach and 0.4781 in the One-vs-One approach. However, all models show relatively low performance across all metrics, reflecting the difficulty of the multi-class classification task under LOPO validation. 

A critical observation is the drastic decline in F1-scores when transitioning from the 80:20 split to the LOPO cross-validation framework. While LightGBM achieved an impressive F1-score of 0.9391 in the 80:20 split, this metric plummeted to 0.1408 under LOPO, with other models showing a similarly poor performance (ranging from 0.1307 to 0.1698). This substantial disparity highlights a lack of inter-subject generalizability in the multi-class context; the models appear to master the idiosyncratic patterns of individuals within the training set but struggle to project these learned severity levels onto entirely unseen physiological profiles. The low F1-scores in {Supplementary Table \ref{tab:Multi_LOPO}} underscore that while the LDF and FS features are highly descriptive for specific subjects, the overlapping physiological signatures across different stress levels and different individuals remain a significant hurdle for robust, universal multi-class prediction.

\subsubsection{Multi-class Classification with 5-folds}

Similarly to the LOPO for multi-class classification, we also employ 5-fold for health issue investigation. {Supplementary Table \ref{tab:Multi-Kfold}} shows the performance metrics for multi-class classification using 5-fold cross-validation. This method helps mitigate the variance seen in LOPO by averaging the performance over multiple splits.

{Supplementary Table \ref{tab:Multi-Kfold}} shows that LightGBM continues to show the highest performance with a Macro ROC AUC of 0.5812 (One-vs-Rest) and 0.5057 (One-vs-One). Gradient Boosting and SVM also perform relatively well, but all models have lower performance metrics compared to the binary classification tasks, illustrating the increased complexity of multi-class classification. The results for the 5-fold cross-validation further illustrate the complexity of the multi-class task. Although this method typically provides a more stable estimate of model performance than LOPO by averaging across multiple data splits, the F1-scores for all models remain notably low, ranging from 0.1411 to 0.1736. The significant performance gap when compared to the 80:20 split suggests that accurately identifying specific stress levels is highly sensitive to the data partitioning strategy. This indicates that while the features are descriptive, there is a high degree of overlap in the physiological signatures between different severity groups, making robust universal classification a remaining challenge.

\subsection{Multimodal Sensor Features}
\label{sec.result.sensorfeatures}

\subsubsection{Binary Classification with LOPO}

Performance metrics for different machine learning models using the LOPO approach are summarized in {Supplementary Table \ref{tab:sensor_features_binary_LOPO}}. The LightGBM model achieved the highest ROC AUC score of 0.698, suggesting that it performed relatively better compared to using all features as illustrated in Table \ref{tab:BC_LOPO}. Gradient Boosting followed with an ROC AUC of 0.6265, indicating moderate discriminative ability. In terms of PR AUC, which measures the trade-off between precision and recall, LightGBM again stands out with a score of 0.9091, demonstrating its robustness in handling imbalanced classes. Other models, including Catboost, SVM, and Random Forest, showed lower ROC AUC and PR AUC scores.

\subsubsection{Binary Classification with 5-folds}
The performance metrics for the 5-fold cross-validation approach are detailed in {Supplementary Table \ref{tab:sensor_features_binary_kfold}}. Here, LightGBM also performed well, achieving an ROC AUC of 0.6601 and a PR AUC of 0.8839, highlighting its consistent performance across different validation techniques. Gradient Boosting followed with an ROC AUC of 0.6137 and a PR AUC of 0.8424, reinforcing its reliability as a robust model for this classification task. The Catboost model showed improved performance in the 5-fold scenario (ROC AUC of 0.5145) compared to LOPO, indicating that it might be better suited for general datasets rather than patient-specific variations. SVM and Random Forest had similar ROC AUC scores, around 0.5389 and 0.5607 respectively, but they showed adequate precision-recall trade-offs with PR AUC scores above 0.82.

\subsection{Top-10 Important Features}
\label{sec.result_top10}
Although, we have features extracted from wearable devices and personal information, using the top 10 important features for classification is a strategic approach aimed at enhancing model efficiency and interpretability. Utilizing the top 10 important features allows us to significantly reduce the time and energy required for data collection and processing, thereby saving valuable resources and expediting the overall analysis workflow. 

\subsubsection{Binary Classification with LOPO}
As shown in {Supplementary Table \ref{tab:top10_binary_LOPO}}, the models evaluated include Gradient Boosting, Catboost, LightGBM, SVM, Random Forest and MLP. The results indicate that LightGBM achieved the highest ROC AUC score of 0.7041, followed by Gradient Boosting with a score of 0.6699. Catboost, SVM, Random Forest and MLP showed moderate performance with ROC AUC scores of 0.5788, 0.578, 0.6232, and 0.5454, respectively. In addition, in terms of the Precision-Recall AUC, LightGBM also led with a score of 0.9087, highlighting its superior ability to handle class imbalances and correctly identify positive instances in this binary classification task.

\subsubsection{Binary Classification with 5-folds}
As shown in Table \ref{tab:top10_binary_Kfolds}, LightGBM consistently performed well, achieving an ROC AUC of 0.7168 and a PR AUC of 0.8852, underscoring its robustness and effectiveness in different cross-validation techniques. Gradient Boosting and Catboost also performed competitively with ROC AUC scores of 0.6594 and 0.6173, respectively, and PR AUC scores of 0.8723 and 0.8512. 

\begin{table}[ht]
\centering
\caption{Performance for different models: Top 10 features with 5-fold, binary classification.}
\begin{adjustbox}{max width=1\textwidth}
\setlength{\tabcolsep}{2pt} 
\begin{tabular}{lcccccc}
\toprule
Model & Gradient Boosting & Catboost & LightGBM & SVM & Random Forest \\
\midrule
{ROC AUC} & 0.6594 & 0.6173 & 0.7168 & 0.5692 & 0.6402 \\
{PR AUC} & 0.8723 & 0.8512 & 0.8852 & 0.841 & 0.8754 \\
\bottomrule
\end{tabular}
\end{adjustbox}
\label{tab:top10_binary_Kfolds}
\end{table}

\subsubsection{Multi-class Classification with LOPO}
We also conducted multi-class classification training using the LOPO method. As shown in {Supplementary Table \ref{tab:top10_multi_LOPO}}, the performance metrics indicate a notable variation among the machine learning models. LightGBM emerged as the top performer with a Macro ROC AUC score of 0.633 (One-vs-Rest) and 0.5244 (One-vs-One), demonstrating its capability to handle multiple classes effectively. Gradient Boosting and Catboost showed moderate performance with Macro ROC AUC scores around 0.4946 and 0.4463, respectively. However, the overall macro precision, recall, and F1-score for all models were relatively low, highlighting the complexity and challenge of multi-class classification tasks using LOPO.

\subsubsection{Multi-class Classification with 5-folds}
Finally, we conducted multi-class classification using the same models with 5-fold cross-validation. {Supplementary Table \ref{tab:top10_multi_kfold}} shows that LightGBM again led with a Macro ROC AUC score of 0.6412 (One-vs-Rest) and 0.5585 (One-vs-One), reinforcing its consistent performance across different evaluation methods. Gradient Boosting and Catboost also showed improved performance with Macro ROC AUC scores of 0.5418 and 0.5315, respectively.

When employing the top 10 features, the binary classification performance under the LOPO scheme shows slightly better performance in ROC AUC and PR AUC metrics across most models compared to using all features. For example, Gradient Boosting's ROC AUC increased from 0.6556 to 0.6699, while LightGBM's PR AUC slightly increased from 0.8998 to 0.9087. Similarly, in multi-class classification, the LOPO results show that models trained with the top 10 features generally have higher Macro ROC AUC and precision scores compared to those trained with all features. By focusing on the top ten important features, we can not only enhance model performance but also significantly reduce the time and energy required for data collection and processing, making the analysis more efficient and cost-effective.

\section{Discussions}
\label{sec.resultXAI}

In addition to interpretability and statistical validation, we further assessed the robustness of the model using repeated cross-validation. All primary experiments were conducted using the full feature set, including both demographic variables and wearable-derived physiological signals. Specifically, repeated 5-fold cross-validation with 10 repetitions (50 evaluations per model) was performed. The LightGBM model achieved the best performance with a mean ROC-AUC of $0.6800$ (95\% CI: $0.6461$--$0.7140$), outperforming Random Forest ($0.6636$, 95\% CI: $0.6357$--$0.6914$) and Gradient Boosting ($0.6629$, 95\% CI: $0.6368$--$0.6891$). Lower performance was observed for CatBoost ($0.5847$, 95\% CI: $0.5491$--$0.6203$) and SVM ($0.5652$, 95\% CI: $0.5375$--$0.5930$). The relatively narrow confidence intervals indicate consistent model performance across different data splits. These confidence intervals were obtained from binary classification experiments using repeated 5-fold cross-validation, and the corresponding bar chart with 95\% confidence intervals is provided in the Supplementary Material (Supplementary Figure \ref{fig:meanCI}).

While in the previous sections, we focused on the performance of various machine learning models for health issue prediction, it is crucial to understand the underlying factors that influence these predictions. This is where Explainable Artificial Intelligence (XAI) techniques come into play. XAI methods allow us to gain insight into the decision-making processes of machine learning models, offering valuable explanations for their predictions. As discussed in Sections \ref{sec.result.8020} to \ref{sec.result_top10}, LightGBM outperforms other machine learning models. Therefore, we used XAI techniques to interpret the highest-performing LightGBM model for stress detection. 

\begin{figure}[p]
    \centering
    \includegraphics[width=\linewidth]{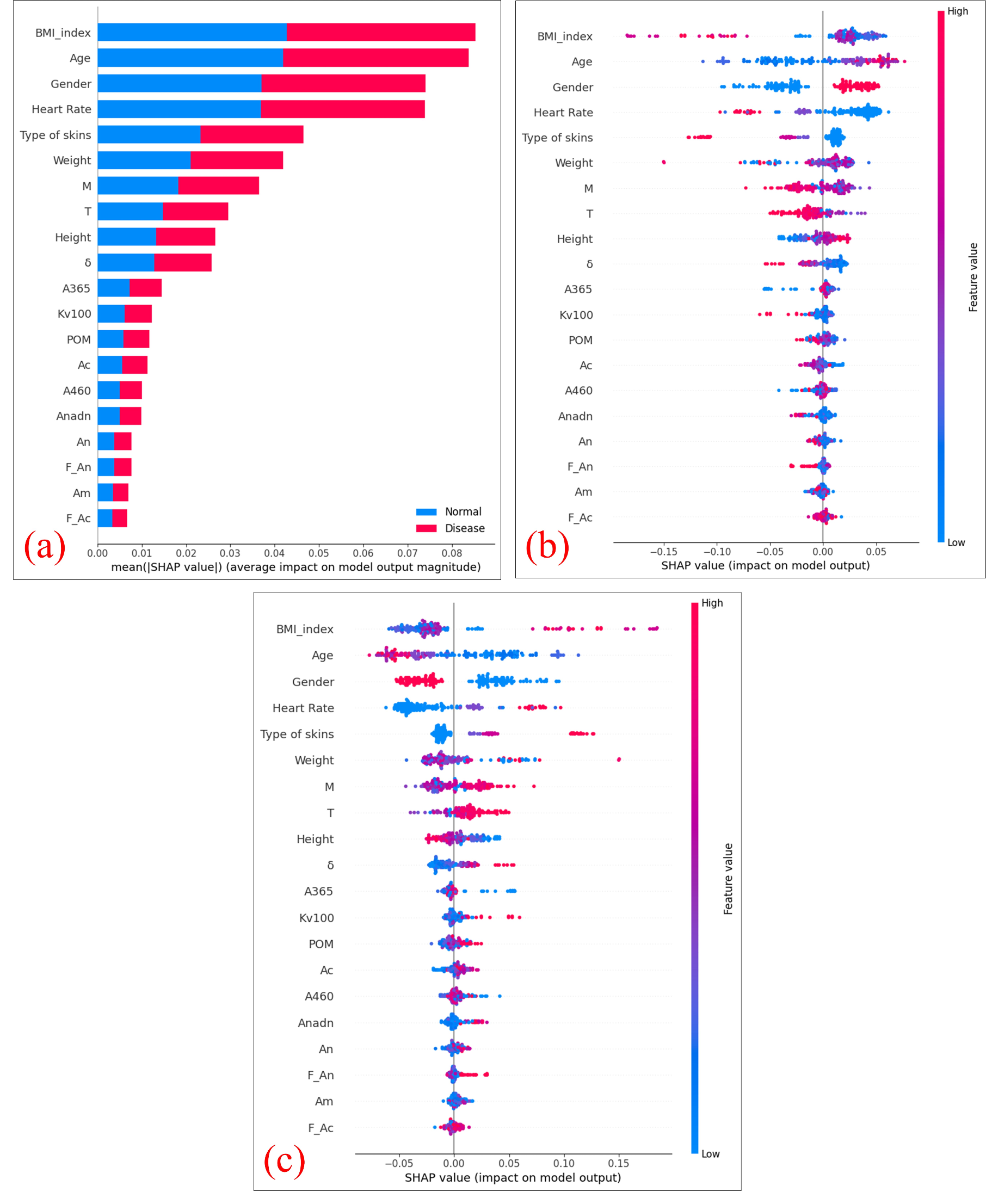}
    \caption{{\textbf{Explainable AI analysis of the LightGBM model using SHAP values.} (a) Global feature importance ranked by mean absolute SHAP values, 
    indicating the overall contribution of each feature to model output across 
    all samples. Features are ordered from most to least important (top to 
    bottom). (b) SHAP summary plot for the normal (well-being) group, showing 
    the distribution of feature contributions to predictions classified as 
    normal. (c) SHAP summary plot for the stress group, illustrating feature 
    contributions associated with stress-related predictions. In all panels, 
    each point represents one sample; color indicates class membership 
    (blue: normal; red: stress). The x-axis shows the SHAP value (impact on 
    model output); positive values increase the likelihood of the predicted 
    class. BMI: Body Mass Index; SHAP: SHapley Additive exPlanations; 
    LightGBM: Light Gradient Boosting Machine; XAI: Explainable Artificial 
    Intelligence.}}
    \label{fig.datadistribution}
\end{figure}

A plot of SHAP values is illustrated in Figure~\ref{fig.datadistribution}(a), in which the features are listed on the left-hand side of the plot, with the most important features at the top. Higher SHAP values indicate a greater impact on the model output. In addition, the blue color represents the normal class and the red color is the stress class. As we can see, the BMI index is the most important feature, followed by age, gender, and heart rate. As we can see, the contribution of each feature in each class is mostly equal.

To understand the distribution of each feature in each class, we plot the SHAP values of each class in Figure~\ref{fig.datadistribution}(b) and Figure~\ref{fig.datadistribution}(c). Similarly to Figure~\ref{fig.datadistribution}(c), the images show scatter plots of the effects of factors on the model output for each class. The x-axis represents the feature value and the y-axis represents the SHAP value. As observed, a low BMI index is associated with a lower likelihood of being classified as stressed by the model. Similarly to the BMI index, Figure~\ref{fig.datadistribution}(a) shows that high age and low heart rate are indicative of a lower likelihood of being stressed according to the model. In addition, women appear to be more stressed than men.

To validate these model-derived observations, we performed formal statistical testing on the relationship between demographic factors and DASS-21 scores:

\begin{itemize}
    \item \textbf{Gender Effects:} An independent samples t-test revealed statistically significant differences in total DASS-21 scores between genders (t = -2.078, p = 0.039). Female participants exhibited significantly higher mean scores (M = 1.04, SD = 1.47) compared to male participants (M = 0.49, SD = 1.08), confirming that women in our sample reported higher levels of psychological distress.
    \item \textbf{Age Effects:} One-way ANOVA demonstrated statistically significant differences across age groups (F = 5.717, p = 0.004). The middle-aged group (25--45 years) showed the highest mean total DASS-21 scores (M = 1.21, SD = 1.58), followed by the young group (<25 years, M = 0.45, SD = 0.93) and senior group (>45 years, M = 0.33, SD = 0.85). This suggests a non-linear relationship between age and mental health status, with peak vulnerability in middle adulthood rather than youth.
\end{itemize}

These statistical findings align with and validate the patterns identified through XAI analysis, providing robust evidence for the demographic risk factors associated with mental health status in our dataset.

To further examine the contribution of wearable-derived signals independently of demographic and clinical factors, an additional experiment was conducted using only features extracted from the wearable device. The results, summarized in Section \ref{sec:wearable_features_appendix} (Supplementary Information), indicate that while overall classification performance decreases, the model retains non-trivial discriminative ability. This suggests that LDF and FS signals capture relevant physiological information associated with mental health status, while demographic variables provide complementary contextual modulation.

\paragraph{Limitation} One limitation of this study is the inherent class imbalance in the collected dataset. Although stratified k-fold cross-validation was employed to maintain class proportions across folds and improve evaluation robustness, no explicit resampling or cost-sensitive learning strategies were applied during training. This decision was made to preserve the original physiological signal distribution and avoid introducing synthetic patterns that may not reflect true psychophysiological characteristics. Future work will systematically investigate imbalance-handling approaches, including class-weighted learning and controlled synthetic augmentation techniques, to further enhance minority-class sensitivity while maintaining physiological plausibility. While the dataset covers a geographically diverse cohort, the overall sample size remains moderate (n = 132). Therefore, caution is warranted when generalizing the findings to larger or more demographically balanced populations, and future multi-center studies are recommended to further validate model robustness.

Our study relies on DASS-21 questionnaire responses for mental health classification, which introduces several important limitations. First, DASS-21 scores reflect self-reported symptoms experienced during the past week only and do not constitute clinical diagnoses. Participants classified as ``normal'' may still experience subclinical psychological conditions, temporary emotional fluctuations, or underlying vulnerabilities not captured by the questionnaire. Second, self-report measures are susceptible to various biases, including social desirability bias (tendency to underreport symptoms), limited self-awareness, or cultural differences in expressing psychological distress. These factors may lead to misclassification of mental health status in our dataset.

\paragraph{Future Work}Future work will investigate deep learning frameworks designed to model the temporal and nonlinear dynamics of wearable-derived physiological signals. By better capturing complex interactions within microcirculatory and metabolic processes, these approaches may provide deeper insights into psychophysiological mechanisms underlying mental health conditions. Future studies will also explore sequence-aware and hybrid models, as well as hierarchical classification strategies, to improve multi-class discrimination and subject-independent generalization. Future work should incorporate clinical validation through structured diagnostic interviews, longitudinal assessment to capture temporal variability in mental health states, and integration of objective biomarkers alongside self-reported measures to improve classification accuracy and clinical validity. In addition, with larger and more balanced datasets, future studies will conduct subgroup-level fairness analyses with statistically reliable confidence intervals to further assess potential demographic bias in model predictions.

\section{Conclusion}
\label{sec.conclusion}
In this study, we introduce a novel approach to predict mental health by training predictive machine learning models for a non-invasive wearable device equipped with LDF/FS sensors. In addition, we establish a large and novel dataset of wearable devices containing physiological signals and the corresponding DASS-21 scores. To our knowledge, this is the largest and most generalized dataset ever collected for both the LDF and the FS studies.
In addition, we also evaluated various machine learning models for predicting DAS levels, prioritizing interpretable models to improve understanding of the relationship between wearable data and mental health. Finally, we employed explainable AI techniques to ensure transparency by identifying features that most influence predictions, providing insights that can help clinicians tailor treatment plans and improve patient outcomes.

Our findings show that: (1) The LightGBM model consistently outperforms others in both binary and multi-class stress level predictions, balancing accuracy and interpretability, making it suitable for practical applications. Using the top 10 important features, LightGBM achieved an ROC AUC of 0.7168 and a PR AUC of 0.8852; (2) Key physiological features like heart rate, BMI, and weight significantly influence stress predictions; (3) Younger individuals and those with a higher BMI or heart rate have a higher chance of experiencing stress; and (4) Females are more likely to be stressed than males.

The marked performance degradation under subject-independent validation highlights the significant inter-individual variability in physiological signals related to mental health states. This finding underscores the importance of rigorous evaluation protocols and suggests that future research should prioritize subject-invariant representation learning to improve generalization across individuals.

\onecolumn
\section*{Acknowledgement}
We would like to extend our sincere appreciation to the Human Ethics Committee of Aston University, Birmingham, UK, and Hai Duong Central College of Pharmacy, Vietnam, for their support and cooperation, including the waiver of informed consent. Their dedication to ethical standards greatly contributed to the success of this study.

The authors also acknowledge support from the British Council Women in STEM Fellowships program (grants No. 2324).


\section*{Data Availability}
The datasets generated and analysed during the current study include de-identified wearable physiological signals and questionnaire-based mental health assessments (DASS-21). All data and full analysis source code are publicly available at \url{https://github.com/leduckhai/Wearable_LDF-FS}. The data are stored in a structured format, including raw signals and extracted features, and can be accessed directly via the repository. All numerical source data underlying the main figures can be reproduced using the provided code. No restrictions apply to the use of these data. Additional information is available from the corresponding author upon reasonable request. The source data for statistical figures (Supplementary Figure \ref{fig.DataRepresentation}) are provided in Supplementary Data.

\section*{Code Availability}
All custom scripts used for data preprocessing, feature extraction, statistical analysis, and machine learning modelling are publicly available at: \url{https://github.com/leduckhai/Wearable_LDF-FS}. The repository contains the complete analysis pipeline required to reproduce the reported results and figures. Analyses were performed using Python (version 3.10) with commonly used scientific libraries including NumPy, pandas, scikit-learn, LightGBM, etc. Default model parameters were used unless otherwise specified in the manuscript. No restrictions apply to access or reuse of the code.

\section*{Competing Interests}

The authors declare no competing interests.

\section*{Author Contributions}

Minh Ngoc Nguyen, Khai Le-Duc, Tan-Hanh Pham, Trong Nhan Nguyen, and Truong-Son Hy performed the data analysis. Minh Ngoc Nguyen, Khai Le-Duc, Tan-Hanh Pham, Trong Nhan Nguyen, Viktor Dremin, and Truong-Son Hy wrote the original draft of the manuscript. Truong-Son Hy and Edik Rafailov jointly supervised this work. Truong-Son Hy served as the corresponding author. All other authors provided scientific advice. All authors reviewed, edited, and approved the final manuscript.

\onecolumn
\bibliographystyle{unsrtnat}
\bibliography{ms}

\clearpage 
\newpage

\begin{center}
{\Large A Wearable Device Dataset for Mental Health Assessment Using Laser Doppler Flowmetry and Fluorescence Spectroscopy Sensors}\\[0.5em]
{\Large \textbf{Supplementary Information}}
\rule{\textwidth}{0.4pt}
\end{center}

\vspace{1em}

\renewcommand{\thetable}{S\arabic{table}}
\setcounter{table}{0}

\appendix
\section*{Supplementary Notes}

\renewcommand{\tablename}{Supplementary Table}
\renewcommand{\thetable}{S\arabic{table}}
\renewcommand{\figurename}{Supplementary Figure}
\renewcommand{\thefigure}{S\arabic{figure}}

\section{Multimodal Wearable Optical Sensors}
\label{sec.full_literature_review}

Global health faces dual challenges from infectious diseases like COVID-19 and rising non-communicable diseases (NCDs). The World Health Organization (WHO) report highlights that the COVID-19 pandemic has caused significant disruptions in chronic disease services. Specifically, 53\% of countries reported disruptions in hypertension treatment, 49\% in diabetes care, 42\% in cancer treatment, and 31\% in cardiovascular emergency services. Additionally, over 50\% of countries postponed public screening programs for breast and cervical cancer due to the reassignment of healthcare staff to COVID-19 duties and the cancellation of planned treatments \cite{world2020impact, restrepo2008medication, yonel2018patients, mularczyk2022preventive}.
The use of remote monitoring devices without intervention is crucial to aid patients and healthcare professionals in timely classification and treatment. These devices can continuously monitor vital health indicators, detect abnormalities early, and alleviate the burden on the healthcare system.

Currently, there is a growing interest in wearable electronic diagnostic devices because daily monitoring of parameters promises a new quality of diagnosis. Recently, multimodal approaches have been actively developed, allowing clinicians to obtain \textit{in vivo} values of physiological and biochemical parameters, as well as to comprehensively assess the viability of the subcutaneous microcirculatory system. One of the first developments of wearable devices for estimating subcutaneous microcirculatory-tissue system parameters is the analyzer, developed by Professor E. Rafailov's research group at Aston University and produced by Aston Medical Technology Ltd. This device integrates a multimodal approach, specifically including two channels for LDF and FS\footnote{\url{https://amedtech.co.uk/}}.

This technology is renowned for its non-invasive measurement capabilities in living tissues. Studies have been conducted at various sites such as the wrist, ankle, thigh, and fingertips. It has many applications, including research on metabolic and vascular complications of diabetes, automatic cerebral vascular analysis, and monitoring cerebral circulation in both healthy individuals and those with disorders. It provides continuous, non-invasive monitoring during diagnostic, treatment, and post-treatment phases. Spectral characteristic changes have been observed in conditions such as malignant tumors, surgical trauma, increased arterial pressure, and many others. It is also used to assess the functional status of the cerebral vascular system in patients with acute and chronic cerebrovascular disorders.

\subsection{Specification of Wearable Devices}

A distinctive feature of the wearable devices under consideration is the absence of optical fibers in the design, which reduces common motion artifacts on the fibers. The wearable devices are placed on the skin for direct irradiation through a window on the underside of the device, recording backscattered or emitted (secondary) radiation from the biological tissue and transmitting measurement data to a PC via Bluetooth or Wi-Fi. The multimodal wearable device with two optical diagnostic channels uses an 850 nm VCSEL chip as the single-mode laser source with 0.8 mW power in the LDF channel. In the FS channel, a UV 365 nm LED is used, with a pulse power of 1.4 mW and an average power of 0.35 mW to excite endogenous fluorophores, particularly NADH. The amplitude of NADH fluorescence intensity (ANADH) is normalized to backscattered radiation to reduce the influence of varying blood filling in the biological tissue, which arises, among other reasons, from artifacts related to different pressures on the skin surface. Thus, sensors can monitor parameters such as perfusion, movement, skin temperature, and metabolic activity, providing crucial information for evaluating various physiological processes. Monte Carlo modeling has shown a penetration depth of up to 2 mm for the LDF channel (deep vessels) and 1 mm for the FS channel \cite{zharkikh2023sampling}. 

Common locations for wearable devices on the body\textquotesingle s biological tissue depend on the diagnostic task: these are usually symmetrical points on the right and left of the upper and lower limbs, areas with direct arterio-venous connections (hand or fingertip) and with predominant nutritional blood flow (forearm or lower leg), and on the forehead at the supraorbital artery regions.

\subsection{Wavelet Analysis Method}

The Morlet wavelet transform was used for the frequency analysis of registered signals 
\cite{tankanag2008application, stefanovska1999, dremin2019dynamic}. 
In short, the LDF signal was decomposed using a wavelet transform as:
\begin{equation}
W(s,\tau) = \frac{1}{\sqrt{s}} \int_{-\infty}^{\infty} x(t)\, \psi^{*} \left( \frac{t - \tau}{s} \right) dt
\end{equation}
where $x(t)$ is the target signal, $\tau$ is the local time index, $s$ is the scaling factor, and $^{*}$ denotes complex conjugation. 

The Morlet wavelet defined in the form
\begin{equation}
\psi(t) = e^{2\pi i t} e^{-t^{2}/\sigma^{2}}
\end{equation}
was used with the decay parameter $\sigma = 1$. 
This wavelet allows one to ensure sufficient time--frequency resolution and is well localized in the time domain.

\subsection{Methodological Validation and Testing Applications}

Before clinical deployment, the wearable LDF system underwent rigorous validation. Saha et al. \cite{saha2020wearable} investigated smoking effects using wearable LDF on fingertip and wrist in non-smokers and smokers. Non-smokers demonstrated significantly higher blood perfusion at both sites, with fingertip values approximately 350 perfusion units versus 250 units in smokers. Wavelet analysis revealed reduced endothelial, neurogenic, and myogenic oscillation amplitudes in smokers, indicating nicotine-induced endothelial dysfunction, increased neurogenic resistance, and elevated vascular smooth muscle tension. Pulse frequency was higher in smokers at approximately 1.2 Hz versus 1.0 Hz in non-smokers, reflecting nicotine's stimulatory effect on heart rate.

Fedorovich et al. \cite{BodyPosition} systematically investigated postural influences using six wearable LDF sensors on forehead, wrists, and shins in ten healthy males across supine, upright, and Trendelenburg positions. At the wrists, upright position decreased endothelial, neurogenic, and myogenic oscillation amplitudes, indicating increased vascular tone, while cardiac oscillations reduced, suggesting decreased arterial inflow. At the shins, upright position decreased perfusion and cardiac oscillations without altering tone-forming mechanisms, suggesting compensatory adaptations maintaining nutritive flow. The forehead demonstrated stable perfusion across all positions, reflecting robust cerebral autoregulation, though neurogenic and myogenic oscillation amplitudes increased during orthostasis, opposite to limb patterns. These findings emphasise the importance of standardising patient position during LDF measurements.

\subsection{Assessment of Microcirculatory Dysfunction in Diabetes}

A comprehensive study using four wearable LDF analysers positioned on fingers, toes, wrists, and shins was conducted in 26 patients with type 2 diabetes mellitus and 31 age-matched healthy controls \cite{zharkikh2024}. The results revealed multidirectional changes in perfusion between upper and lower extremities. Toe perfusion was significantly reduced compared to control, while wrist perfusion was increased. Nutritive blood flow decreased in the toes but increased in the fingers and wrists of diabetic patients. Wavelet analysis demonstrated reduced endothelial oscillation amplitudes in the wrists and shins, decreased neurogenic oscillations in the shins, and a diminished contribution of endothelial oscillations to total spectral power across multiple sites, indicating systemic endothelial dysfunction. Respiratory oscillations increased in the fingers and shins, while cardiac oscillations showed divergent patterns with increases in the shins but decreases in the fingers.

Another investigation focused specifically on diabetic foot microcirculation, examining eleven diabetic patients and twelve healthy adults across four foot regions: the first metatarsal head, fifth metatarsal head, heel, and dorsal foot \cite{Xing2024}. Diabetic patients showed significantly reduced mean blood flow in the neurogenic components at both metatarsal heads and in the cardiac component at the first metatarsal head. Sample entropy analysis revealed reduced signal complexity in diabetic patients across neurogenic and myogenic components at the first metatarsal head, the endothelial component at the fifth metatarsal head, and the myogenic component at the dorsal foot. These findings demonstrate differential involvement of hairy skin, represented by the dorsal foot, versus glabrous skin, represented by the plantar surface, in diabetic microangiopathy. Arteriovenous anastomosis dilation in plantar regions, resulting from sympathetic denervation, explains the preservation of total perfusion despite compromise of nutritive capillary flow.

A comparative analysis of ageing and diabetes effects on microcirculation was performed in 37 diabetic patients, 37 older healthy controls, and 58 younger healthy controls \cite{Zherebtsov2023}. Average perfusion increased with both age and diabetes, though no statistically significant difference emerged between diabetic patients and older controls. The energy of blood flow oscillations was reduced in diabetic patients across endothelial, neurogenic, myogenic, and respiratory frequency ranges, with significant differences between diabetic patients and older controls specifically in neurogenic and myogenic oscillations. These findings support the sequential model of diabetic microvascular dysfunction whereby neural control is affected earliest, followed by endothelial function, and finally smooth muscle responses. Logistic regression models achieved excellent classification performance, with accuracy of 0.92 for distinguishing younger controls from diabetic patients using cardiac and myogenic oscillation parameters.

The utility of wearable LDF for therapeutic monitoring was demonstrated in diabetic patients receiving five daily intravenous infusions of alpha-lipoic acid \cite{article2022}. Following treatment, patients showed decreased microcirculation index in both upper and lower extremities, decreased nutritive blood flow, and increased shunt index. Critically, post-treatment parameters approximated the values of the healthy control group. Myogenic oscillation amplitudes in the lower limbs decreased below control values, suggesting increased precapillary sphincter tone that may be protective against oedema and inflammation. This redistribution of blood flow from nutritive to shunt pathways with concurrent decrease in total perfusion may represent a positive therapeutic effect, protecting capillaries from pressure damage while maintaining tissue oxygenation through enhanced oxygen extraction efficiency.

\subsection{COVID-19 Microcirculatory Impairment}

Zharkikh et al. \cite{diagnostics13050920} conducted a two-phase investigation of COVID-19 effects using wearable LDF. The first phase followed one patient 10 days pre-infection and 26 days post-recovery. Following recovery, myogenic oscillations decreased in wrists, while respiratory and cardiac oscillations decreased in shins. Dynamic observations revealed cardiac fluctuations reduced immediately post-disease, while respiratory fluctuations changed during the first week after recovery.

The second phase compared twenty-three post-COVID patients undergoing rehabilitation with thirteen pre-pandemic controls. In post-COVID patients, shin perfusion was reduced, and nutritive blood flow was significantly decreased in both wrists and shins. Neurogenic, respiratory and cardiac oscillation amplitudes showed marked increases in wrists. The increased neurogenic oscillation amplitude indicates decreased neurogenic tone, leading to arteriolar dilation, arterio-venous anastomosis dilation, increased shunting away from capillaries, and consequent venular dilation. This pattern of increased oscillations with decreased nutritive flow describes post-COVID microvascular dysregulation that may underlie long COVID symptoms, including fatigue and cognitive dysfunction.

Summarizing all the given data, the presented wearable diagnostic devices with LDF and FS channels are a promising approach for evaluating the functional state of microcirculation under normal and pathological conditions. However, more detailed studies with larger patient cohorts and extended analysis of physiological conditions should be conducted for further clinical implementation.

One of the crucial tasks is to investigate the effects of various treatment protocols and lifestyle changes on microcirculatory and metabolic parameters using these wearable devices. Another important direction is developing machine learning algorithms for automated data analysis and interpretation, which could significantly enhance the diagnostic capabilities of wearable devices.

The development and application of wearable diagnostic devices with LDF and FS channels represent a significant advancement in medical diagnostics, offering non-invasive, real-time monitoring of the microcirculatory system and metabolic state. These devices hold great potential for improving patient care, particularly in managing chronic diseases and monitoring treatment efficacy. Our research aims to leverage this technology to build a dataset for stress detection. The volunteers in our work have diverse medical histories, including migraine, diabetes, STEMI, and hypertension. By exploring cutaneous blood microcirculation parameters using a non-invasive wearable device equipped with LDF and FS channels, we can gain valuable insights. To the best of our knowledge, our work pioneers in publishing a large LDF/FS wearable device dataset for mental health assessment.

\onecolumn
\section{Detailed Study Design and Dataset Description}
\label{sec.appendix_study_design_and_data}

\subsection{Clinical Definition and Data Collection}
\label{sec.clinical_definition}

The criteria of this study are described as follows:

\begin{itemize}
     \item This study focuses on volunteers aged 18 and above.
     \item Volunteers must not have any medical conditions related to dermatological diseases on both hands and middle fingers.
     \item  A total of 132 volunteers, aged 18 and above, of all genders, various occupations, and different ages, who are healthy and alert, were included.
    \item Feature groups are organized into four categories, summarized in Table~\ref{tab:features}.
   
\end{itemize}

Table \ref{tab:das21responses} shows the depression, anxiety and stress scale (DASS21) questionnaire responses which we gave to our volunteers.

\subsection{Data Analysis}

Our dataset focuses on understanding the factors influencing depression, anxiety, and stress (DAS) levels. To achieve this, we have collected and integrated data from three key sources: personal information, wearable sensor readings, and the DASS-21 questionnaire. This diverse data representation allows us to create a comprehensive picture of each individual's background, psychological state, and physiological responses. 
The dataset includes 132 participants recruited from 19 countries, providing geographic diversity while maintaining a controlled experimental protocol.

Similar to a medical record, the dataset includes essential demographic details for each participant. Our research involved a wide range of ages (18--94) with an average participant age of 40. We also ensured participant diversity by including a variety of races (Asian, White, African) and genders (55.2\% male, 44.8\% female), as shown in Fig.~\ref{fig.DataRepresentation}(a) and Fig.~\ref{fig.DataRepresentation}(b), respectively. 
Although participants were drawn from multiple countries, the ethnic distribution is predominantly Asian, which should be considered when interpreting model generalizability across broader populations.

In addition to the demographic information, we also investigate the effect of patients' routines and medical history, which is used to further understand potential contributors to DAS. As shown in Fig.~\ref{fig.DataRepresentation}(c), most of the participants do not smoke cigarettes, and the rate of people who used to smoke is small compared to non-smokers and current smokers. A similar observation is seen in the health issues attribute, in which the percentage of people who have problems with health issues is significantly smaller than normal people, as illustrated in Fig.~\ref{fig.DataRepresentation}(d).

Table \ref{tab.Wellbeing1} and Fig. \ref{fig.wellbeing-nonwellbeing1} indicate that non-wellbeing individuals exhibit higher perfusion parameters and amplitude variations compared to wellbing individuals. For the perfusion parameter M, wellbeing individuals have a mean value of 21.02 (95\% CI, 4.73 - 35.59), whereas non-wellbeing individuals have a significantly higher M value of 26.49 (95\% CI, 8.59 - 36.71), (p=0.016, Mann-Whitney U test). The endothelial, neural, muscle, respiratory, and cardiovascular amplitude variations all tend to be higher by more than 0.2, but the differences are not statistically significant.

As shown in Table \ref{tab:Wellbeing_Kv100} and Fig. \ref{fig.wellbeing-nonwellbeing2}, non-wellbeing individuals exhibit a statistically significant higher amplitude of perfusion parameter fluctuations ($\delta$) compared to their wellbeing counterparts, with values of 4.79 (95\% CI, 2.25 - 7.65) versus 3.64 (95\% CI, 0.98 - 6.93), (p=0.007, Mann-Whitney U test). Additionally, they have a significantly higher temperature at the measurement site, 33.21 (30.14 - 35.82) compared to 30.65 (95\% CI, 22.68 - 35.6), (p=0.01, Mann-Whitney U test). Moreover, the metabolic index (POM) at the measurement site is also significantly elevated in non-wellbeing individuals, with values of 11.42 (95\% CI, 1.95 - 24.42) versus 7.7 (95\% CI, 0.85 - 20.43), (p=0.01, Mann-Whitney U test).

\subsection{Statistical Analysis of Demographic Factors}
To validate demographic patterns in mental health status, we performed statistical significance testing on gender and age effects. Data were first aggregated by Patient\_ID to ensure each participant contributed a single observation, with total DASS-21 score computed as the sum of Depression, Anxiety, and Stress subscales.

For gender analysis, an independent samples t-test was conducted to compare total DASS-21 scores between male and female participants. For age analysis, participants were categorized into three groups: Young (<25 years), Middle (25--45
years), and Senior (>45 years), and one-way ANOVA was performed to assess differences across age groups. All statistical analyses were conducted using scipy.stats in Python, with significance level set at $\sigma = 0.05$.

\onecolumn
\section{Data Processing}
\label{sec.details_experimental_setup}
This Appendix Section details the machine learning models employed for predicting depression, anxiety, and stress (DAS) levels, along with the Explainable Artificial Intelligence (XAI) technique used to interpret their decision-making processes.

\subsection{Data Pre-processing Pipeline}
Prior to model training, a standardized preprocessing pipeline was applied to ensure data quality and consistency. Raw wearable and DASS-21 data were consolidated into a unified master table. Missing values and irrelevant features were removed, data types were verified, and categorical variables were appropriately encoded. Outliers were examined and treated to reduce noise.

For models sensitive to feature magnitude (SVM and MLP), feature standardization (zero mean, unit variance) was applied. All preprocessing steps were conducted strictly within each training fold during cross-validation to prevent data leakage, ensuring fair model comparison and reliable performance evaluation.

\subsection{Machine Learning Models}

To identify the most effective approach for predicting Depression, Anxiety, and stress (DAS) levels, we explored various machine learning algorithms. These algorithms leverage the collected wearable device data and DASS-21 questionnaire scores to estimate a patient's mental issues.

\textbf{Support Vector Machine} (SVM) is a well-established method that is known for its ability to effectively handle high-dimensional datasets, even with a relatively small sample size. SVMs aim to find a hyperplane in the feature space that maximizes the margin between different classes. In SVM, new data points are then classified based on which side of the hyperplane they fall on. SVMs are powerful for classification tasks with high-dimensional data, such as ours with potentially many features extracted from wearable sensors. They are effective even with limited data and offer good generalization capabilities. However, SVMs can be computationally expensive for very large datasets and may be less interpretable than other algorithms on this list.

\textbf{Random Forest Classifier} is an ensemble learning algorithm that combines the predictions of multiple, independently trained decision trees. Each tree is built using a random subset of features and data points, promoting diversity within the ensemble. The final prediction is made by majority vote or averaging the individual tree predictions. The algorithms are robust to overfitting due to their inherent diversity. They can handle various data types and perform well even with missing values. This approach is particularly useful for datasets with potential noise or inconsistencies.

\textbf{Gradient Boosting Classifier} is an algorithm that works by iteratively building an ensemble of weak decision trees. Each tree learns from the errors of the previous one, ultimately leading to a more robust and accurate model. Gradient boosting is known for its flexibility and ability to handle various data types, making it a strong contender for our analysis.

Building upon gradient boosting, \textbf{CatBoost} specifically addresses challenges in healthcare data. It incorporates advanced techniques for handling categorical features, such as one-hot encoding or custom loss functions, which can be problematic for traditional gradient boosting. Additionally, CatBoost prioritizes model interpretability by providing feature importance scores and visualizations of decision boundaries. CatBoost excels in scenarios with a high volume of categorical features, common in healthcare data. It offers improved interpretability compared to standard gradient boosting, allowing us to understand the factors influencing model predictions

\textbf{LightGBM} (Light Gradient Boosting Machine) is a highly efficient implementation of the gradient boosting algorithm specifically designed for speed and performance. It utilizes techniques similar to gradient-based one-side sampling and feature importance sampling to focus on informative data points and reduce computational costs. LightGBM offers exceptional speed and accuracy, making it a compelling choice for large datasets. It is particularly efficient for memory usage, allowing for training in resource-constrained environments. LightGBM excels at handling large and complex datasets, making it suitable for our analysis where we have a high volume of data points from wearable devices. 

In addition to the machine learning algorithms above, we also implemented a \textbf{Multi-layer Perceptron} (MLP) for health issue prediction. Unlike simpler models, MLPs excel at identifying complex, non-linear relationships within the data. This capability could be particularly valuable for uncovering subtle patterns between physiological signals and mental health states. The MLP neural network has two hidden layers and a final output layer with a unit. The network employs layer normalization, ReLU activation for hidden layers, and dropout to prevent overfitting. Finally, a sigmoid or a softmax activation function is applied to the output layer to transform the final values to the probability of classes.

\textbf{EEGNet} is a compact convolutional neural network specifically designed for EEG-based classification tasks. It efficiently captures both spatial and temporal patterns in multi-channel EEG signals using depthwise and separable convolutions. In our study, EEGNet is adapted to process wearable-derived physiological signals, enabling the model to learn subtle spatio-temporal dependencies related to Depression, Anxiety, and Stress (DAS) levels. Its lightweight architecture allows fast training and evaluation even on relatively small datasets.

By evaluating the performance of these diverse algorithms, we aim to identify the one that best predicts DAS levels in the context of our specific dataset and research goals.

\subsection{Explainable AI}

In healthcare applications, ensuring trustworthy AI requires models to be not only accurate but also interpretable. Understanding the reasoning behind a model's predictions for DAS levels is crucial for building trust and confidence in its outputs. This empowers healthcare professionals and researchers to make informed decisions based on the predicted DAS levels and the underlying factors influencing those predictions. In this study, we leverage SHAP (Shapley Additive Explanations) to achieve interpretability and gain insights into the model's decision-making process for DAS prediction \cite{lundberg2017unified}.

SHAP, a powerful approach for achieving interpretability, assigns an attribution value (SHAP value) to each feature for a given DAS prediction. This value represents the contribution of that specific feature (e.g., a specific physiological sensor reading or a DASS-21 questionnaire response) to the model's final prediction. High positive SHAP values indicate that the feature has a strong positive influence on the predicted DAS level (potentially indicating a higher likelihood of depression, anxiety, or stress). Conversely, low negative SHAP values signify a negative influence (potentially indicating a lower likelihood). By analyzing the SHAP values for each feature, we can gain insights into the relative importance of various factors shaping the model's predictions about a user's mental state.

This interpretability allows us to answer several key questions:

\begin{itemize}
    \item \textbf{Identification of key physiological and psychological indicators:} What are the features from wearable sensor data and questionnaire scores of a patient that have the most significant influence on the model's predictions? Understanding these can pinpoint crucial physiological and psychological indicators associated with DAS levels. This knowledge can inform the development of targeted interventions and preventative measures for mental health.
    \item \textbf{Validation of model fairness and mitigation of bias:} Are the model's predictions fair across different demographics (age, gender, etc.)? Examining SHAP values across these groups helps ensure that the model is not unfairly biased towards certain populations. This is crucial in healthcare applications where fairness and unbiased decision-making are paramount.
    \item \textbf{Enhanced model transparency and trust:} How does the model arrive at its predictions? By explaining the rationale behind the model's predictions through SHAP values, we can foster trust and confidence in its use among healthcare professionals and researchers. This transparency is essential for the adoption and responsible use of AI in mental health assessments.
    
\end{itemize}

\onecolumn
\section{Training and Evaluation Metrics}

\subsection{Case Study}
\label{sec.casestudy}
Our study employs two primary approaches to train machine learning models for predicting DAS levels: binary classification and multi-class classification, as shown in Table \ref{table:all_features} and Table \ref{table:sensor_important_features}. Both approaches leverage data from the DASS-21 questionnaire alongside potentially other features from the collected dataset. In addition, we consider three cases to investigate the models' performances: Using all collected features as shown in Table \ref{table:all_features}, using only features extracted from wearable devices, and using top-10 important features as illustrated in Table \ref{table:sensor_important_features}.

For binary classification, this approach simplifies the prediction task by transforming the DAS levels into a binary classification problem. In particular, we utilize the information from the DASS-21 questionnaire as labels, focusing on the mental health state of the participants. We categorize participants into two classes based on their DASS-21 scores:
\begin{itemize}
    \item \textbf{Normal:} This class comprises participants who score within the normal range for depression, anxiety, and stress according to established DASS-21 scoring guidelines.
    \item \textbf{Abnormal:} This class encompasses participants whose DASS-21 scores indicate potential symptoms of depression, anxiety, or stress.
\end{itemize}

By converting the problem into a binary classification task, we can train models to distinguish between individuals with normal mental health states and those with potential mental health concerns based on their DASS-21 responses and potentially other features from the dataset.

For multi-class classification, this approach aims for a more granular prediction by treating DAS levels as a multi-class classification problem. Instead of collapsing mental health states into two categories, we define multiple classes based on the established DASS-21 scoring ranges: Normal, stress, stress Anxiety, and stress anxiety depression.

To provide a clearer overview of the dataset composition and the experimental setup used in this case study, we summarize the dataset structure and validation strategy in Table 23. In addition, the main hyperparameters used for the machine learning and deep learning models are reported in Table 24 to ensure experimental reproducibility.

\subsection{Data Split}

\subsubsection{Random 80:20 Split}
In machine learning, dividing the dataset into training and testing subsets is essential for reliable model evaluation. A commonly adopted strategy is the random 80:20 split, where 80\% of the data are used for model training and the remaining 20\% are reserved for performance testing. This approach provides a practical balance between sufficient training data and an independent evaluation set.

To prevent data leakage and ensure robust generalization assessment, the 80:20 split in this study was performed at the subject level rather than at the sample level. Specifically, all samples from a given individual were assigned exclusively to either the training set or the testing set.

\begin{python}
from sklearn.model_selection import train_test_split

# Assuming `X` is the feature matrix and `y` is the target vector
X_train, X_test, y_train, y_test = train_test_split(X, y, test_size=0.2, random_state=42)

\end{python}

\subsubsection{Leave-one-patient-out (LOPO)}
Predictive models for diagnosis or treatment outcomes must generalize well across different patients. 
Leave-one-patient-out (LOPO) \cite{hastie2009elements} ensures that the model is evaluated on its ability to perform on new patients not seen during training.
For each iteration, all data points from one patient are used as the test set, while the data from all other patients are used to train the model. 
This process is repeated for each individual in the dataset, ensuring that every patient's data is used as a test set exactly once.

\begin{python}
def leave_one_person_out(patient_ids):
  loo = dict()
  for patient_id in set(patient_ids):
    loo[patient_id] = None

  for patient_id in loo:
    train_indices, test_indices = [], []

    for i in range(len(patient_ids)):
      if patient_ids[i] == patient_id:
        test_indices.append(i)
      else:
        train_indices.append(i)

    loo[patient_id] = [train_indices, test_indices]

  return loo

patient_ids = list(df_rawdata['Patient_ID'])
loo = leave_one_person_out(patient_ids)

for patient_id in loo:
  # Leave one out
  train_indices = np.array(loo[patient_id][0])
  test_indices = np.array(loo[patient_id][1])

  X_train = X[train_indices]
  X_test = X[test_indices]
  y_train = y[train_indices]
  y_test = y[test_indices]
  
\end{python}

\subsubsection{K-folds}
K-fold cross-validation is a robust technique for training machine learning models in healthcare, ensuring reliable and unbiased performance evaluation \cite{hastie2009elements}. 
K-fold cross-validation mitigates the risk of overfitting, maximizes data utilization, and enhances the model's generalizability, which is crucial for accurately predicting patient outcomes and making informed medical decisions.
It involves randomly partitioning the dataset into k equally sized folds, where the model is trained on k-1 folds and validated on the remaining fold. 
This process is repeated k times, with each fold serving as the validation set once, and the performance metrics are averaged to provide a comprehensive assessment.
Note that for each fold there is no patient overlapping.
In the scope of this work, we used 5-fold cross-validation and 10 seeds for training, which means 5-fold was trained for 10 times to avoid varying results.

\begin{python}
def k_fold(patient_ids, number_of_folds = 5):
  patient_id_dict = dict()
  for patient_id in set(patient_ids):
    patient_id_dict[patient_id] = []

  for i in range(len(patient_ids)):
    patient_id = patient_ids[i]
    patient_id_dict[patient_id].append(i)

  kfold_dict = dict()
  i, k = 0, len(patient_id_dict.keys())//number_of_folds
  queue = list(patient_id_dict.keys())

  while queue:
    random.shuffle(queue)
    patient_id_test = [j for j in queue[0:k] ]
    patient_id_train = [j for j in set(patient_id_dict.keys())-set(patient_id_test) ]

    train_indices, test_indices = [], []
    for patient_id in patient_id_train:
      train_indices += patient_id_dict[patient_id]
    for patient_id in patient_id_test:
      test_indices += patient_id_dict[patient_id]
    kfold_dict[i] = [train_indices, test_indices]

    queue = queue[k:]
    i += 1

  return kfold_dict

patient_ids = list(df_rawdata['Patient_ID'])

for seed in range(10):
  kfold = k_fold(patient_ids, number_of_folds=5)
  for k in kfold:
    # K-fold
    train_indices = np.array(kfold[k][0])
    test_indices = np.array(kfold[k][1])

    X_train = X[train_indices]
    X_test = X[test_indices]
    y_train = y[train_indices]
    y_test = y[test_indices]

\end{python}
\subsection{Evaluation Metrics}
\label{sec.evaluation}
To assess the performance of the machine learning models for predicting Depression, Anxiety, and stress (DAS) levels, we employ two key evaluation metrics: Receiver Operating Characteristic (ROC) AUC (Area Under the Curve) and Precision-Recall (PR) AUC. These metrics provide a comprehensive assessment of the model's discriminative ability and its performance in handling class imbalances.

The ROC AUC metric measures the ability of the model to distinguish between classes. It plots the True Positive Rate (TPR) against the False Positive Rate (FPR) at various threshold settings. The AUC value ranges from 0 to 1, where a value of 1 indicates perfect classification and a value of 0.5 suggests performance no better than random guessing. A higher ROC AUC indicates that the model is better at differentiating between the positive and negative classes.

In our study, we calculate the ROC AUC to assess how well our model can predict DAS levels across the entire range of possible thresholds. This metric is particularly useful in healthcare applications, where the cost of false positives and false negatives can vary, and understanding the trade-offs between sensitivity and specificity is crucial.

While ROC AUC provides a good overall measure of performance, it can be insensitive to class imbalance.  In our case, the prevalence of depression, anxiety, and stress might be lower than the prevalence of healthy individuals.  The PR curve addresses this by plotting Precision against Recall.

\onecolumn
\section{ROC Plots}
This section shows the ROC AUC and ROC PR plots for all binary classification results above. The discrepancy between ROC AUC and PR AUC highlights the impact of class imbalance in mental health classification. While ROC AUC reflects overall separability, PR AUC provides a more sensitive measure of minority-class detection, which is particularly relevant in mental health assessment where pathological states are less frequent.

\subsection{All Features with 80:20 Split (Binary Classification)}
Figures \ref{fig:AUCplot-standard_classification-binary_gradientboosting}, \ref{fig:AUCplot-standard_classification-binary_catboost}, \ref{fig:AUCplot-standard_classification-binary_lightgbm}, \ref{fig:AUCplot-standard_classification-binary_svm}, \ref{fig:AUCplot-standard_classification-binary_randomforest}, and \ref{fig:AUCplot-standard_classification-binary_mlp} show the ROC AUC and ROC PR plots for all features binary classification models using Gradient Boosting, Catboost, LightGBM, SVM, Random Forest, and MLP respectively, with a random 80:20 split.

\subsection{All Features with Cross-Validation}

\subsubsection{Binary Classification: LOPO}
Figures \ref{fig:AUCplot-allfeatures_binary_classification-gradientboosting_lopo}, \ref{fig:AUCplot-allfeatures_binary_classification-catboost_lopo}, \ref{fig:AUCplot-allfeatures_binary_classification-lightGBM_lopo}, \ref{fig:AUCplot-allfeatures_binary_classification-svm_lopo}, \ref{fig:AUCplot-allfeatures_binary_classification-randomforest_lopo}, and \ref{fig:AUCplot-allfeatures_binary_classification-mlp_lopo} show the ROC AUC and ROC PR plots for all features binary classification models using Gradient Boosting, Catboost, LightGBM, SVM, Random Forest, and MLP respectively, with a LOPO split.

\subsubsection{Binary Classification: K-folds}
Figures \ref{fig:AUCplot-allfeatures_binary_classification-gradientboosting_5fold}, \ref{fig:AUCplot-allfeatures_binary_classification-catboost_5fold}, \ref{fig:AUCplot-allfeatures_binary_classification-lightGBM_5fold}, \ref{fig:AUCplot-allfeatures_binary_classification-svm_5fold}, \ref{fig:AUCplot-allfeatures_binary_classification-randomforest_5fold}, and \ref{fig:AUCplot-allfeatures_binary_classification-mlp_5fold} show the ROC AUC and ROC PR plots for all features binary classification models using Gradient Boosting, Catboost, LightGBM, SVM, Random Forest, and MLP respectively, with a K-folds split.

\subsection{Using Multimodal Sensor Features Classification}

\subsubsection{Binary Classification: LOPO}
Figures \ref{fig:AUCplot-sensorfeatures_binary_classification-gradientboosting_lopo}, \ref{fig:AUCplot-sensorfeatures_binary_classification-catboost_lopo}, \ref{fig:AUCplot-sensorfeatures_binary_classification-lightgbm_lopo}, \ref{fig:AUCplot-sensorfeatures_binary_classification-svm_lopo}, \ref{fig:AUCplot-sensorfeatures_binary_classification-randomforest_lopo}, and \ref{fig:AUCplot-sensorfeatures_binary_classification-mlp_lopo} show the ROC AUC and ROC PR plots for binary classification models of multimodal sensor features using Gradient Boosting, Catboost, LightGBM, SVM, Random Forest, and MLP respectively, with a split LOPO.

\subsubsection{Binary Classification: K-folds}
Figures \ref{fig:AUCplot-sensorfeatures_binary_classification-gradientboosting_kfold}, \ref{fig:AUCplot-sensorfeatures_binary_classification-catboost_kfold}, \ref{fig:AUCplot-sensorfeatures_binary_classification-lightgbm_kfold}, \ref{fig:AUCplot-sensorfeatures_binary_classification-svm_kfold}, \ref{fig:AUCplot-sensorfeatures_binary_classification-randomforest_kfold}, and \ref{fig:AUCplot-sensorfeatures_binary_classification-mlp_kfold} show the ROC AUC and ROC PR plots for multimodal sensor features binary classification models using Gradient Boosting, Catboost, LightGBM, SVM, Random Forest, and MLP respectively, with a K-folds split.

\subsubsection{Binary Classification: Top 10 Features with K-folds}
Figures \ref{fig:AUCplot-top10features_binary_classification-gradientboosting_5fold}, \ref{fig:AUCplot-top10features_binary_classification-catboost_5fold}, \ref{fig:AUCplot-top10features_binary_classification-lightgbm_5fold}, \ref{fig:AUCplot-top10features_binary_classification-svm_5fold}, and \ref{fig:AUCplot-top10features_binary_classification-randomforest_5fold} show the ROC AUC and ROC PR plots for top 10 features binary classification models using Gradient Boosting, Catboost, LightGBM, SVM, Random Forest, and MLP respectively, with a K-folds split.

\subsection{Using Wearable Features Classification}
\label{sec:wearable_features_appendix}

To isolate the contribution of wearable-derived signals, we excluded all demographic and clinical variables and trained models using only LDF and FS features. Binary classification was evaluated using both LOPO and k-fold cross-validation. Despite reduced performance compared to the full-feature model, wearable-only models retained measurable discriminative ability. Feature importance further confirmed physiologically meaningful contributions from vascular and metabolic parameters. These results support the independent relevance of wearable physiological signals for mental health assessment.

\subsubsection{Binary Classification: LOPO}

Table~\ref{tab:wearable_features_binary_LOPO} presents the performance of different machine learning models using wearable-derived features under the LOPO validation scheme for binary classification. Overall, the classification performance is moderate, with ROC AUC values ranging from 0.4730 to 0.5829. Among the evaluated models, the MLP achieved the highest ROC AUC (0.5829), followed closely by CatBoost (0.5667). In contrast, PR AUC values are relatively higher across models, with Random Forest (0.8173), CatBoost (0.8171), and LightGBM (0.8019) showing comparatively better performance. The modest performance observed under LOPO further highlights the challenges of generalizing wearable-derived physiological features across individuals, reinforcing the importance of incorporating complementary contextual or demographic information.

\subsubsection{Binary Classification: K-fold}

Table~\ref{tab:wearable_features_binary_kFold} presents the performance of wearable-only models under k-fold cross-validation. Compared to the LOPO setting, ROC AUC values are slightly improved (0.4973--0.5419), with MLP achieving the highest ROC AUC (0.5419), indicating better internal validation performance. PR AUC values remain consistently high across models (0.7969--0.8201), with Random Forest showing the best precision--recall trade-off (0.8201). These results suggest that wearable-derived features provide moderate discriminative capability, with more stable performance under k-fold validation than strict inter-participant LOPO evaluation.

\subsubsection{Binary Classification: Feature Importance }

Table~\ref{tab:merged_featuresbinary_prediction} indicates that the most important features across Gradient Boosting, Random Forest, and LightGBM are primarily related to microcirculatory perfusion and metabolic fluorescence signals. The microcirculation index (M) consistently ranks highest, highlighting the central role of tissue perfusion dynamics. Metabolic-related parameters such as A460 (NADH fluorescence amplitude) and POM (oxidative metabolism index) also contribute substantially, suggesting a link between cellular metabolic activity and psychophysiological status. The recurrent presence of vascular variability and oscillatory features further supports the relevance of autonomic and endothelial regulation mechanisms in mental health classification. These findings are consistent with the XAI-based interpretation presented in Health Issue Explanation Section, where similar microcirculatory and metabolic features were identified as key contributors to model decision-making.

\renewcommand{\thefigure}{S\arabic{figure}}
\setcounter{figure}{0}

\onecolumn
\section*{Supplementary Figures}

\begin{figure}[h]
    \centering
    \includegraphics[width=1\linewidth]{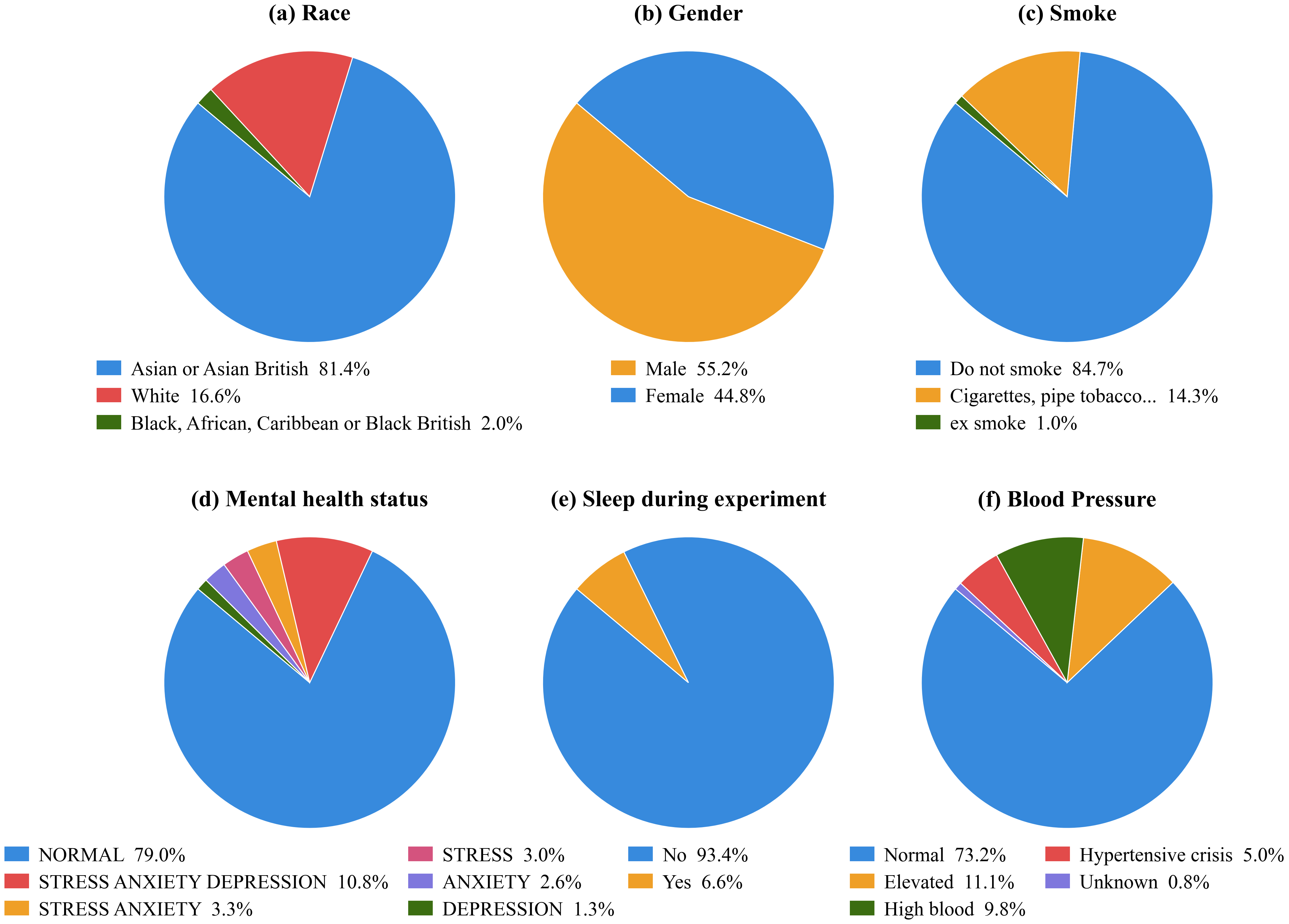} \\
    \caption{\textbf{Demographic and clinical characteristics of the study cohort.}
Pie charts showing the distribution of (a) racial background, (b) gender, 
(c) smoking status, (d) mental health status classified by DASS-21, 
(e) sleep quality during the experiment, and (f) blood pressure status.}
    \label{fig.DataRepresentation}
\end{figure}

\begin{figure}[h]
    \centering
    \textbf{Microvascular perfusion and oscillatory amplitudes in wellbeing and non-wellbeing groups}
    \includegraphics[width=1\linewidth]{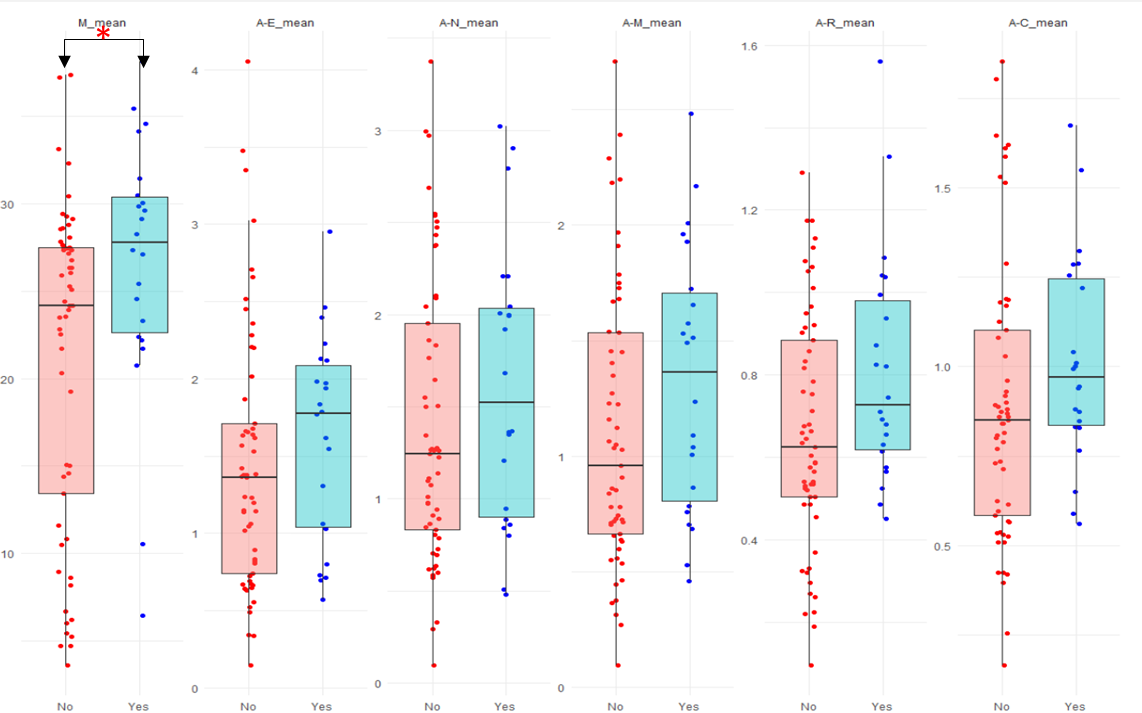} \\
    \caption{Blood perfusion (M*) with standard deviation and Maximum amplitude with a standard deviation of the endothelial (A-E), neurogenic (A-N), myogenic (A-M), breath (A-R) and pulse (A-C) mechanism for Wellbeing vs Non-wellbeing, *, p$<$0.01, Mann-Whitney U test.}
    \label{fig.wellbeing-nonwellbeing1}
\end{figure}

\begin{figure}
    \centering
    \textbf{Physiological and metabolic parameters in wellbeing and non-wellbeing groups}
    \includegraphics[width=1\linewidth]{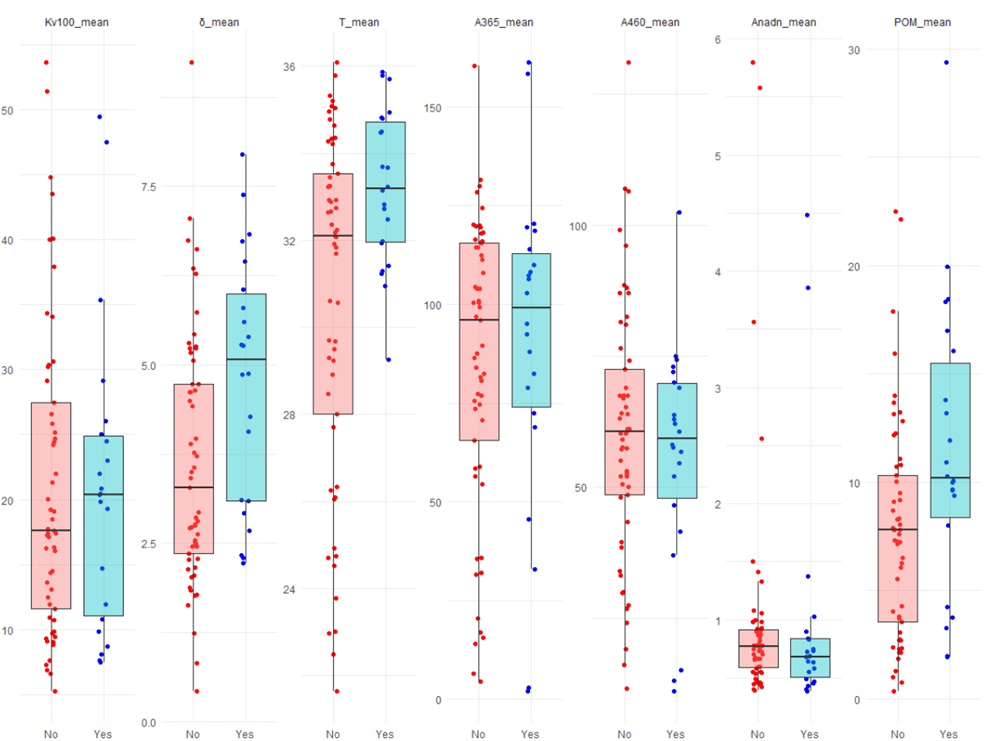}
    \caption{The parameters with standard deviation for Wellbeing vs Non-wellbeing: Kv100, $\delta$*, T*, A365, A460, Anadn, POM*, F-E; F-N; F-M; F-R; F-C, *, p$<$0.01, Mann-Whitney U test.}
    \label{fig.wellbeing-nonwellbeing2}
\end{figure}

\begin{figure}[h]
    \centering
    \includegraphics[width=.7\textwidth]{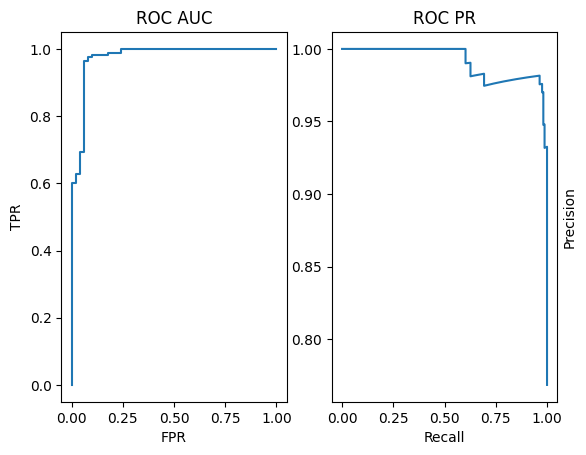}\hfill
    \caption{All features with 80:20 split, binary classification: Gradient Boosting}
    \label{fig:AUCplot-standard_classification-binary_gradientboosting}
\end{figure}

\begin{figure}[h]
    \centering
    \includegraphics[width=.7\textwidth]{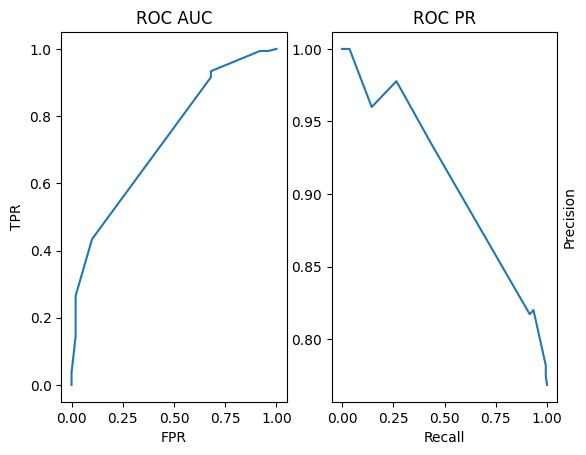}\hfill
    \caption{All features with 80:20 split, binary classification: Catboost}
    \label{fig:AUCplot-standard_classification-binary_catboost}
\end{figure}

\begin{figure}[h]
    \centering
    \includegraphics[width=.7\textwidth]{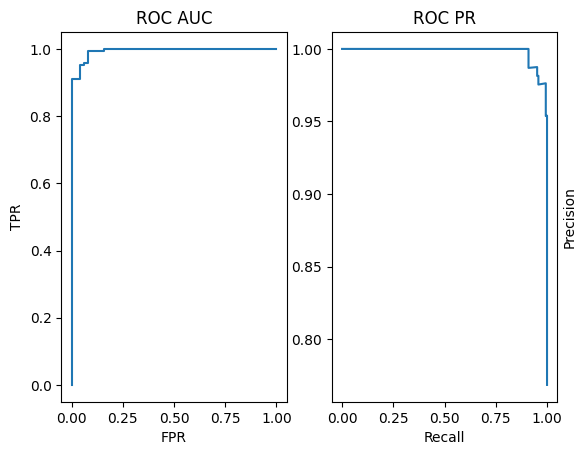}\hfill
    \caption{All features with 80:20 split, binary classification: LightGBM}
    \label{fig:AUCplot-standard_classification-binary_lightgbm}
\end{figure}

\begin{figure}[h]
    \centering
    \includegraphics[width=.7\textwidth]{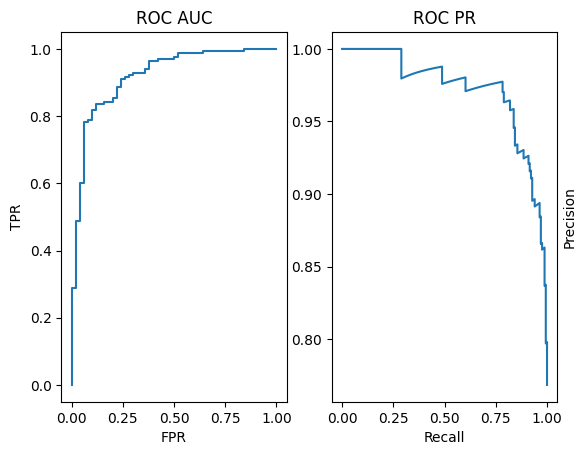}\hfill
    \caption{All features with 80:20 split, binary classification: SVM}
    \label{fig:AUCplot-standard_classification-binary_svm}
\end{figure}

\begin{figure}[h]
    \centering
    \includegraphics[width=.7\textwidth]{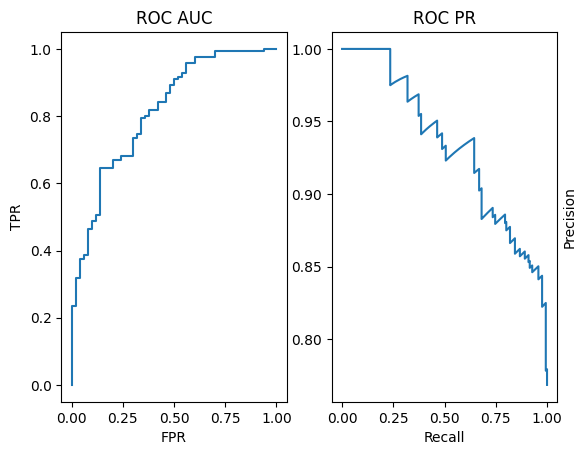}\hfill
    \caption{All features with 80:20 split, binary classification: Random forest}
    \label{fig:AUCplot-standard_classification-binary_randomforest}
\end{figure}

\begin{figure}[h]
    \centering
    \includegraphics[width=.7\textwidth]{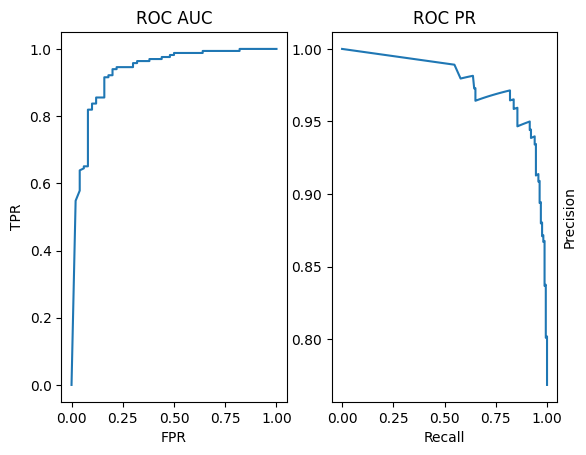}\hfill
    \caption{All features with 80:20 split, binary classification: MLP}
    \label{fig:AUCplot-standard_classification-binary_mlp}
\end{figure}

\begin{figure}[h]
    \centering
    \includegraphics[width=.7\textwidth]{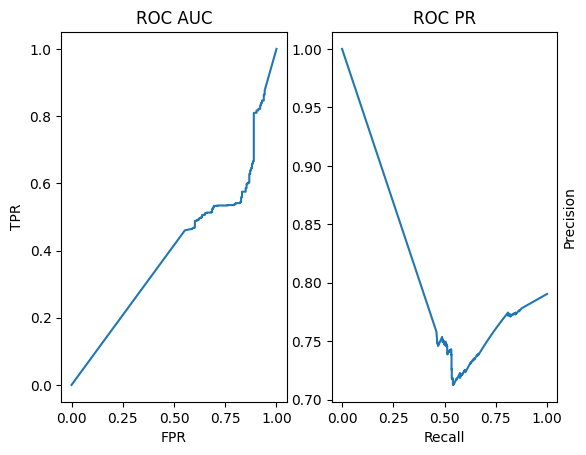}\hfill
    \caption{All features with 80:20 split, binary classification: EEGNet}
    \label{fig:AUCplot-standard_classification-binary_eeg}
\end{figure}

\begin{figure}[h]
    \centering
    \includegraphics[width=.7\textwidth]{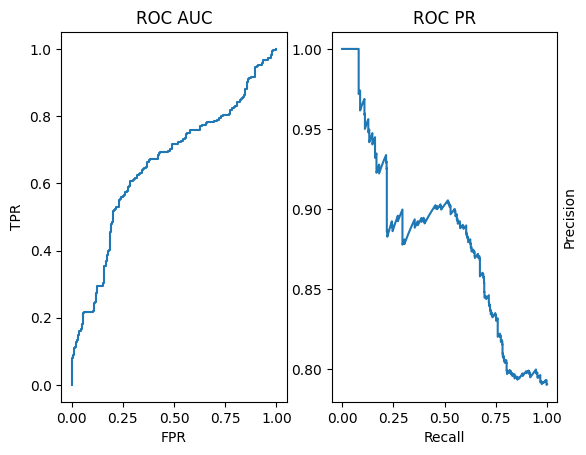}\hfill
    \caption{All features with LOPO, binary classification: Gradient Boosting}
    \label{fig:AUCplot-allfeatures_binary_classification-gradientboosting_lopo}
\end{figure}

\begin{figure}[h]
    \centering
    \includegraphics[width=.7\textwidth]{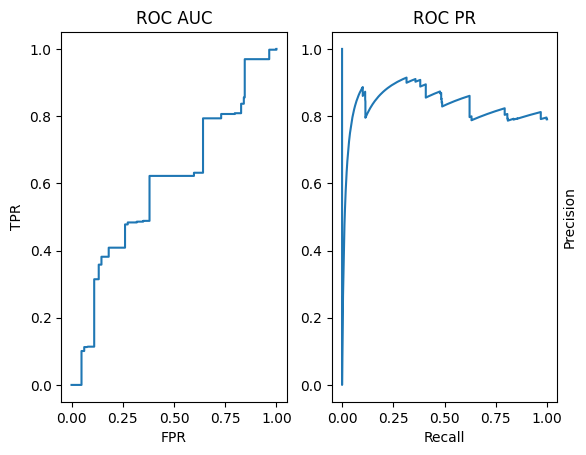}\hfill
    \caption{All features with LOPO, binary classification: Catboost}
    \label{fig:AUCplot-allfeatures_binary_classification-catboost_lopo}
\end{figure}

\begin{figure}[h]
    \centering
    \includegraphics[width=.7\textwidth]{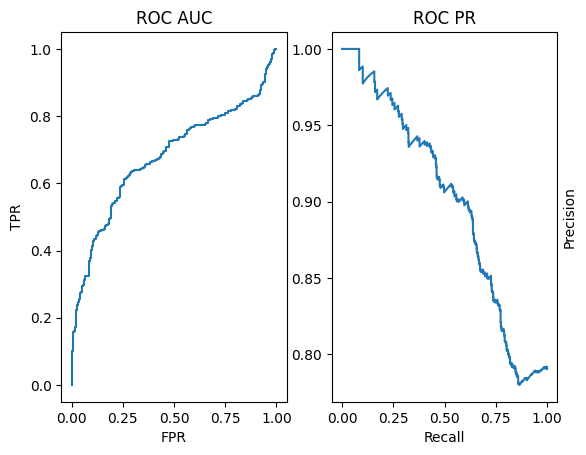}\hfill
    \caption{All features with LOPO, binary classification: LightGBM}
    \label{fig:AUCplot-allfeatures_binary_classification-lightGBM_lopo}
\end{figure}

\begin{figure}[h]
    \centering
    \includegraphics[width=.7\textwidth]{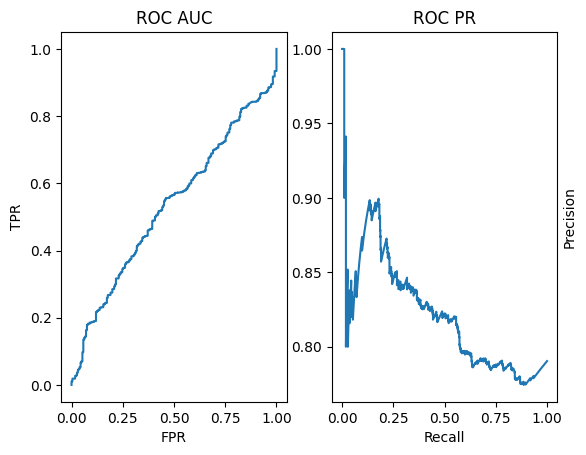}\hfill
    \caption{All features with LOPO, binary classification: SVM}
    \label{fig:AUCplot-allfeatures_binary_classification-svm_lopo}
\end{figure}

\begin{figure}[h]
    \centering
    \includegraphics[width=.7\textwidth]{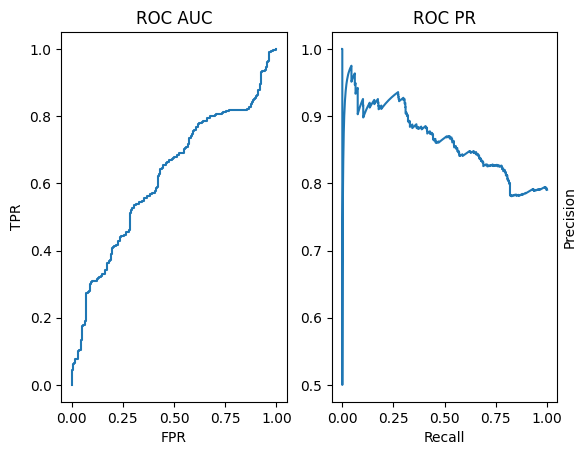}\hfill
    \caption{All features with LOPO, binary classification: Random Forest}
    \label{fig:AUCplot-allfeatures_binary_classification-randomforest_lopo}
\end{figure}

\begin{figure}[h]
    \centering
    \includegraphics[width=.7\textwidth]{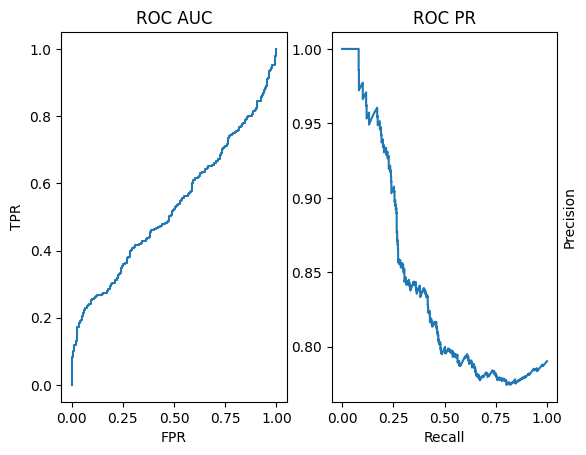}\hfill
    \caption{All features with LOPO, binary classification: MLP}
    \label{fig:AUCplot-allfeatures_binary_classification-mlp_lopo}
\end{figure}

\begin{figure}[h]
    \centering
    \includegraphics[width=.7\textwidth]{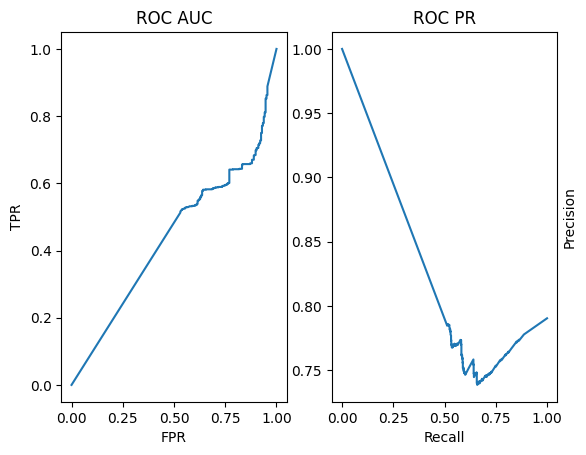}\hfill
    \caption{All features with LOPO, binary classification: EEGNet}
    \label{fig:AUCplot-allfeatures_binary_classification-eegnet_lopo}
\end{figure}

\begin{figure}[h]
    \centering
    \includegraphics[width=.7\textwidth]{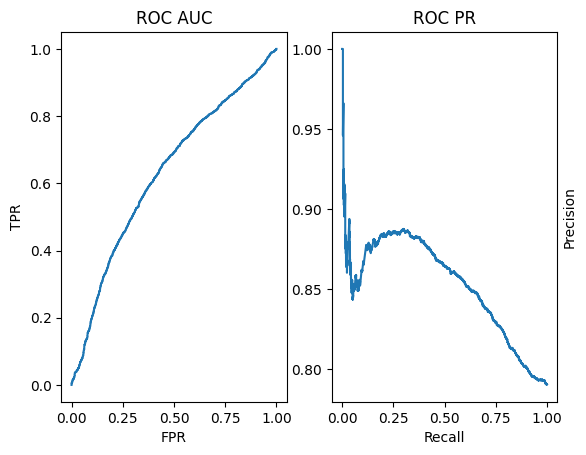}\hfill
    \caption{All features with K-folds, binary classification: Gradient Boosting}
    \label{fig:AUCplot-allfeatures_binary_classification-gradientboosting_5fold}
\end{figure}

\begin{figure}[h]
    \centering
    \includegraphics[width=.7\textwidth]{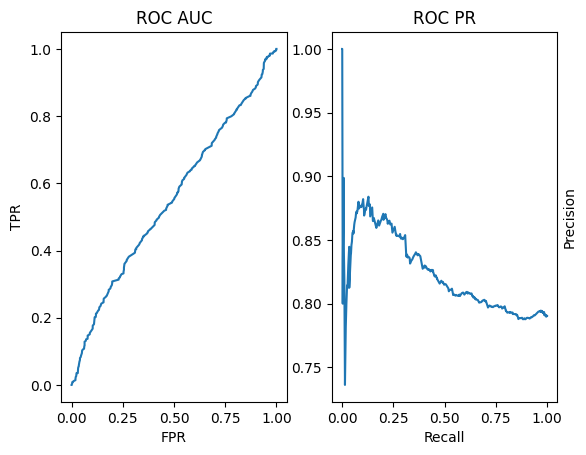}\hfill
    \caption{All features with K-folds, binary classification: Catboost}
    \label{fig:AUCplot-allfeatures_binary_classification-catboost_5fold}
\end{figure}

\begin{figure}[h]
    \centering
    \includegraphics[width=.7\textwidth]{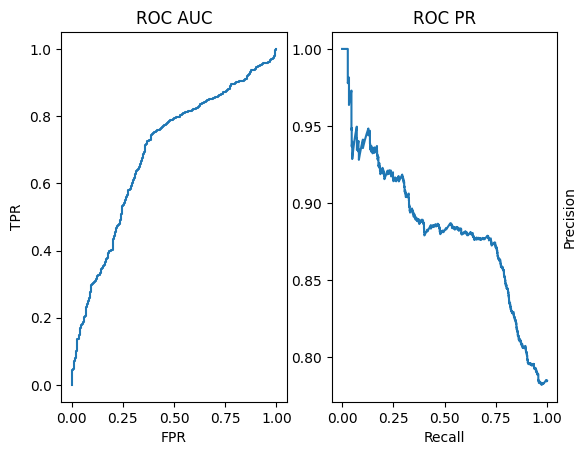}\hfill
    \caption{All features with K-folds, binary classification: LightGBM}
    \label{fig:AUCplot-allfeatures_binary_classification-lightGBM_5fold}
\end{figure}

\begin{figure}[h]
    \centering
    \includegraphics[width=.7\textwidth]{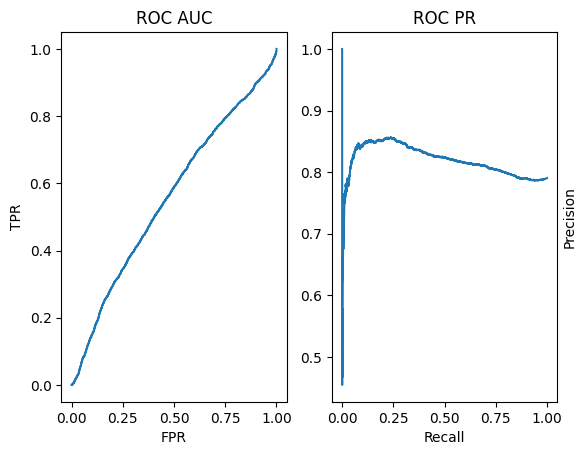}\hfill
    \caption{All features with K-folds, binary classification: SVM}
    \label{fig:AUCplot-allfeatures_binary_classification-svm_5fold}
\end{figure}

\begin{figure}[h]
    \centering
    \includegraphics[width=.7\textwidth]{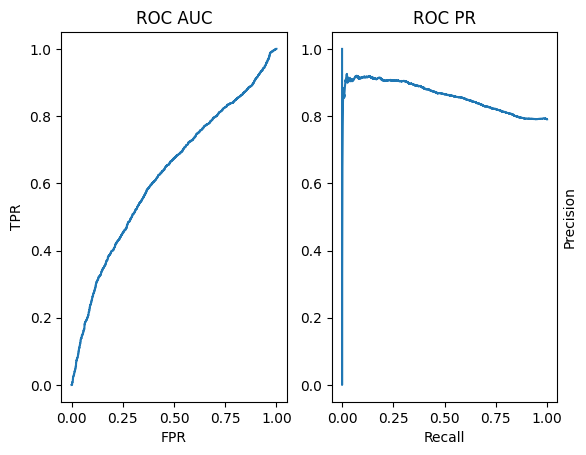}\hfill
    \caption{All features with K-folds, binary classification: Random Forest}
    \label{fig:AUCplot-allfeatures_binary_classification-randomforest_5fold}
\end{figure}

\begin{figure}[h]
    \centering
    \includegraphics[width=.7\textwidth]{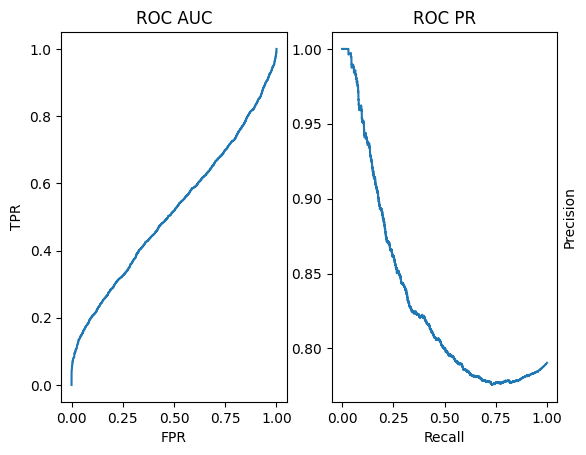}\hfill
    \caption{All features with K-folds, binary classification: MLP}
    \label{fig:AUCplot-allfeatures_binary_classification-mlp_5fold}
\end{figure}

\begin{figure}[h]
    \centering
    \includegraphics[width=.7\textwidth]{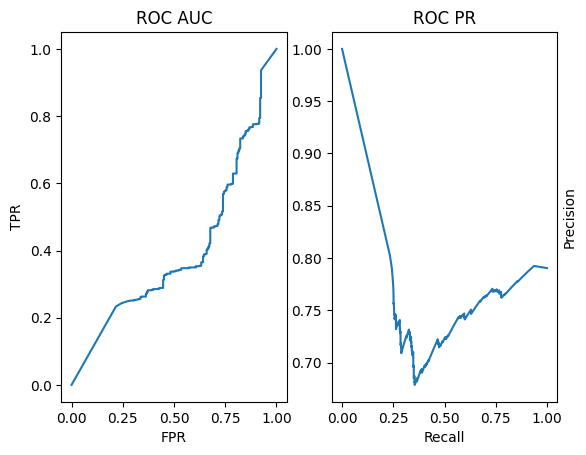}\hfill
    \caption{All features with K-folds, binary classification: EGGNet}
    \label{fig:AUCplot-allfeatures_binary_classification-eegnet_5fold}
\end{figure}

\begin{figure}[h]
    \centering
    \includegraphics[width=.7\textwidth]{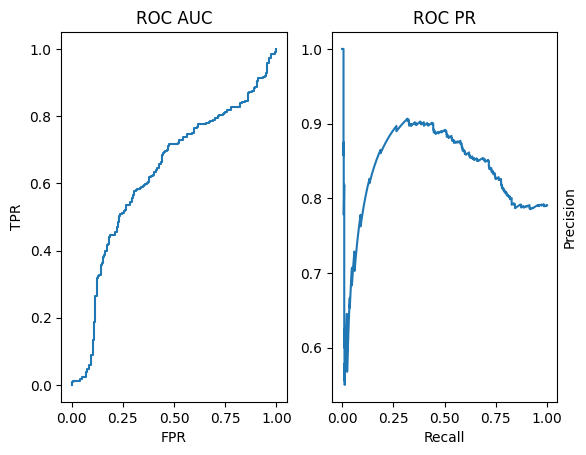}\hfill
    \caption{Multimodal sensor features with LOPO, binary classification: Gradient Boosting}
    \label{fig:AUCplot-sensorfeatures_binary_classification-gradientboosting_lopo}
\end{figure}

\begin{figure}[h]
    \centering
    \includegraphics[width=.7\textwidth]{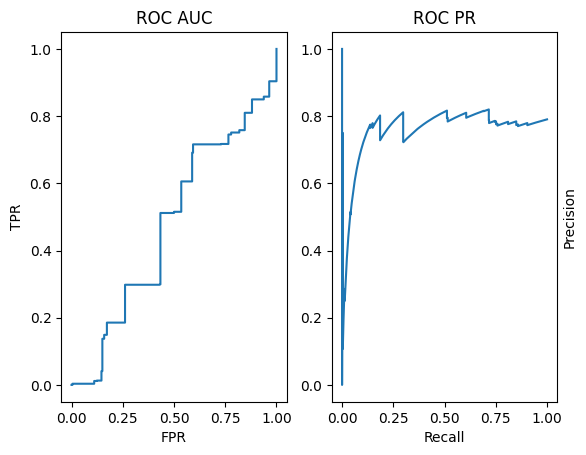}\hfill
    \caption{Multimodal sensor features with LOPO, binary classification: Catboost}
    \label{fig:AUCplot-sensorfeatures_binary_classification-catboost_lopo}
\end{figure}

\begin{figure}[h]
    \centering
    \includegraphics[width=.7\textwidth]{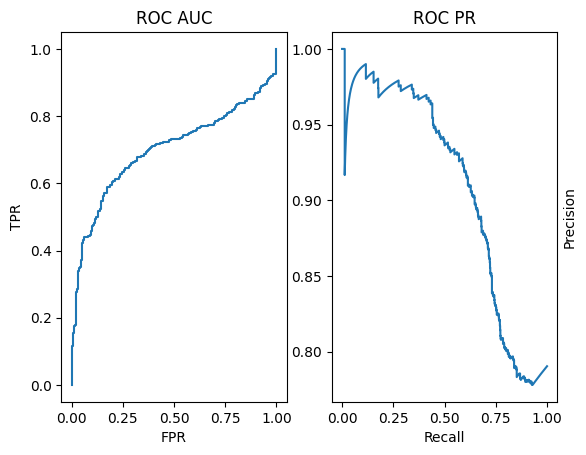}\hfill
    \caption{Multimodal sensor features with LOPO, binary classification: LightGBM}
    \label{fig:AUCplot-sensorfeatures_binary_classification-lightgbm_lopo}
\end{figure}

\begin{figure}[h]
    \centering
    \includegraphics[width=.7\textwidth]{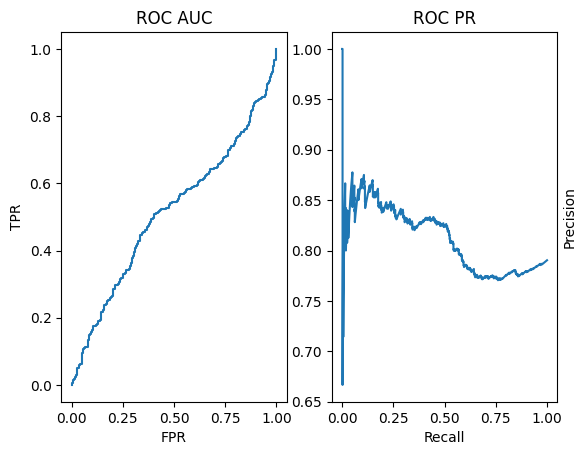}\hfill
    \caption{Multimodal sensor features with LOPO, binary classification: SVM}
    \label{fig:AUCplot-sensorfeatures_binary_classification-svm_lopo}
\end{figure}

\begin{figure}[h]
    \centering
    \includegraphics[width=.7\textwidth]{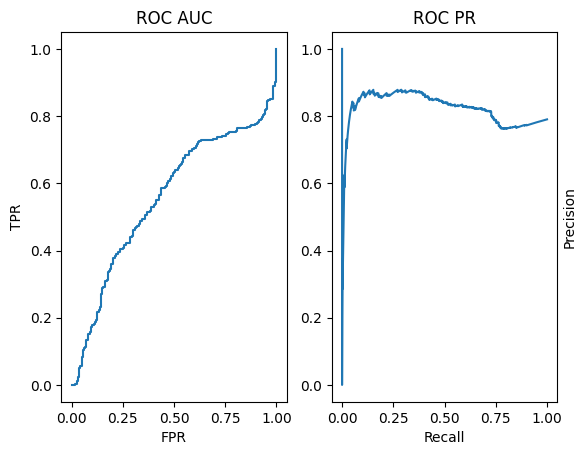}\hfill
    \caption{Multimodal sensor features with LOPO, binary classification: Random Forest}
    \label{fig:AUCplot-sensorfeatures_binary_classification-randomforest_lopo}
\end{figure}

\begin{figure}[h]
    \centering
    \includegraphics[width=.7\textwidth]{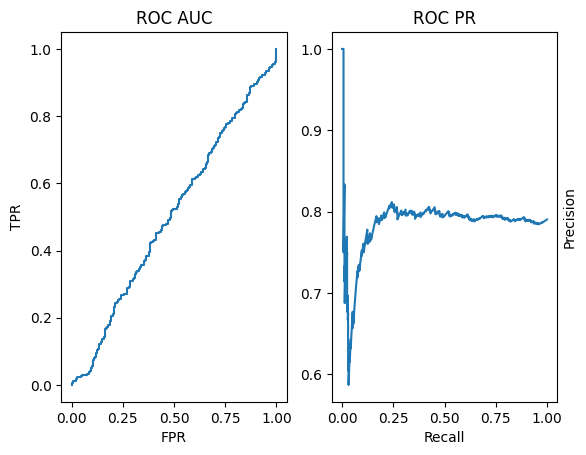}\hfill
    \caption{Multimodal sensor features with LOPO, binary classification: MLP}
    \label{fig:AUCplot-sensorfeatures_binary_classification-mlp_lopo}
\end{figure}

\begin{figure}[h]
    \centering
    \includegraphics[width=.7\textwidth]{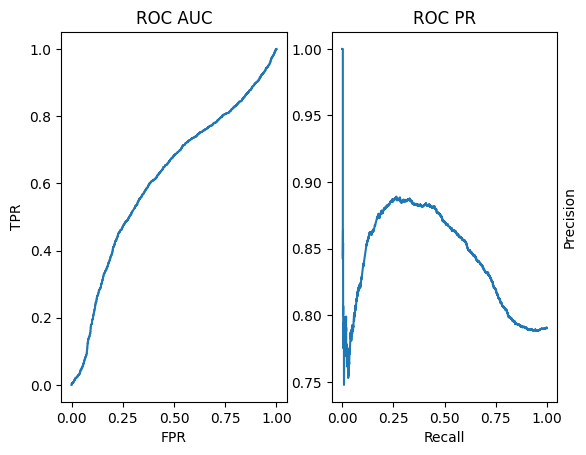}\hfill
    \caption{Multimodal sensor features with K-folds, binary classification: Gradient Boosting}
    \label{fig:AUCplot-sensorfeatures_binary_classification-gradientboosting_kfold}
\end{figure}

\begin{figure}[h]
    \centering
    \includegraphics[width=.7\textwidth]{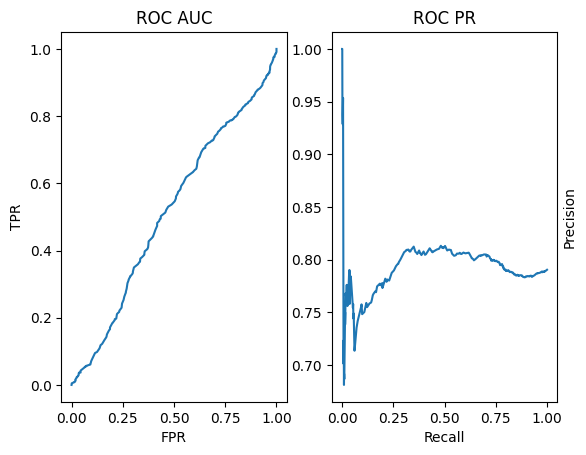}\hfill
    \caption{Multimodal sensor features with K-folds, binary classification: Catboost}
    \label{fig:AUCplot-sensorfeatures_binary_classification-catboost_kfold}
\end{figure}

\begin{figure}[h]
    \centering
    \includegraphics[width=.7\textwidth]{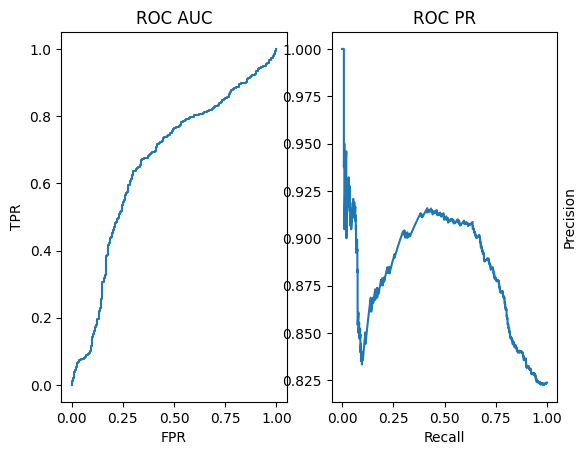}\hfill
    \caption{Multimodal sensor features with K-folds, binary classification: LightGBM}
    \label{fig:AUCplot-sensorfeatures_binary_classification-lightgbm_kfold}
\end{figure}

\begin{figure}[h]
    \centering
    \includegraphics[width=.7\textwidth]{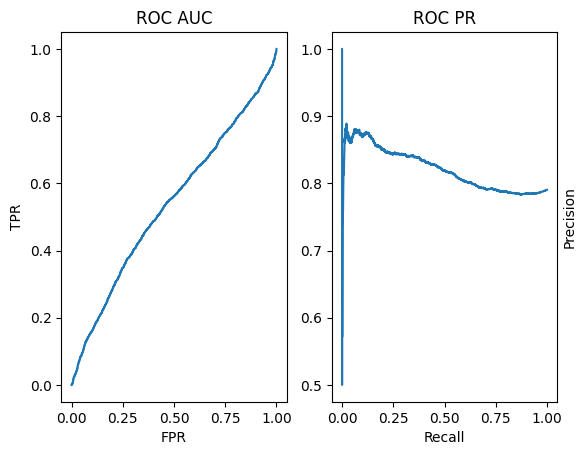}\hfill
    \caption{Multimodal sensor features with K-folds, binary classification: SVM}
    \label{fig:AUCplot-sensorfeatures_binary_classification-svm_kfold}
\end{figure}

\begin{figure}[h]
    \centering
    \includegraphics[width=.7\textwidth]{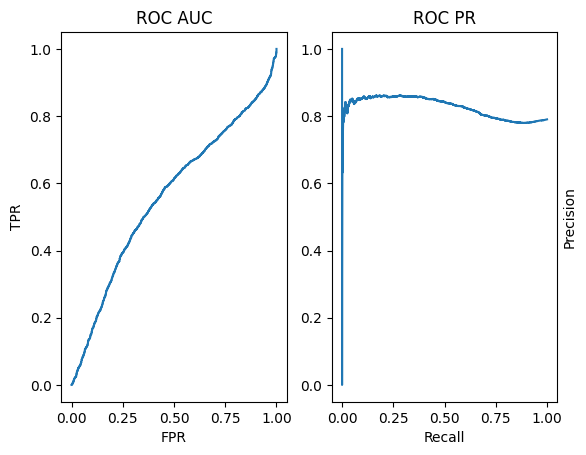}\hfill
    \caption{Multimodal sensor features with K-folds, binary classification: Random Forest}
    \label{fig:AUCplot-sensorfeatures_binary_classification-randomforest_kfold}
\end{figure}

\begin{figure}[h]
    \centering
    \includegraphics[width=.7\textwidth]{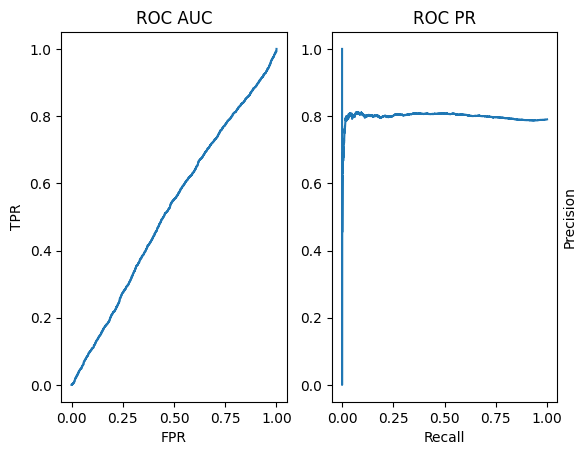}\hfill
    \caption{Multimodal sensor features with K-folds, binary classification: MLP}
    \label{fig:AUCplot-sensorfeatures_binary_classification-mlp_kfold}
\end{figure}

\begin{figure}[h]
    \centering
    \includegraphics[width=.7\textwidth]{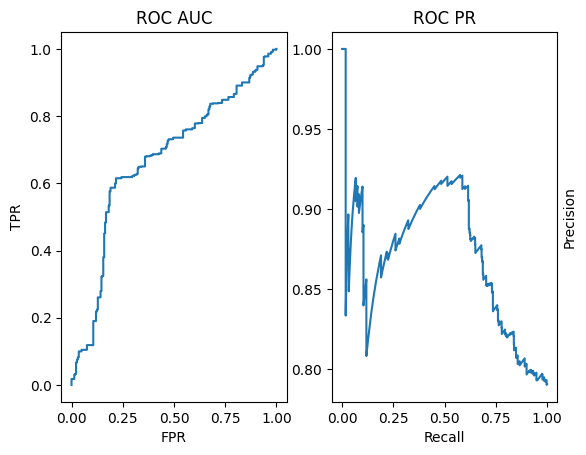}\hfill
    \caption{Top 10 features with LOPO, binary classification: Gradient Boosting}
    \label{fig:AUCplot-top10features_binary_classification-gradientboosting_lopo}
\end{figure}

\begin{figure}[h]
    \centering
    \includegraphics[width=.7\textwidth]{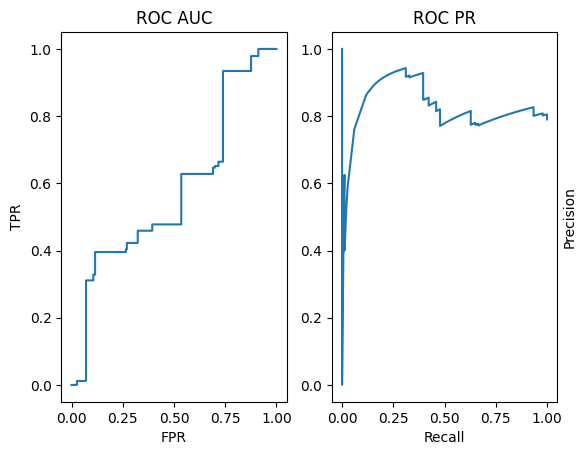}\hfill
    \caption{Top 10 features with LOPO, binary classification: Catboost}
    \label{fig:AUCplot-top10features_binary_classification-catboost_lopo}
\end{figure}

\begin{figure}[h]
    \centering
    \includegraphics[width=.7\textwidth]{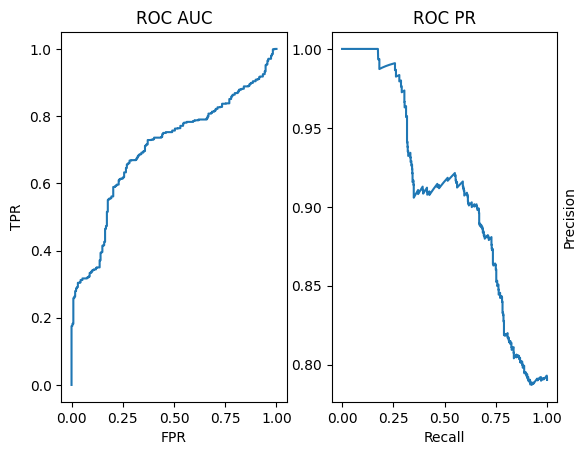}\hfill
    \caption{Top 10 features with LOPO, binary classification: LightGBM}
    \label{fig:AUCplot-top10features_binary_classification-lightgbm_lopo}
\end{figure}

\begin{figure}[h]
    \centering
    \includegraphics[width=.7\textwidth]{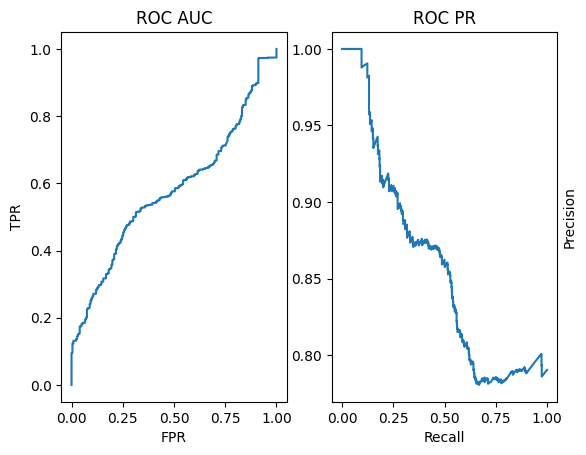}\hfill
    \caption{Top 10 features with LOPO, binary classification: SVM}
    \label{fig:AUCplot-top10features_binary_classification-svm_lopo}
\end{figure}

\begin{figure}[h]
    \centering
    \includegraphics[width=.7\textwidth]{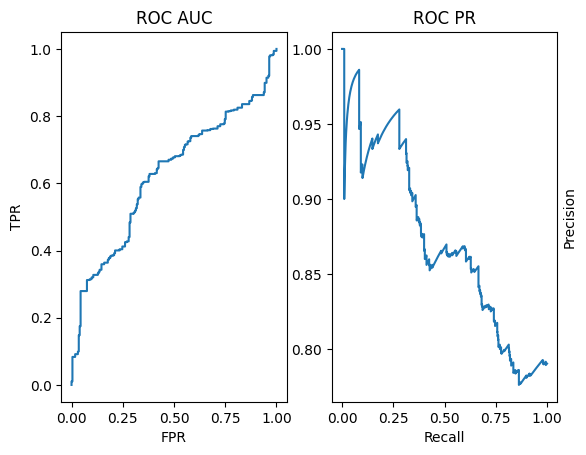}\hfill
    \caption{Top 10 features with LOPO, binary classification: Random Forest}
    \label{fig:AUCplot-top10features_binary_classification-randomforest_lopo}
\end{figure}

\begin{figure}[h]
    \centering
    \includegraphics[width=.7\textwidth]{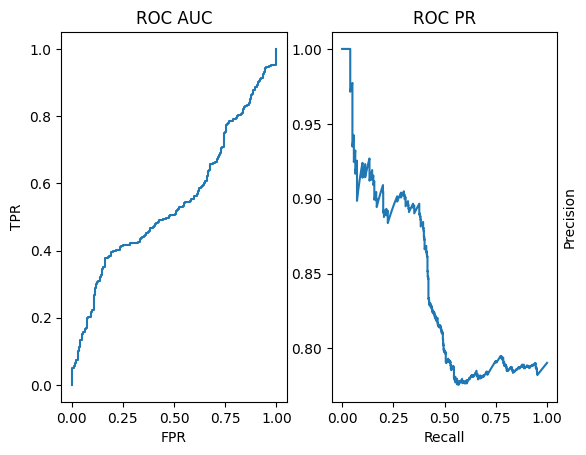}\hfill
    \caption{Top 10 features with LOPO, binary classification: MLP}
    \label{fig:AUCplot-top10features_binary_classification-mlp_lopo}
\end{figure}

\begin{figure}[h]
    \centering
    \includegraphics[width=.7\textwidth]{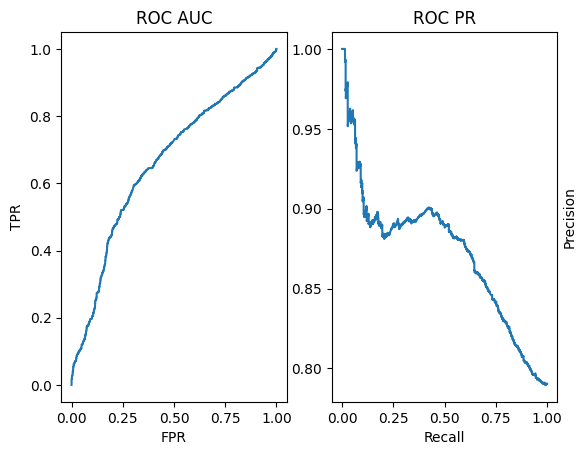}\hfill
    \caption{Top 10 features with K-folds, binary classification: Gradient Boosting}
    \label{fig:AUCplot-top10features_binary_classification-gradientboosting_5fold}
\end{figure}

\begin{figure}[h]
    \centering
    \includegraphics[width=.7\textwidth]{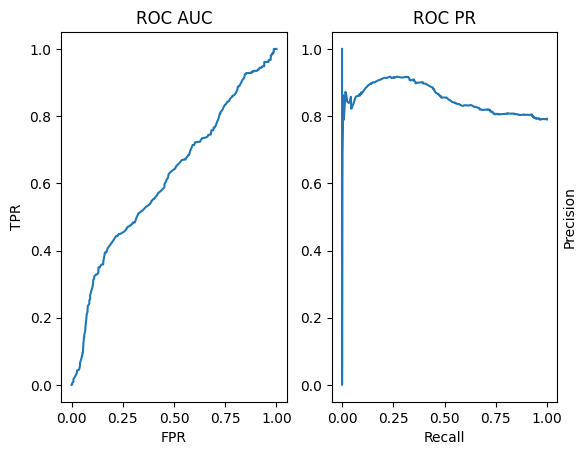}\hfill
    \caption{Top 10 features with K-folds, binary classification: Catboost}
    \label{fig:AUCplot-top10features_binary_classification-catboost_5fold}
\end{figure}

\begin{figure}[h]
    \centering
    \includegraphics[width=.7\textwidth]{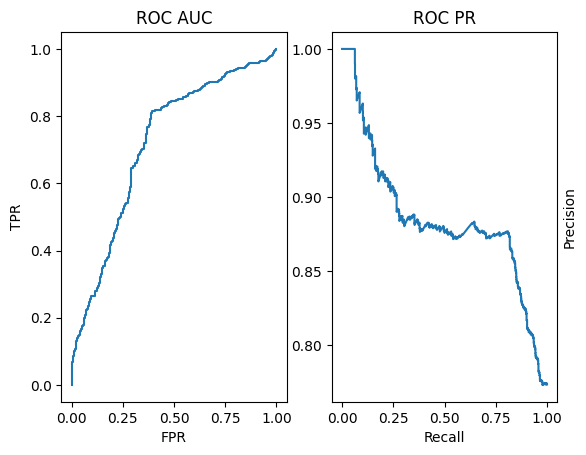}\hfill
    \caption{Top 10 features with K-folds, binary classification: LightGBM}
    \label{fig:AUCplot-top10features_binary_classification-lightgbm_5fold}
\end{figure}

\begin{figure}[h]
    \centering
    \includegraphics[width=.7\textwidth]{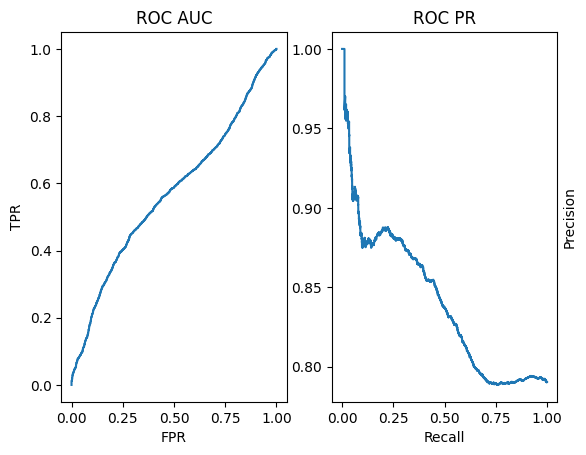}\hfill
    \caption{Top 10 features with K-folds, binary classification: SVM}
    \label{fig:AUCplot-top10features_binary_classification-svm_5fold}
\end{figure}

\begin{figure}[h]
    \centering
    \includegraphics[width=.7\textwidth]{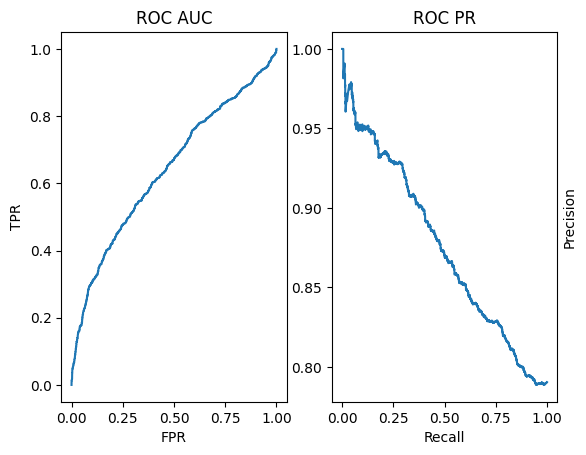}\hfill
    \caption{Top 10 features with K-folds, binary classification: Random Forest}
    \label{fig:AUCplot-top10features_binary_classification-randomforest_5fold}
\end{figure}

\begin{figure}[h]
    \centering
    \includegraphics[width=.7\textwidth]{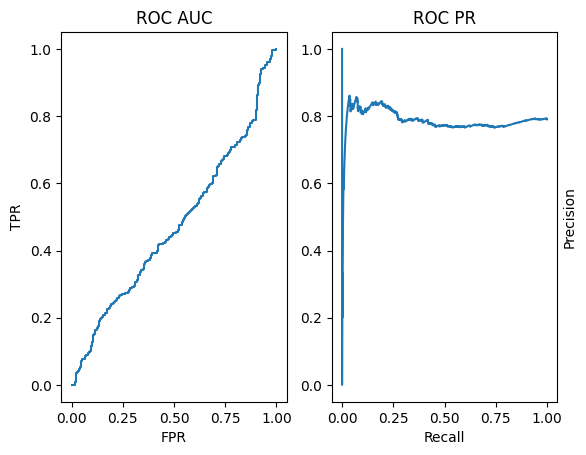}\hfill
    \caption{Wearable features with LOPO, binary classification: Gradient Boosting}
    \label{fig:AUCplot-wearable_device_features_binary_classification-gradientboosting_lopo}
\end{figure}

\begin{figure}[h]
    \centering
    \includegraphics[width=.7\textwidth]{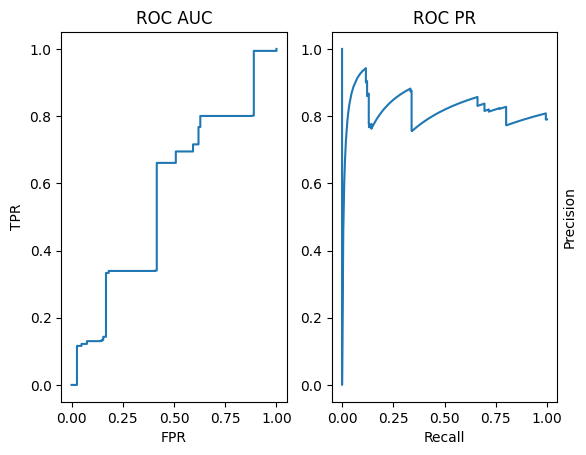}\hfill
    \caption{Wearable features with LOPO, binary classification: CatBoost}
    \label{fig:AUCplot-wearable_device_features_binary_classification-catboost_lopo}
\end{figure}

\begin{figure}[h]
    \centering
    \includegraphics[width=.7\textwidth]{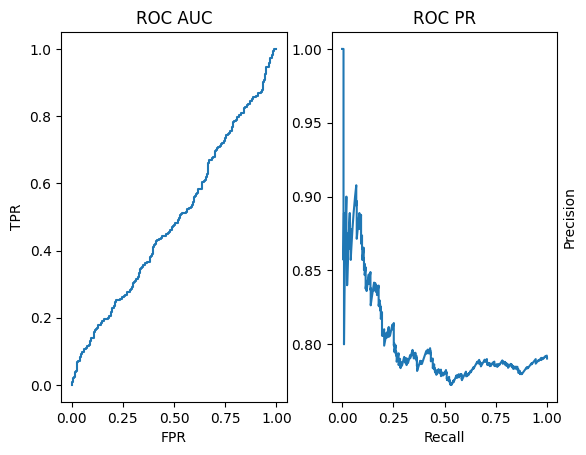}\hfill
    \caption{Wearable features with LOPO, binary classification: LightGBM}
    \label{fig:AUCplot-wearable_device_features_binary_classification-lightgbm_lopo}
\end{figure}

\begin{figure}[h]
    \centering
    \includegraphics[width=.7\textwidth]{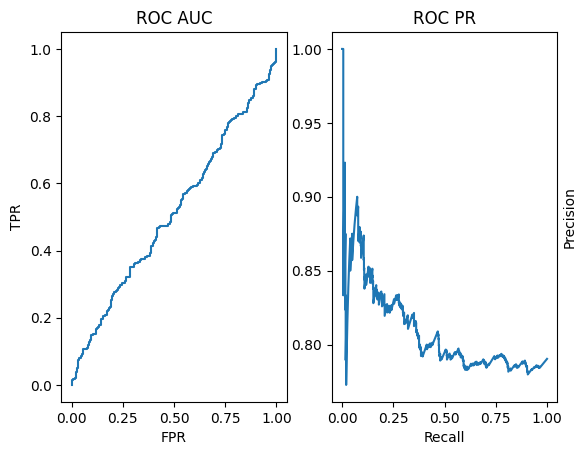}\hfill
    \caption{Wearable features with LOPO, binary classification: SVM}
    \label{fig:AUCplot-wearable_device_features_binary_classification-svm_lopo}
\end{figure}

\begin{figure}[h]
    \centering
    \includegraphics[width=.7\textwidth]{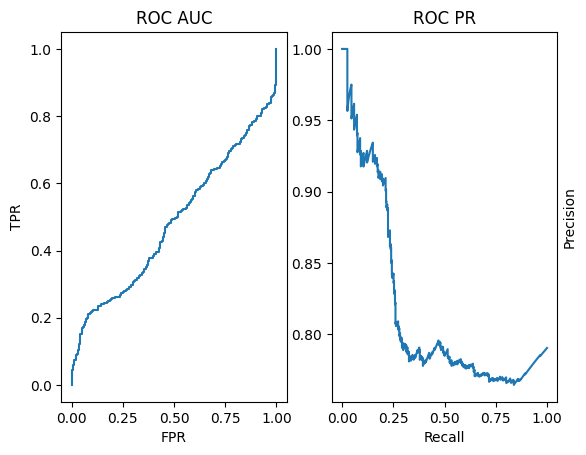}\hfill
    \caption{Wearable features with LOPO, binary classification: Random Forest}
    \label{fig:AUCplot-wearable_device_features_binary_classification-random_forest_lopo}
\end{figure}
 \clearpage
\begin{figure}[h]
    \centering
    \includegraphics[width=.7\textwidth]{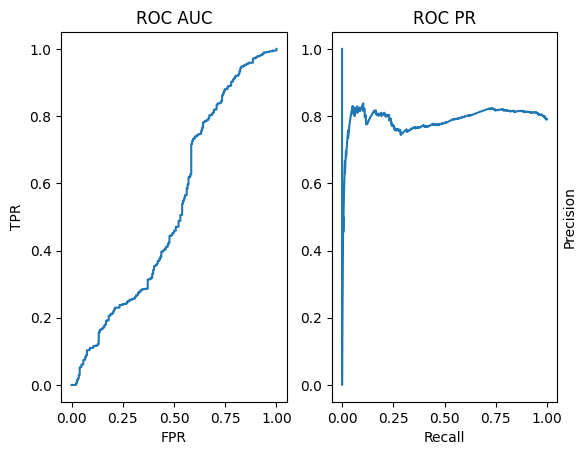}\hfill
    \caption{Wearable features with LOPO, binary classification: MLP}
    \label{fig:AUCplot-wearable_device_features_binary_classification-mlp_lopo}
\end{figure}

\begin{figure}[h]
    \centering
    \includegraphics[width=.7\textwidth]{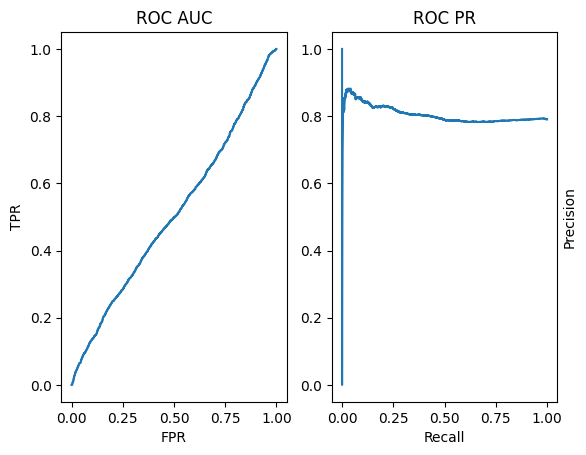}\hfill
    \caption{Wearable features with k-Fold, binary classification: Gradient Boosting}
    \label{fig:E5_2_gradientboosting_kfold}
\end{figure}

\begin{figure}[h]
    \centering
    \includegraphics[width=.7\textwidth]{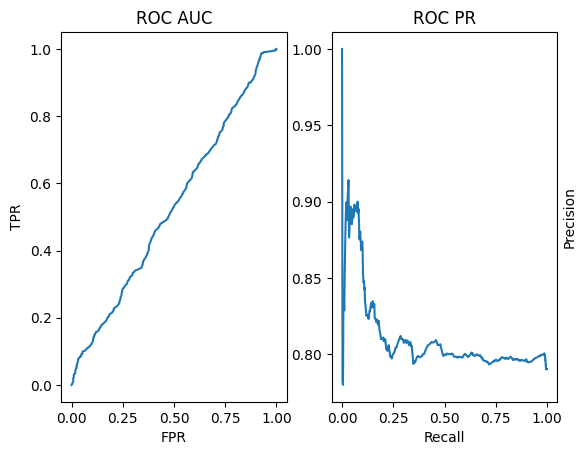}\hfill
    \caption{Wearable features with k-Fold, binary classification: CatBoost}
    \label{fig:E5_2_catboost_kfold}
\end{figure}

\begin{figure}[h]
    \centering
    \includegraphics[width=.7\textwidth]{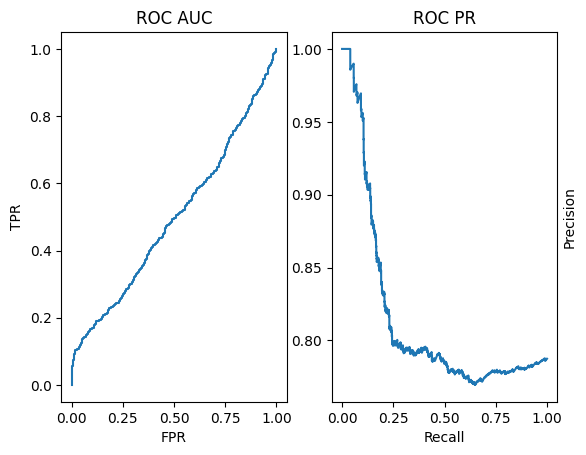}\hfill
    \caption{Wearable features with k-Fold, binary classification: LightGBM}
    \label{fig:E5_2_lightgbm_kfold}
\end{figure}

\begin{figure}[h]
    \centering
    \includegraphics[width=.7\textwidth]{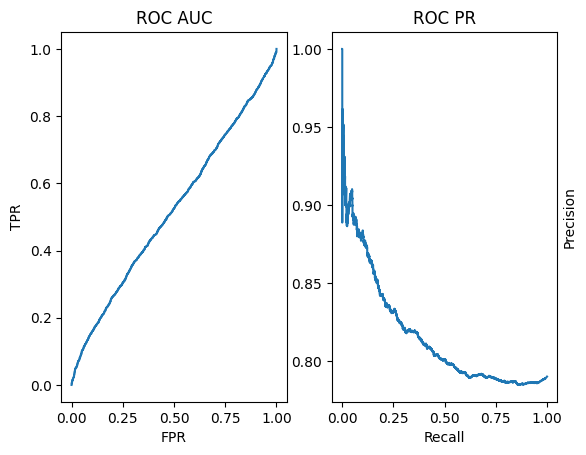}\hfill
    \caption{Wearable features with k-Fold, binary classification: SVM}
    \label{fig:E5_2_svm_kfold}
\end{figure}

\begin{figure}[h]
    \centering
    \includegraphics[width=.7\textwidth]{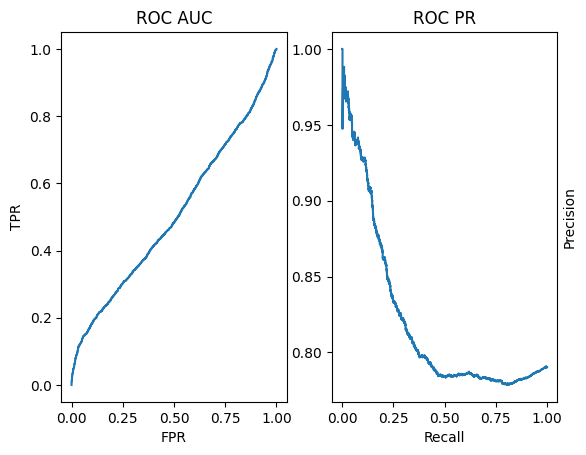}\hfill
    \caption{Wearable features with k-Fold, binary classification: Random Forest}
    \label{fig:E5_2_randomforest_kfold}
\end{figure}

\begin{figure}[h]
    \centering
    \includegraphics[width=.7\textwidth]{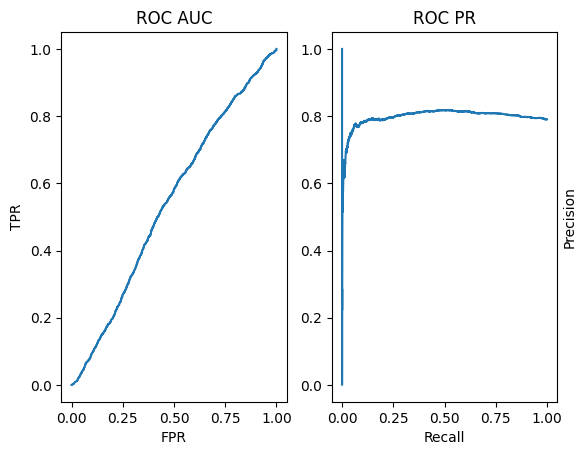}\hfill
    \caption{Wearable features with k-Fold, binary classification: MLP}
    \label{fig:E5_2_mlp_kfold}
\end{figure}

\onecolumn

\begin{figure}[t]
    \centering
    \includegraphics[width=0.85\linewidth]{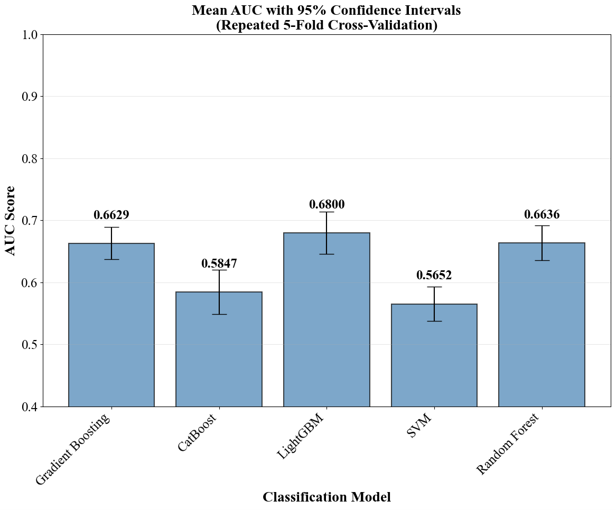}
    \caption{Mean ROC-AUC with 95\% Confidence Intervals Across Classification Models 
        (Repeated 5-Fold Cross-Validation).
        Bar chart comparing the mean ROC-AUC scores of five binary classification models. Gradient Boosting, CatBoost, LightGBM, SVM, and Random Forest evaluated 
        using repeated 5-fold cross-validation with 10 repetitions (50 evaluations per model). 
        Error bars represent 95\% confidence intervals computed using the $t$-distribution. 
        LightGBM achieved the highest mean AUC of 0.6800 (95\% CI: 0.6461--0.7140), 
        while SVM performed lowest at 0.5652 (95\% CI: 0.5375--0.5930).}
    \label{fig:meanCI}
\end{figure}

\onecolumn
\section*{Supplementary Tables}

\begin{table}[ht]
    \centering
    \caption{Performance for different models: All features with 80:20 split, multi-class classification}
    \begin{adjustbox}{max width=1\textwidth}
    \setlength{\tabcolsep}{2pt} 
        \begin{tabular}{lcccccc}
            \toprule
            \centering Model & Gradient Boosting & Catboost & LightGBM & SVM & Random Forest \\
            \midrule
            {\begin{tabular}{@{}c@{}}Macro ROC AUC\\ One-vs-Rest\end{tabular}} & 0.8043 & 0.6932 & 0.9962 & 0.973 & 0.8695  \\
            {\begin{tabular}{@{}c@{}}Macro ROC AUC\\ One-vs-One\end{tabular}}  & 0.8302 & 0.6875 & 0.993  & 0.9574 & 0.7952  \\
            {Macro Precision}           & 0.6238 & 0.1723 & 0.9875 & 0.4417 & 0.2966  \\
            {Recall}                    & 0.5319 & 0.2108 & 0.9085 & 0.3799 & 0.1845\\
            {F1-score}                  & 0.5152 & 0.1808 & 0.9391 & 0.375  & 0.1783  \\
            \bottomrule
        \end{tabular}
    \end{adjustbox}
    \label{tab:ml_metrics_All_features_multi_classification}
\end{table}

\begin{table}[ht]
    \centering
    \caption{Top 10 important features using Gradient Boosting, Catboost, and LightGBM when conducting binary prediction with an 80:20 split. The meaning of each feature is explained in Table \ref{table.wearabeldeviceparameters}.}
    \begin{adjustbox}{max width=1\textwidth}
    \setlength{\tabcolsep}{2pt} 
        \begin{tabular}{c>{\raggedright\arraybackslash}p{1.5cm}c>{\raggedright\arraybackslash}p{1.5cm}c>{\raggedright\arraybackslash}p{1.5cm}cc}
            \toprule
            \multirow{2}{*}{{Order}} & \multicolumn{2}{c}{{Gradient Boosting}} & \multicolumn{2}{c}{{Catboost}} & \multicolumn{2}{c}{{LightGBM}} \\
            \cmidrule(lr){2-3} \cmidrule(lr){4-5} \cmidrule(lr){6-7}
            & \centering Feature & Importance & \centering Feature & Importance & \centering Feature & Importance \\
            \midrule
            1 & BMI\_index & 0.279819 & Age & 55.244066 & BMI\_index & 22 \\
            2 & Heart Rate & 0.163589 & Type of skins & 29.933423 & Heart Rate & 13 \\
            3 & Age & 0.160837 & Weight & 11.040225 & Age & 13 \\
            4 & Type of skins & 0.156214 & $\delta$ & 3.782286 & Weight & 9 \\
            5 & Weight & 0.097077 & Type of data & 0.000000 & Height & 8 \\
            6 & T & 0.050864 & F\_Ae & 0.000000 & M & 6 \\
            7 & Height & 0.044463 & Level of BP & 0.000000 & T & 6 \\
            8 & A460 & 0.011797 & Smoking routine & 0.000000 & A460 & 5 \\
            9 & Anadn & 0.009720 & BMI\_index & 0.000000 & Kv100 & 2 \\
            10 & M & 0.009113 & Height & 0.000000 & Type of skins & 2 \\
            \bottomrule
        \end{tabular}
    \end{adjustbox}
    \label{tab:merged_featuresbinary_prediction}
\end{table}

\begin{table}[ht]
\centering
\caption{Top 10 important features using Gradient Boosting, Catboost, and LightGBM for multi-class classification with an 80:20 split. The meaning of each feature is explained in Table \ref{table.wearabeldeviceparameters}. }
\begin{adjustbox}{max width=1\textwidth}
\setlength{\tabcolsep}{2pt} 
    \begin{tabular}{c>{\raggedright\arraybackslash}p{1.5cm}c>{\raggedright\arraybackslash}p{1.5cm}c>{\raggedright\arraybackslash}p{1.5cm}cc}
        \toprule
        \multirow{2}{*}{{Order}} & \multicolumn{2}{c}{{Gradient Boosting}} & \multicolumn{2}{c}{{Catboost}} & \multicolumn{2}{c}{{LightGBM}} \\
        \cmidrule(lr){2-3} \cmidrule(lr){4-5} \cmidrule(lr){6-7}
        & \centering Feature & Importance & \centering Feature & Importance & \centering Feature & Importance \\
        \midrule
        1 & Heart Rate & 0.682942 & Heart Rate & 87.150766 & Weight & 18 \\
        2 & A365 & 0.179464 & Type of skins & 8.122651 & Height & 9 \\
        3 & BMI\_index & 0.089107 & Anadn & 3.149739 & BMI\_index & 8 \\
        4 & Type of skins & 0.043119 & $\delta$ & 1.576843 & Heart Rate & 7 \\
        5 & Height & 0.004020 & F\_An & 0.000000 & Type of skins & 7 \\
        6 & Age & 0.001002 & Level of BP & 0.000000 & A365 & 7 \\
        7 & POM & 0.000173 & Smoking routine & 0.000000 & Age & 5 \\
        8 & T & 0.000173 & BMI\_index & 0.000000 & POM & 5 \\
        9 & F\_An & 0.000000 & Height & 0.000000 & T & 5 \\
        10 & Level of BP & 0.000000 & Weight & 0.000000 & F\_Ar & 2 \\
        \bottomrule
    \end{tabular}
\end{adjustbox}

\label{tab:multiclass_features}
\end{table}

\begin{table}[ht]
\centering
\caption{Performance for different models: All features with LOPO, multi-class classification.}
\begin{adjustbox}{max width=1\textwidth}
\setlength{\tabcolsep}{2pt} 
\begin{tabular}{lcccccc}
\toprule
\centering Model & Gradient Boosting & Catboost & LightGBM & SVM & Random Forest  \\
\midrule
{\begin{tabular}{@{}c@{}}Macro ROC AUC\\ One-vs-Rest\end{tabular}} & 0.4466 & 0.3279 & 0.5678 & 0.4208 & 0.3092  \\
{Macro Precision} & 0.1719 & 0.1346 & 0.1336 & 0.1311 & 0.1317  \\
{Recall} & 0.1776 & 0.1384 & 0.1493 & 0.1631 & 0.1667  \\
{F1-score} & 0.1698 & 0.1307 & 0.1408 & 0.1454 & 0.1472\\
\bottomrule
\end{tabular}
\end{adjustbox}
\label{tab:Multi_LOPO}
\end{table}

\begin{table}[ht]
\centering
\caption{Performance for different models: All features with 5-fold, multi-class classification.}
\begin{adjustbox}{max width=1\textwidth}
\setlength{\tabcolsep}{2pt} 
\begin{tabular}{lcccccc}
\toprule
\centering Model & Gradient Boosting & Catboost & LightGBM & SVM & Random Forest  \\
\midrule
{\begin{tabular}{@{}c@{}}Macro ROC AUC\\ One-vs-Rest\end{tabular}} & 0.4804 & 0.4103 & 0.5812 & 0.4412 & 0.4663  \\
{\begin{tabular}{@{}c@{}}Macro ROC AUC\\ One-vs-One\end{tabular}} & 0.4492 & 0.4132 & 0.5057 & 0.4539 & 0.4207  \\
{Macro Precision} & 0.1783 & 0.1434 & 0.1554 & 0.1357 & 0.1223  \\
{Recall} & 0.1746 & 0.1539 & 0.1652 & 0.1628 & 0.1667 \\
{F1-score} & 0.1736 & 0.1474 & 0.1578 & 0.1465 & 0.1411  \\
\bottomrule
\end{tabular}
\end{adjustbox}
\label{tab:Multi-Kfold}
\end{table}

\begin{table}[ht]
\centering
\caption{Performance for different models: Multimodal sensor features with LOPO, binary classification.}
\begin{adjustbox}{max width=1\textwidth}
\setlength{\tabcolsep}{2pt} 
\begin{tabular}{lcccccc}
\toprule
Model & Gradient Boosting & Catboost & LightGBM & SVM & Random Forest & MLP \\
\midrule
{ROC AUC} & 0.6265 & 0.4753 & 0.698 & 0.5124 & 0.556 & 0.5034 \\
{PR AUC} & 0.8379 & 0.7556 & 0.9091 & 0.8113 & 0.8209 & 0.7855 \\
\bottomrule
\end{tabular}
\end{adjustbox}
\label{tab:sensor_features_binary_LOPO}
\end{table}

\begin{table}[ht]
\centering
\caption{Performance for different models: Multimodal sensor features with 5-fold, binary classification.}
\begin{adjustbox}{max width=1\textwidth}
\setlength{\tabcolsep}{2pt} 
\begin{tabular}{lcccccc}
\toprule
Model & Gradient Boosting & Catboost & LightGBM & SVM & Random Forest & MLP \\
\midrule
{ROC AUC} & 0.6137 & 0.5145 & 0.6601 & 0.5389 & 0.5607 & 0.5216 \\
{PR AUC} & 0.8424 & 0.7914 & 0.8839 & 0.8207 & 0.8261 & 0.7983 \\
\bottomrule
\end{tabular}
\end{adjustbox}
\label{tab:sensor_features_binary_kfold}
\end{table}

\begin{table}[ht]
\centering
\caption{Performance for different models: Top 10 features with LOPO, binary classification.}
\begin{adjustbox}{max width=1\textwidth}
\setlength{\tabcolsep}{2pt} 
\begin{tabular}{lcccccc}
\toprule
Model & Gradient Boosting & Catboost & LightGBM & SVM & Random Forest & MLP \\
\midrule
{ROC AUC} & 0.6699 & 0.5788 & 0.7041 & 0.578 & 0.6232 & 0.5454 \\
{PR AUC} & 0.8689 & 0.8213 & 0.9087 & 0.8591 & 0.8714 & 0.8413 \\
\bottomrule
\end{tabular}
\end{adjustbox}
\label{tab:top10_binary_LOPO}
\end{table}

\begin{table}[ht]
\centering
\caption{Performance for different models: Top 10 features with LOPO, multi-class classification.}
\begin{adjustbox}{max width=1\textwidth}
\setlength{\tabcolsep}{2pt} 
\begin{tabular}{lcccccc}
\toprule
\centering Model & Gradient Boosting & Catboost & LightGBM & SVM & Random Forest \\
\midrule
{\begin{tabular}{@{}c@{}}Macro ROC AUC\\ One-vs-Rest\end{tabular}} & 0.4946 & 0.4463 & 0.633 & 0.4466 & 0.3352  \\
{\begin{tabular}{@{}c@{}}Macro ROC AUC\\ One-vs-One\end{tabular}} & 0.4933 & 0.4084 & 0.5244 & 0.4344 & 0.3007 \\
{Macro Precision} & 0.1935 & 0.1558 & 0.1636 & 0.1335 & 0.1317 \\
{Recall} & 0.2182 & 0.1737 & 0.1742 & 0.1552 & 0.1667  \\
{F1-score} & 0.1947 & 0.159 & 0.1679 & 0.1429 & 0.1472  \\
\bottomrule
\end{tabular}
\end{adjustbox}
\label{tab:top10_multi_LOPO}
\end{table}

\begin{table}[ht]
\centering
\caption{Performance for different models: Top 10 features with 5-fold, multi-class classification.}
\begin{adjustbox}{max width=1\textwidth}
\setlength{\tabcolsep}{2pt} 
\begin{tabular}{lcccccc}
\toprule
Model & Gradient Boosting & Catboost & LightGBM & SVM & Random Forest \\
\midrule
{\begin{tabular}{@{}c@{}}Macro ROC AUC\\ One-vs-Rest\end{tabular}} & 0.5418 & 0.5315 & 0.6412 & 0.5022 & 0.3857 \\
{\begin{tabular}{@{}c@{}}Macro ROC AUC\\ One-vs-One\end{tabular}} & 0.5507 & 0.4474 & 0.5585 & 0.4755 & 0.3615 \\
{Macro Precision} & 0.2314 & 0.1401 & 0.1778 & 0.1525 & 0.2896 \\
{Recall} & 0.2347 & 0.1476 & 0.2031 & 0.1705 & 0.1711 \\
{F1-score} & 0.2224 & 0.1416 & 0.1885 & 0.1596 & 0.1502 \\
\bottomrule
\end{tabular}
\end{adjustbox}
\label{tab:top10_multi_kfold}
\end{table}


\begin{table}[h!]
\centering
\begin{tabularx}{\textwidth}{l X}  
\toprule
\textbf{Feature Group} & \textbf{Feature Names} \\
\midrule
Laser Doppler Flowmetry (LDF) & $M$, $\delta$, $T$, $Kv_{100}$ \\
Fluorescence Spectroscopy (FS) & $A_{365}$, $A_{460}$, $A_{nadn}$, $POM$ \\
Wavelet-Extracted Rhythms & $Ae$, $An$, $Am$, $Ar$, $Ac$, $F\_Ae$, $F\_An$, $F\_Am$, $F\_Ar$, $F\_Ac$ \\
Demographic Features & Age, Gender, Weight, Height, BMI, Smoking status, Blood Pressure, Heart Rate \\
\midrule
\textbf{Total} & 27 features \\
\bottomrule
\end{tabularx}

\caption{Summary of all features grouped by type.}
\label{tab:features}
\end{table}

\begin{table}[h!]
    \centering
    \caption{Depression, anxiety and stress scale (DASS21) questionnaire responses.}
    \setlength{\tabcolsep}{3pt}
    \renewcommand{\arraystretch}{1.6}
    \adjustbox{width=\columnwidth}{
    \begin{tabular}{|c|m{10cm}|c|c|c|c|}
        \hline
        No.& \centering \textbf{Question} & \textbf{0} & \textbf{1} & \textbf{2} & \textbf{3} \\ \hline
        
        1 (s) & I found it hard to wind down & 0 & 1 & 2 & 3 \\ \hline
      
        2 (a) & I was aware of dryness of my mouth & 0 & 1 & 2 & 3 \\ \hline
    
        3 (d) & I couldn't seem to experience any positive feeling at all & 0 & 1 & 2 & 3 \\ \hline
     
        4 (a) & I experienced breathing difficulty (e.g. excessively rapid breathing, breathlessness in the absence of physical exertion) & 0 & 1 & 2 & 3 \\ \hline
     
        5 (d) & I found it difficult to work up the initiative to do things & 0 & 1 & 2 & 3 \\ \hline
        6 (s) & I tended to over-react to situations & 0 & 1 & 2 & 3 \\ \hline
        
        7 (a) & I experienced trembling (e.g. in the hands) & 0 & 1 & 2 & 3 \\ \hline

        8 (s) & I felt that I was using a lot of nervous energy & 0 & 1 & 2 & 3 \\ \hline
     
        9 (a) & I was worried about situations in which I might panic and make a fool of myself & 0 & 1 & 2 & 3 \\ \hline

        10 (d) & I felt that I had nothing to look forward to & 0 & 1 & 2 & 3 \\ \hline
 
        11 (s) & I found myself getting agitated & 0 & 1 & 2 & 3 \\ \hline
      
        12 (s) & I found it difficult to relax & 0 & 1 & 2 & 3 \\ \hline
    
        13 (d) & I felt down-hearted and blue & 0 & 1 & 2 & 3 \\ \hline
       
        14 (s) & I was intolerant of anything that kept me from getting on with what I was doing & 0 & 1 & 2 & 3 \\ \hline
    
        15 (a) & I felt I was close to panic & 0 & 1 & 2 & 3 \\ \hline
     
        16 (d) & I was unable to become enthusiastic about anything & 0 & 1 & 2 & 3 \\ \hline
    
        17 (d) & I felt I wasn't worth much as a person & 0 & 1 & 2 & 3 \\ \hline
      
        18 (s) & I felt that I was rather touchy & 0 & 1 & 2 & 3 \\ \hline
    
        19 (a) & I was aware of the action of my heart in the absence of physical exertion (e.g. sense of heart rate increase, heart missing a beat) & 0 & 1 & 2 & 3 \\ \hline
     
        20 (a) & I felt scared without any good reason & 0 & 1 & 2 & 3 \\ \hline
  
        21 (d) & I felt that life was meaningless & 0 & 1 & 2 & 3 \\ \hline

    \end{tabular}
    }
    \label{tab:das21responses}
\end{table}

\begin{table}[h]
\centering
\setlength{\tabcolsep}{2pt}
\renewcommand{\arraystretch}{3.5}
\caption{Blood perfusion (M*) with standard deviation and Maximum amplitude with standard
deviation of the endothelial (A-E), neurogenic (A-N), myogenic (A-M), breath (A-R) and pulse (A-C)
mechanism for Wellbeing vs Non-wellbeing, *, p<0.01, Mann-Whitney U test.}
\begin{adjustbox}{width=\textwidth}
\begin{tabular}{lcccccccccccccc}
\toprule
\textbf{Subgroup} & \textbf{M\_mean} & \textbf{p-value} & \textbf{A-E\_mean} & \textbf{p-value} & \textbf{A-N\_mean} & \textbf{p-value} & \textbf{A-M\_mean} & \textbf{p-value} & \textbf{A-R\_mean} & \textbf{p-value} & \textbf{A-C\_mean} & \textbf{p-value} \\ \midrule

All & \begin{tabular}{@{}c@{}}22.54 \\ (4.73 - 37.23)\end{tabular} & & \begin{tabular}{@{}c@{}}1.49 \\ (0.34 - 3.36)\end{tabular} & & \begin{tabular}{@{}c@{}}1.46 \\ (0.33 - 3)\end{tabular} & & \begin{tabular}{@{}c@{}}1.17 \\ (0.31 - 2.4)\end{tabular} & & \begin{tabular}{@{}c@{}}0.7 \\ (0.22 - 1.29)\end{tabular} & & \begin{tabular}{@{}c@{}}0.93 \\ (0.39 - 1.68)\end{tabular} & \\

Wellbeing  & \begin{tabular}{@{}c@{}}21.02 \\ (4.73 - 35.59)\end{tabular} & 0.016121 & \begin{tabular}{@{}c@{}}1.44 \\ (0.34 - 3.42)\end{tabular} & 0.171585 & \begin{tabular}{@{}c@{}}1.4 \\ (0.31 - 2.99)\end{tabular} & 0.27169 & \begin{tabular}{@{}c@{}}1.11 \\ (0.29 - 2.35)\end{tabular} & 0.123042 & \begin{tabular}{@{}c@{}}0.66 \\ (0.2 - 1.18)\end{tabular} & 0.0614 & \begin{tabular}{@{}c@{}}0.89 \\ (0.31 - 1.74)\end{tabular} & 0.06069 \\

Non-Wellbeing & \begin{tabular}{@{}c@{}}26.49 \\ (8.59 - 36.71)\end{tabular} & & \begin{tabular}{@{}c@{}}1.62 \\ (0.64 - 2.7)\end{tabular} & & \begin{tabular}{@{}c@{}}1.61 \\ (0.49 - 2.96)\end{tabular} & & \begin{tabular}{@{}c@{}}1.33 \\ (0.5 - 2.32)\end{tabular} & & \begin{tabular}{@{}c@{}}0.81 \\ (0.47 - 1.44)\end{tabular} & & \begin{tabular}{@{}c@{}}1.02 \\ (0.58 - 1.61)\end{tabular} & \\

\bottomrule
\end{tabular}
\end{adjustbox}
\label{tab.Wellbeing1}
\end{table}

\begin{table}[h]
\centering
\setlength{\tabcolsep}{2pt}
\renewcommand{\arraystretch}{3.5}
\caption{The parameters with standard deviation for Wellbeing vs Non-wellbeing: Kv100, $\delta^{*}$, $T^*$,
A365, A460, Anadn, ${POM}^{*}$, F-E; F-N; F-M; F-R; F-C, *, $p<0.01$, Mann-Whitney U test.}
\begin{adjustbox}{width=\textwidth}
\begin{tabular}{lccccccccccccc}
\toprule
\textbf{Subgroup}  & \textbf{Kv100\_mean} & \textbf{$\delta$\_mean} & \textbf{p-value} & \textbf{T\_mean} & \textbf{p-value} & \textbf{A365\_mean} & \textbf{A460\_mean} & \textbf{Anadn\_mean} & \textbf{POM\_mean} & \textbf{p-value} \\ \midrule

All & \begin{tabular}{@{}c@{}}21.09 \\ (6.86 - 49.55)\end{tabular} & \begin{tabular}{@{}c@{}}3.96 \\ (1.21 - 7.41)\end{tabular} & & \begin{tabular}{@{}c@{}}31.36 \\ (22.95 - 35.79)\end{tabular} & & \begin{tabular}{@{}c@{}}86.82 \\ (4.42 - 158.6)\end{tabular} & \begin{tabular}{@{}c@{}}59.3 \\ (12.92 - 106.52)\end{tabular} & \begin{tabular}{@{}c@{}}1.01 \\ (0.4 - 4.54)\end{tabular} & \begin{tabular}{@{}c@{}}8.74 \\ (0.99 - 22.15)\end{tabular} \\

Wellbeing & \begin{tabular}{@{}c@{}}21.12 \\ (6.71 - 48.75)\end{tabular} & \begin{tabular}{@{}c@{}}3.64 \\ (0.98 - 6.93)\end{tabular} & 0.007252 & \begin{tabular}{@{}c@{}}30.65 \\ (22.68 - 35.6)\end{tabular} & 0.018427 & \begin{tabular}{@{}c@{}}85.43 \\ (9.5 - 130.9)\end{tabular} & \begin{tabular}{@{}c@{}}60.64 \\ (17.2 - 106.8)\end{tabular} & \begin{tabular}{@{}c@{}}1.01 \\ (0.41 - 4.77)\end{tabular} & \begin{tabular}{@{}c@{}}7.7 \\ (0.85 - 20.43)\end{tabular} & 0.010656 \\

Non-Wellbeing & \begin{tabular}{@{}c@{}}21 \\ (7.56 - 48.44)\end{tabular} & \begin{tabular}{@{}c@{}}4.79 \\ (2.25 - 7.65)\end{tabular} & & \begin{tabular}{@{}c@{}}33.21 \\ (30.14 - 35.82)\end{tabular} & & \begin{tabular}{@{}c@{}}90.43 \\ (2.52 - 159.81)\end{tabular} & \begin{tabular}{@{}c@{}}55.81 \\ (12.05 - 88.06)\end{tabular} & \begin{tabular}{@{}c@{}}1 \\ (0.39 - 4.16)\end{tabular} & \begin{tabular}{@{}c@{}}11.42 \\ (1.95 - 24.42)\end{tabular} \\

\bottomrule
\end{tabular}
\end{adjustbox}
\label{tab:Wellbeing_Kv100}
\end{table}

\begin{table}[h!]
\centering
\caption{Model investigation with all collected features, in which we trained models with different data split methods: 80:20, 5-fold, and leave one patient out (LOPO). In addition, "x" presents the cases that are being trained in our investigation.}
\begin{adjustbox}{max width=\textwidth}
\begin{tabular}{|l|c|c|c|c|c|c|}
\hline
\multirow{2}{*}{Model} & \multicolumn{6}{c|}{All Features} \\
\cline{2-7}
 & \multicolumn{2}{c|}{Split 80:20} & \multicolumn{2}{c|}{5-Fold Validation} & \multicolumn{2}{c|}{LOPO} \\
\cline{2-7}
 & Binary & Multi-class & Binary & Multi-class & Binary & Multi-class \\
\hline
Gradient Boosting & x & x & x & x & x & x \\ \hline
Catboost & x & x & x & x & x & x \\ \hline
LightGBM & x & x & x & x & x & x \\ \hline
SVM & x & x & x & x & x & x \\ \hline
Random Forest & x & x & x & x & x & x \\ \hline
MLP & x &  & x &  & x &  \\ \hline
EEGNet & x & & x &  & x & \\
\hline
\end{tabular}
\end{adjustbox}
\label{table:all_features}
\end{table}

\begin{table}[h!]
\centering
\renewcommand{\arraystretch}{1.2} 
\caption{Model investigation with sensor features and top-10 important features. We trained the models with various data split methods: 80/20 train-test split, 5-fold cross-validation, and LOPO cross-validation. The "x" marks in the table indicate the specific model-data combinations investigated in this study.}
\begin{adjustbox}{max width=\textwidth}
\begin{tabular}{|l|c|c|c|c|c|c|c|c|}
\hline
\multirow{2}{*}{Model} & \multicolumn{4}{c|}{Sensor Features} & \multicolumn{4}{c|}{Top-10 Important Features} \\
\cline{2-9}
 & \multicolumn{2}{c|}{5-Fold Validation} & \multicolumn{2}{c|}{LOPO} & \multicolumn{2}{c|}{5-Fold Validation} & \multicolumn{2}{c|}{LOPO} \\
\cline{2-9}
 & Binary & Multi-class & Binary & Multi-class & Binary & Multi-class & Binary & Multi-class \\
\hline
mGradient Boosting & x & x & x & x & x & x & x & x \\ \hline
Catboost & x & x & x & x & x & x & x & x \\ \hline
LightGBM & x & x & x & x & x & x & x & x \\ \hline
SVM & x & x & x & x & x & x & x & x \\ \hline
Random Forest & x & x & x & x & x & x & x & x \\ \hline
MLP & x & x & x & x &  &  &  &  \\
\hline
\end{tabular}
\end{adjustbox}
\label{table:sensor_important_features}
\end{table}

\begin{table}[h!]
\centering
\renewcommand{\arraystretch}{1.2}
\caption{ {Dataset structure and validation configuration used in the case study.}}
\begin{adjustbox}{max width=\textwidth}
\begin{tabular}{|l|l|}
\hline
 {Category} &  {Count / Details} \\ \hline
 {Total Participants} &  {132 unique individuals} \\ \hline
 {Age Range} &  {18--94 years (Mean: 40 years)} \\ \hline
 {Geographic Distribution} &  {Participants from 19 countries} \\ \hline
 {80:20 Random Split} &  {Training: 862 samples / Testing: 216 samples} \\ \hline
 {k-Fold Cross-Validation} &  {5 patient-disjoint folds} \\ \hline
 {LOPO Cross-Validation} &  {132 folds (1 patient held out per fold)} \\ \hline
 {Total Features} &  {27 features (LDF + FS + Demographics)} \\ \hline
\end{tabular}
\end{adjustbox}
\label{table:dataset_structure}
\end{table}

\begin{table}[h!]
\centering
\renewcommand{\arraystretch}{1.2}
{\color{black}
\caption{ {Parameter configuration for the machine learning and deep learning models used in this study.}}
\begin{adjustbox}{max width=\textwidth}
\begin{tabular}{|l|p{11cm}|}
\hline
\textbf{Methods} & \textbf{Parameters} \\ \hline
Gradient Boosting & Learning rate = 1.0; n\_estimators = 100; random\_state = 0; criterion = friedman\_mse; max\_depth = 1 \\ \hline
CatBoost & Learning rate = 1.0; depth = 2 \\ \hline
LightGBM & Objective = binary; metric = binary\_logloss; num\_leaves = 31; learning\_rate = 0.1; n\_estimators = 100; verbose = -1 \\ \hline
SVM & Gamma = auto; C = 1.0; kernel = rbf; probability = True \\ \hline
Random Forest & Max depth = 2; random\_state = 0; n\_estimators = 100; criterion = gini; max\_features = sqrt \\ \hline
MLP & Learning rate = $1\times10^{-3}$; batch size = 32; dropout probability = 0.05; hidden size = 64 \\ \hline
EEGNet & Learning rate = $1\times10^{-5}$; dropout probability = 0.05; batch size = 32; C = 3; T = 9 \\ \hline
\end{tabular}
\end{adjustbox}
}
\label{table:parameter_information}
\end{table}

\begin{table}[ht]
\centering
\caption{Performance for different models: Wearable features with LOPO, binary classification.}
\begin{adjustbox}{max width=1\textwidth}
\setlength{\tabcolsep}{2pt}

\begin{tabular}{lcccccc}
\toprule
Model & Gradient Boosting & Catboost & LightGBM & SVM & Random Forest & MLP\\
\midrule
\textbf{ROC AUC} & 0.4730 & 0.5667 & 0.4933 & 0.5115 & 0.4829 & 0.5289\\
\textbf{PR AUC} & 0.7868 & 0.8171 & 0.8019 & 0.8089 & 0.8173 & 0.7900\\
\bottomrule
\end{tabular}
\end{adjustbox}
\label{tab:wearable_features_binary_LOPO}
\end{table}

\begin{table}[ht]
\centering
\caption{Performance for different models: Wearable features with k-Fold, binary classification.}
\begin{adjustbox}{max width=1\textwidth}
\setlength{\tabcolsep}{2pt}
\begin{tabular}{lcccccc}
\toprule
Model & Gradient Boosting & Catboost & LightGBM & SVM & Random Forest & MLP\\
\midrule
\textbf{ROC AUC} & 0.5076 & 0.5299 & 0.4973 & 0.5210 & 0.5066 & 0.5419\\
\textbf{PR AUC} & 0.8041 & 0.8109 & 0.8149 & 0.8162 & 0.8201 & 0.7969\\
\bottomrule
\end{tabular}
\end{adjustbox}
\label{tab:wearable_features_binary_kFold}
\end{table}

\begin{table}[ht]
    \centering
    \caption{Top 10 important features using Gradient Boosting, Random Forest, and LightGBM when conducting binary prediction with an 80:20 split.}
    \begin{adjustbox}{max width=1\textwidth}
    \setlength{\tabcolsep}{2pt}
        \begin{tabular}{c>{\raggedright\arraybackslash}p{1.5cm}c>{\raggedright\arraybackslash}p{1.5cm}c>{\raggedright\arraybackslash}p{1.5cm}cc}
            \toprule
            \multirow{2}{*}{\textbf{Order}} & \multicolumn{2}{c}{\textbf{Gradient Boosting}} & \multicolumn{2}{c}{\textbf{Random Forest}} & \multicolumn{2}{c}{\textbf{LightGBM}} \\
            \cmidrule(lr){2-3} \cmidrule(lr){4-5} \cmidrule(lr){6-7}
            & \centering Feature & Importance & \centering Feature & Importance & \centering Feature & Importance \\
            \midrule
            1 & M & 0.140254 & M & 0.162013 & Ae & 12 \\
            2 & A460 & 0.129633 & A460 & 0.125790 & Kv100 & 11 \\
            3 & A365 & 0.094329 & T & 0.117138 & Anadn & 8 \\
            4 & POM & 0.084874 & Ac & 0.109214 & M & 7 \\
            5 & Kv100 & 0.083186 & A365 & 0.102838 & T & 7 \\
            6 & Anadn & 0.073758 & POM & 0.090019 & A365 & 6 \\
            7 & T & 0.063749 & $\delta$ & 0.060251 & POM & 6 \\
            8 & $\delta$ & 0.061283 & Anadn & 0.052906 & An & 5 \\
            9 & Ae & 0.049073 & Ar & 0.037379 & F\_An & 5 \\
            10 & F\_Ar & 0.041633 & F\_Ac & 0.028451 & Ac & 5 \\
            \bottomrule
        \end{tabular}
    \end{adjustbox}
    \label{tab:merged_featuresbinary_prediction_2}
\end{table}



\clearpage 
\newpage
\onecolumn 

\clearpage 

\onecolumn
\tableofcontents

\end{document}